\documentclass{article} 
\usepackage{iclr2022_conference,times}

\usepackage{siunitx}        
\usepackage{multirow}       
\usepackage{graphicx}       
\usepackage{wrapfig}        
\usepackage{floatflt}
\usepackage{amsmath}
\usepackage{mathtools}      
\usepackage{float}          
\usepackage{upgreek}        
\usepackage{booktabs}
\usepackage{xurl}

\usepackage{amsmath,amsfonts,bm}

\def\eqref#1{equation~\ref{#1}}

\def\1{\bm{1}}

\DeclareMathAlphabet{\mathsfit}{\encodingdefault}{\sfdefault}{m}{sl}
\SetMathAlphabet{\mathsfit}{bold}{\encodingdefault}{\sfdefault}{bx}{n}

\usepackage{hyperref}
\usepackage{url}

\title{Fine-grained Differentiable Physics: A  Yarn-level Model for Fabrics}

\author{Deshan Gong\textsuperscript{1}, Zhanxing Zhu\textsuperscript{2,3},  Andrew J. Bulpitt\textsuperscript{1}  and He Wang\textsuperscript{1}\thanks{corresponding author}\\
\textsuperscript{1}School of Computing, University of Leeds\\
\textsuperscript{2}School of Informatics, University of Edinburgh\\
\textsuperscript{3}Peking University\\
{\footnotesize \texttt{\{scdg, A.J.Bulpitt, h.e.wang\}@leeds.ac.uk, zhanxing.zhu@pku.edu.cn}} \\
}

\iclrfinalcopy 
\begin{document}

\maketitle

\begin{abstract}
Differentiable physics modeling combines physics models with gradient-based learning to provide model explicability and data efficiency. It has been used to learn dynamics, solve inverse problems and facilitate design, and is at its inception of impact. Current successes have concentrated on general physics models such as rigid bodies, deformable sheets, etc, assuming relatively simple structures and forces. Their granularity is intrinsically coarse and therefore incapable of modelling complex physical phenomena. Fine-grained models are still to be developed to incorporate sophisticated material structures and force interactions with gradient-based learning. Following this motivation, we propose a new differentiable fabrics model for composite materials such as cloths, where we dive into the granularity of yarns and model individual yarn physics and yarn-to-yarn interactions. To this end, we propose several differentiable forces, whose counterparts in empirical physics are indifferentiable, to facilitate gradient-based learning. These forces, albeit applied to cloths, are ubiquitous in various physical systems. Through comprehensive evaluation and comparison, we demonstrate our model's \textit{explicability} in learning meaningful physical parameters, \textit{versatility} in incorporating complex physical structures and heterogeneous materials, \textit{data-efficiency} in learning, and \textit{high-fidelity} in capturing subtle dynamics. Code is available in: {\small \url{https://github.com/realcrane/Fine-grained-Differentiable-Physics-A-Yarn-level-Model-for-Fabrics.git}}
\end{abstract}

\section{Introduction}
Differentiable physics models (DPMs) have recently spiked interests, e.g. rigid bodies~\citep{heiden_interactive_2020}, cloth~\citep{liang_differentiable_2019}, and soft bodies~\citep{hu_2019_chainqueen}. The essence of DPMs is making physics models differentiable, so that gradient-based learning can be used to make systems adhere strictly to physical dynamics. This is realized via back-propagation through a series of observed actions, where the system can quickly learn the underlying dynamics. While enjoying neural networks' capability of modeling arbitrary non-linearity, DPMs also improve the model explicability as the learnable model parameters bear physical meanings. As a result, such models provide a new avenue for many applications such as inverse problems, e.g. estimating the mass of a moving rigid body~\citep{de_avila_belbute-peres_end--end_2018}, and control, e.g. learning to shake a bottle to shape the fluid in it~\citep{li_learning_2019}. 

Early research attempted to model simple and general physical systems such as rigid bodies~\citep{de_avila_belbute-peres_end--end_2018}, followed by a range of systems including deformable objects~\citep{li_learning_2019}, cloth~\citep{liang_differentiable_2019}, contacts~\citep{zhong_differentiable_2021}, etc. However, existing models are only generally-purposed which do not consider complex structures/topologies and force interactions. Taking cloth (i.e. fabrics) as an example, existing models~\citep{liang_differentiable_2019,li_learning_2019} can learn general cloth dynamics, but only when the cloth is relative simple and homogeneous. Recent research~\citep{wang_first_2020} has started to explore articulated systems but the model capacity is insufficient to capture the full dynamics of complex systems such as fabrics. Since real-world physical systems (e.g. materials in engineering) often have sophisticated structures and consist of heterogeneous materials, we argue that it is crucial to design fine-grained DPMs, for differentiable physics to be truly applicable and meaningful to real-world applications. 

This paper focuses on fabrics which are composite materials consisting of basic slim units arranged in different patterns. A common example in fabrics is woven cloth which is made from yarns of different materials (silk, cotton, nylon, etc.) interlaced in various patterns (e.g. plain, satin, twill). Fabrics present new challenges in differentiable modeling. First, the dynamics heterogeneity caused by material and structural diversity needs to be incorporated into modeling, which is especially crucial for solving inverse problems where the physical properties are learned from data. General DPMs without sufficient granularity can only approximate the dynamics and are unable to learn meaningful parameters. Second, certain forces that are essential for fabrics dynamics are indifferentiable. One such example is friction. The standard Coulomb model for rigid bodies has been made differentiable recently~\citep{de_avila_belbute-peres_end--end_2018,zhong_differentiable_2021}. However, it is overly simplified for fabrics because the yarn-to-yarn friction shows richer dynamics~\citep{zhou_overview_2017} that is beyond the capacity of existing methods. Further, the contact modeling together with friction requires new treatments that previous methods did not have to consider.

To overcome these challenges, we propose a new DPM for fabrics at a \emph{more fine-grained} level and apply it to cloth modeling. Unlike general DPMs, we start with a fine-grained yarn-level model. By modeling each yarn individually, we provide the capacity of modeling fabrics with mixed yarns and different woven patterns, which could not be handled previously. To facilitate gradient-based learning, we propose new differentiable forces on/between yarns, including contact, friction and shear. Finally, we incorporate implicit Euler and implicit differentiation to compute gradients induced by an optimization problem embedded in the simulation.


To our best knowledge, our model is the first differentiable physics model which provides sufficient granularity for heterogeneous materials such as fabrics. We comprehensively evaluate its learning capability, data efficiency and fidelity. Since there is no similar model, we compare our model with the most similar work~\citep{liang_differentiable_2019} and traditional Bayesian optimization on inverse problems. We also compare our work on control learning with popular Reinforcement Learning methods. We show that our model is more explicable, has higher data efficiency, generates more accurate predictions in inverse problem and control learning respectively.

\section{Related work}
\label{gen_inst}

\textbf{Differentiable physics simulator.} A differentiable simulator integrates differentiable physics engine into the forward and backward propagation of learning. As a strong inductive bias, these simulation engines increase data efficiency and learning accuracy over gradient-free models. Due to these advantages, differentiable simulation demonstrates superiority in a number of problem domains such as inverse problem, robot control and motion planning. The early works focused initially on simple rigid bodies~\citep{de_2018_end, degrave_2019_differentiable} and later simulation of high degrees of freedom systems, such as fluids~\citep{schenck_2018_spnets}, elastic bodies~\citep{hu_2019_chainqueen, huang_2021_plasticinelab}, and cloth~\citep{liang_differentiable_2019}. More recently, \citet{jatavallabhula_2021_gradsim} introduced an end-to-end differentiable simulator that can learn from images by combining differentiable rendering and differentiable simulation. Comparatively, we explore fine-grained DPMs for composite materials, which leads to new challenges in differentiable modeling.

\textbf{Cloth simulation.} Cloth simulation initially appeared in textile engineering and was then introduced to computer graphics~\citep{long_2011_cloth}. Cloth has been modeled as particle systems~\citep{breen_1992_physically}, mass-spring systems~\citep{provot_1995_deformation}, and continuum~\citep{narain_2012_adaptive}. \citet{kaldor_2008_simulating} proposed a yarn-level knit cloth simulator and found that cloth microstructures have a considerable influence on cloth dynamics. Since then the cloth simulation community has shifted the focus to yarn-level cloth simulation. Based on the objectives, the recent research can be classified to increasing efficiency~\citep{kaldor_2010_efficient, cirio_2016_yarn}, combining continuum models and yarn-level models~\citep{casafranca_2020_mixing, sperl_2020_homogenized}, introducing woven cloth simulation~\citep{cirio_yarn_2014}, and optimizations~\citep{pizana_2020_bending, sanchez_2020_robust}. Our work is orthogonal to these papers in that we introduce a new methodology to incorporate differentiable physics into yarn-level models.

\textbf{Machine learning and cloth simulation.} Machine learning was initially introduced to cloth simulation to make data-driven simulators, which have inherent advantages in simulation efficiency over physical-based methods \citep{james_2003_precomputing, kim_2008_drivenshape}, and can help improve fidelity~\citep{lahner2018deepwrinkles}. In parallel, machine learning has been applied to discover the physical properties from visual information. \citet{bouman_2013_estimating} proposed a linear regression model for evaluating cloth density and stiffness from the dynamics of wind-blown cloth. \citet{yang2017learning} introduced a neural network for classifying cloths based on how their dynamics are affected by stretching and bending stiffness. \citet{rasheed_2020_learning} proposed a model for estimating the friction coefficient between cloth and other objects. By combining physically-based cloth simulators and neural networks, \citet{runia_2020_cloth} estimated cloth parameters by training neural networks to adjust a simulator's parameters so that the simulated cloth mimics the observed one in videos. Different from these gradient-free models, \citet{liang_differentiable_2019} and \citet{li2021diffcloth} proposed sheet-level differentiable cloth models that can be used to estimate cloth parameters. In this work, we dive into fine-grained physics and propose a new yarn-level differentiable fabrics model which can be embedded into deep neural networks as a layer. 

\section{Methodology}
Since cloth is employed as an application in this paper, we use the terms `cloth' and `fabric' interchangeably. We first explain the cloth representation (Sec.~\ref{sec:representation}) and the (physics) system equation for simulation (Sec.~\ref{sec:systemEquation}). Then we present our new force models (Sec.~\ref{sec:force}), and how we solve the system equation to enable back-propagation (Sec.~\ref{sec:physicsSolve}).

\label{headings}
\subsection{Cloth representation}
\label{sec:representation}
\begin{wrapfigure}[14]{r}{0.3\textwidth}
    \centering
    \includegraphics[trim=0 0 0 110, width=0.3\textwidth]{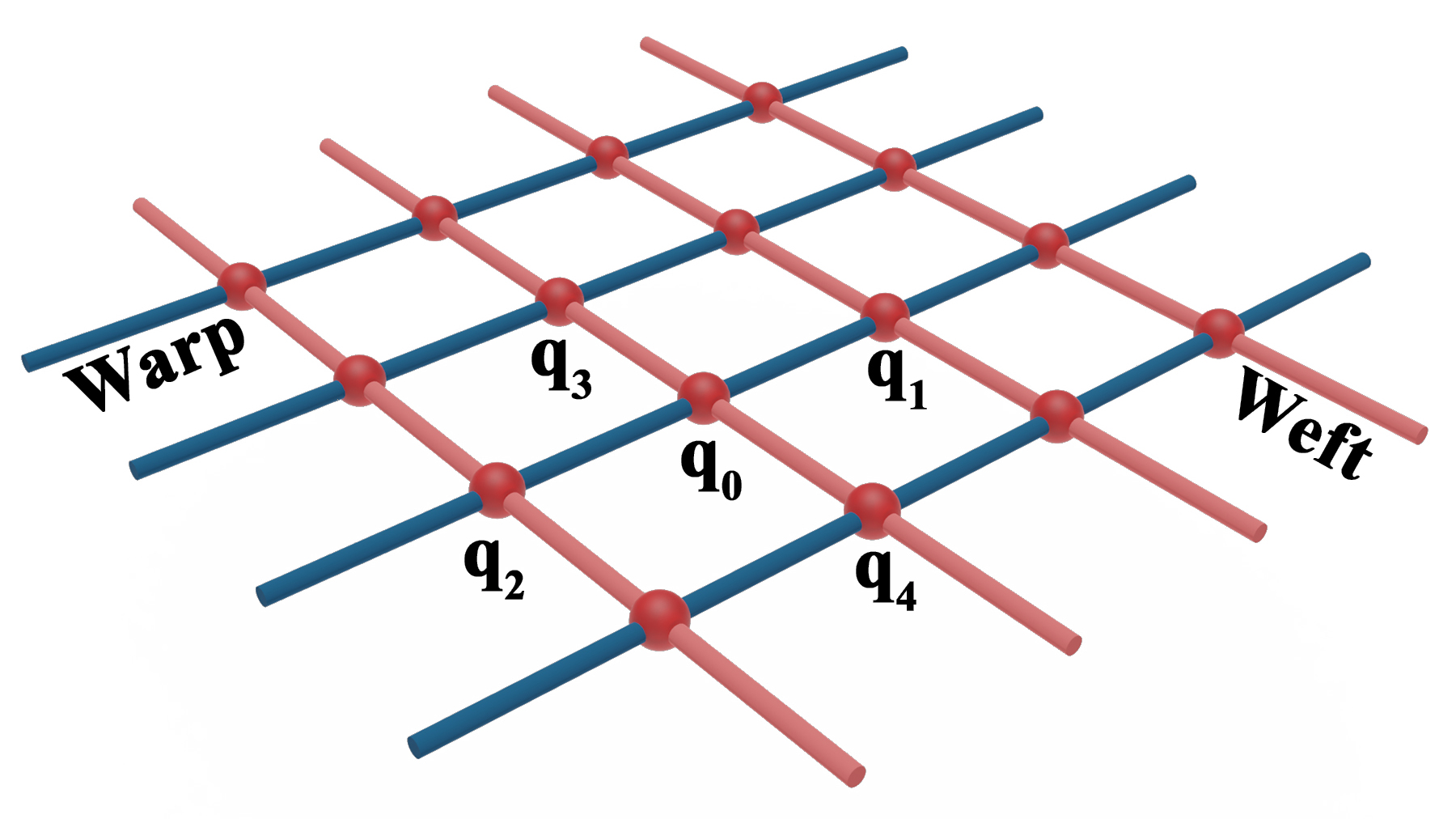}
    \caption{Blue and red rods denote warps and wefts respectively. $\mathbf{q}$s are the crossing nodes.}
    \label{fig:grid}
\end{wrapfigure}

Similar to~\citet{cirio_yarn_2014}, our cloth consists of two perpendicular groups of parallel yarns named \textit{warps} and \textit{wefts}. Every pair of warp and weft are in contact with each other at one crossing node (Figure \ref{fig:grid}), with a persistent contact. We employ an Eulerian-on-Lagrangian discretization~\citep{sueda_large_2011}, and denote the Degrees of Freedom (DoFs) of every crossing node as $\mathbf{q}_i \equiv (\mathbf{x}_i,u_i,v_i)$. $\mathbf{x}_i\in \mathbb{R}^3$ is the Lagrangian coordinates indicating spatial locations and $(u_i,v_i)$ is the Eulerian coordinates indicating sliding movements between yarns. The end points of yarns do not contact with other yarns and hence  they are treated as special crossing nodes that have no Eulerian terms, i.e. $\mathbf{q}_j \equiv \mathbf{x}_i$. Therefore, on a $r (rows) \times c (columns)$ cloth, there are $ (r-2) \times (c-2)$ crossing nodes with five DoFs and $ 2r + 2c - 4$ crossing nodes with three DoFs. Every two neighboring crossing nodes on the same warp/weft delimit a warp/weft segment. A warp segment with end points $\mathbf{q}_0$ and $\mathbf{q}_1$ is denoted as $\mathbf{[q_0,q_1]}$ and its position is $(\mathbf{x}_0,\mathbf{x}_1,u_0, u_1)$ (Figure~\ref{fig:grid}). This way, a woven cloth is discretized into crossing nodes and segments which are the primitive units of the cloth. Every segment is assumed to be straight so that linear interpolation can be employed on the segment, i.e. the
spatial location of a point in the segment $\mathbf{[q_0,q_1]}$ is $\mathbf{x}(u) = \frac{u - u_0}{\Delta u} \mathbf{x}_0 + \frac{u_1 - u}{\Delta u} \mathbf{x}_1$, where $u$ is the point's position in Eulerian coordinates and $\Delta u = u_1 - u_0$ is the crossing nodes distance in Eulerian coordinates. We use $L$ to denote the rest length of the yarn segment and $R$ to denote the yarn radius.

\subsection{System equation for simulation}
\label{sec:systemEquation}
A cloth’s state at time $t$, $\mathcal{S}_{(t)}=\{\mathcal{Q}_{(t)},\dot{\mathcal{Q}}_{(t)}\}$, includes all the crossing node DoFs $\mathcal{Q} = \{\mathbf{q}_i |i=1,2,\dots,N\}$ and their velocities $\dot{\mathcal{Q}}=\{\dot{\mathbf{q}}_i | i=1,2, \dots,N\}$, where $N$ is the number of crossing nodes. Knowing the state, we can calculate the internal and external forces:
\begin{equation}
\label{eqn:Lag}
    \mathbf{F} = \mathbf{M}\ddot{\mathbf{q}} = \frac{\partial T}{\partial \mathbf{q}} - \frac{\partial V}{\partial \mathbf{q}} - \dot{\mathbf{M}}\dot{\mathbf{q}}
\end{equation}
where $\mathbf{q}$, $\dot{\mathbf{q}}$, and $\ddot{\mathbf{q}}$ are the \emph{general} position, velocity, and acceleration respectively, with a dimension $l = 3 \times r \times c + 2 \times (r-2) \times (c-2)$. $\mathbf{M} \in \mathbb{R}^{l \times l}$ is the general mass matrix.  The model assumes mass is distributed homogeneously. $T$ and $V$ are the kinetic and potential energy. As force is related to the partial derivative of energy with respect to position, the right hand terms in Eq.~\ref{eqn:Lag} are inertia, conservative forces, and part of the time derivative of $\mathbf{M}\dot{\mathbf{q}}$. Non-conservative forces are added to the right side of the equation. Section~\ref{sec:force} gives the details of all the forces considered in our model. Using implicit Euler~\citep{baraff_large_1998}, we can derive the system equation for simulation:


\begin{equation}
\label{eqn:linear}
    \left(\mathbf{M} - \frac{\partial \mathbf{F}_{(t)}}{\partial \mathbf{q}}h^2 - \frac{\partial \mathbf{F}_{(t)}}{\partial \dot{\mathbf{q}}}h\right)\dot{\mathbf{q}}_{(t+1)} = h \left(\mathbf{F}_{(t)} - \frac{\partial \mathbf{F}_{(t)}}{\partial \dot{\mathbf{q}}} \dot{\mathbf{q}}_{(t)} \right) + \mathbf{M} \dot{\mathbf{q}}_{(t)}
\end{equation}
where the subscript in brackets $t$ indicates the associate variable at time $t$. Detailed deduction is in Appendix.
\subsection{Force models}
\label{sec:force}
To simulate cloth, we need to compute the inertia, internal and external forces in Equation \ref{eqn:linear}. Inertia is the derivative of kinetic energy with respect to node positions. As woven cloths are interlaced yarns, the internal forces can be further classified into 1) forces caused by yarn deformation and 2) forces resulting from yarn-to-yarn interactions. We treat each yarn as an elastic rod that can generate elastic energy, including stretching and bending~\citep{jawed_2018_primer}, ignoring the twisting due to its triviality in cloth dynamics~ \citep{cirio_yarn_2014}. The elastic energy is $V^{e} = V^{s} + V^{b}$, where $V^{s}$ and $V^{b}$ are the stretching and bending energy respectively. The yarn-to-yarn interaction forces include friction, shear, and parallel yarn collisions. We refer the reader to Appendix for the inertia, stretching, and bending force as it is straightforward to show that they are differentiable.

\begin{wrapfigure}[11]{r}{4.5cm}
    \centering
    \includegraphics[width=0.3\textwidth, trim= 210 40 210 110]{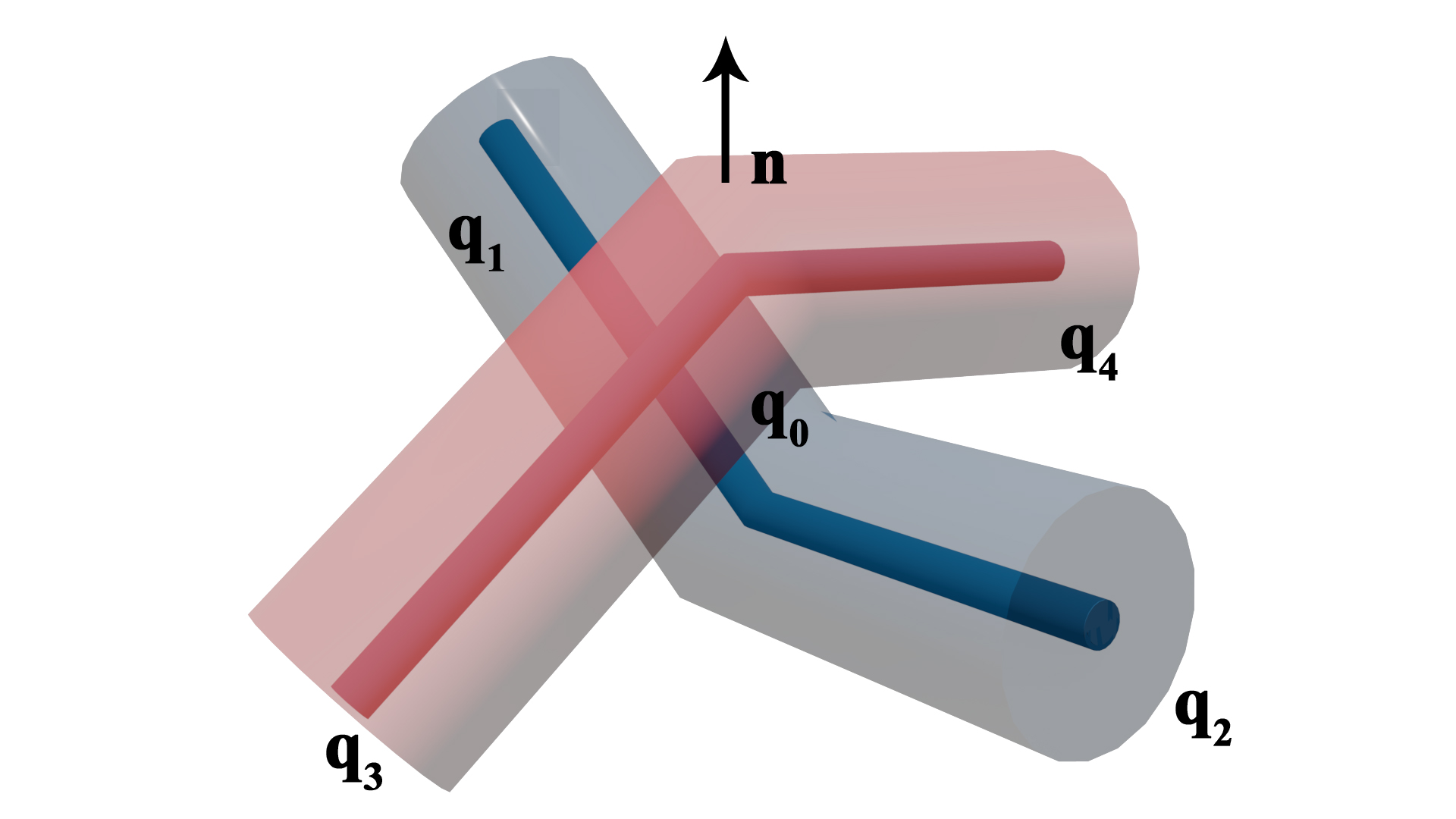}
    \caption{Compression force on $\mathbf{q}_0$ along normal $\mathbf{n}$ at $\mathbf{q}_0$. }
    \label{fig:compression}
\end{wrapfigure}
\textbf{Yarn-to-yarn contact}. 
While the aforementioned forces are differentiable, the yarn-to-yarn forces are not. Existing differentiable contact models mainly correct after-contact positions and velocities~\citep{de_avila_belbute-peres_end--end_2018,liang_differentiable_2019,zhong_differentiable_2021} via (multiple) optimization solves, which is too simplistic for fabrics. Yarn-to-yarn contact has its unique features. It is relatively sticky and often has small relative velocities. We need a contact model that reflects this and leads to a differentiable contact force which affects the friction/shear. The contact force is a combination of the stretching $\mathbf{F}_s$ and bending forces $\mathbf{F}_b$ at every crossing node along the contact normal $\mathbf{n}$. We assume no-slip contact and compute the contact force by:
\begin{equation}
    F_n = \mbox{ReLU}(\frac{1}{2} \mathbf{n}^{\top}(\mathbf{F}_s^{u} + \mathbf{F}_b^{u} - \mathbf{F}_s^{v} - \mathbf{F}_b^{v}))
\end{equation}
where $u$ and $v$ represent the forces from warp and weft segments. The rectified linear unit (ReLU) ensures the non-negativity of the contact force. The normal $\mathbf{n}$, from warp to weft yarn (Fig. \ref{fig:compression}), is approximated by the normal of the best-fit plane of $\mathbf{q_0}$-$\mathbf{q_4}$.

\textbf{Friction}. The friction between warps and wefts prohibits relative movements, which is crucial to the overall dynamics of the fabric. In differentiable physics, contacts under simple settings have been modeled, such as the standard Coulomb model for kinetic friction~\citep{zhong_differentiable_2021}. But this is insufficient for our purpose for two reasons. First, the static friction plays a key role in stick-slip behaviours of yarns~\citep{zhou_overview_2017} and needs to be modeled. Second, the standard Coulomb friction model is a piece-wise function, which is intrinsically indifferentiable at the static-to-kinetic transition point. Therefore, we need a new differentiable friction model.

\begin{figure}[tb]
    \centering
    \includegraphics[width=\textwidth]{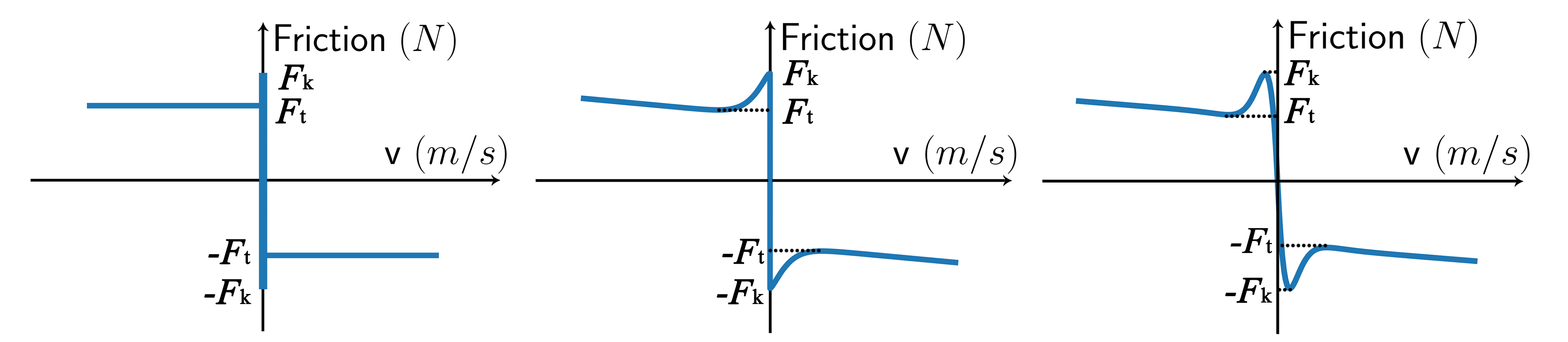}
    \caption{$F_k$ and $F_t$ are the static and kinetic friction. Coulomb model (left) is an indifferentiable multi-value function. Stribeck effect (middle) is empirically observed~\citep{stribeck_wesentlichen_1902}. Our model (right) incorporates the Stribeck effect and also simulates self-excited vibrations around $\mathsf{v}=0$.}
     \label{fig:friction}
 \end{figure}

The low relative speed between yarns is a special situation where the static-to-kinetic transition could actually be continuous (as opposed to the Coulomb model), experimentally shown by Stribeck~\citep{stribeck_wesentlichen_1902}. This indicates that a continuous and differentiable model has the potential to be more accurate for yarns than the widely used Coulomb model. Further, the breakaway force causing the static-to-kinetic transition depends on the rate of the external force~\citep{johannes_role_1973}, and the nonlinear stick-slip behavior is related to self-excited vibrations before transition~\citep{awrejcewicz_chaotic_1988}. Inspired by the above research, we propose a new differentiable yarn-to-yarn friction model (Figure~\ref{fig:friction}):
\begin{equation}
\label{eq:friction}
    F_{Slide} = -\Big( 
    \frac{k_f\delta u - K(\delta u)\mu F_n}{2} K(\mu F_n - F_u)
    + \frac{k_f\delta u + K(\delta u)\mu F_n}{2} \Big) - d_f \dot{u}_0
\end{equation}
where $\delta u = u_0 - \bar{u}_0$ and $K(x) = \tanh(px)$. $\bar{u}_0$ is the anchor position when there is no relative movement between the warp and the weft segment, $\mu$ is the friction coefficient, and $F_u$ is the external force. We introduce a hyperparameter $p$ to control the conversion speed between static and kinetic friction. To understand Equation~\ref{eq:friction}, there are three situations: $F_u = 0$ (no external force), $0 \leq F_u \leq \mu F_n$ (static friction), and $F_u > \mu F_n$ (kinetic friction).  When $F_u = 0$, $\delta u=0$, $K(\mu F_n - F_u)=1$ and the speed $\dot{u}_0=0$, so $F_{Slide}=0$; when $0 \leq F_u \leq \mu F_n$, we allow a small displacement $\delta u$ to mimic the self-excited vibration in static friction, governed by a Hooke's spring $k_f\delta u$ with stiffness $k_f$. When $F_u$ is small, i.e. $K(\mu F_n - F_u)$ is close to 1, $F_{Slide} \approx -k_f\delta u - d_f \dot{u}_0$ where $d_f$ is a damping coefficient. $F_{Slide}$  is mainly the static friction minus a small damping term (as $\dot{u}_0$ is small). Due to the time discretization in simulation, the spring force has a delayed response which causes small-range vibrations. When $K(\mu F_n - F_u)$ starts to decrease to 0 and the breakaway force is achieved $F_u = \mu F_n$, $F_{Slide} = -\frac{k_f\delta u + K(\delta u)\mu F_n}{2} - d_f \dot{u}_0$  and $\frac{k_f\delta u + K(\delta u)\mu F_n}{2}$ is the average of the spring force and the maximum static friction. Finally when $F_u > \mu F_n$, $K(\mu F_n - F_u)$ quickly becomes -1 and $K(\delta u)$ becomes 1 as $\delta u$ increases. Then $F_{Slide} = -\mu F_n - d_f \dot{u}_0$, which is the kinetic friction minus damping. Fig.~\ref{fig:friction} shows our friction can closely approximate the Stribeck effect while maintaining differentiability, as opposed to the indifferentiable Coulomb model. Also, it incorporates self-excited vibrations within $F_{Slide} \in [-F_k, F_k]$ which is when $0 \leq F_u \leq \mu F_n$.
 
\begin{wrapfigure}[12]{r}{7cm}
    \centering
    \includegraphics[trim=0 0 0 85, width=0.5\textwidth]{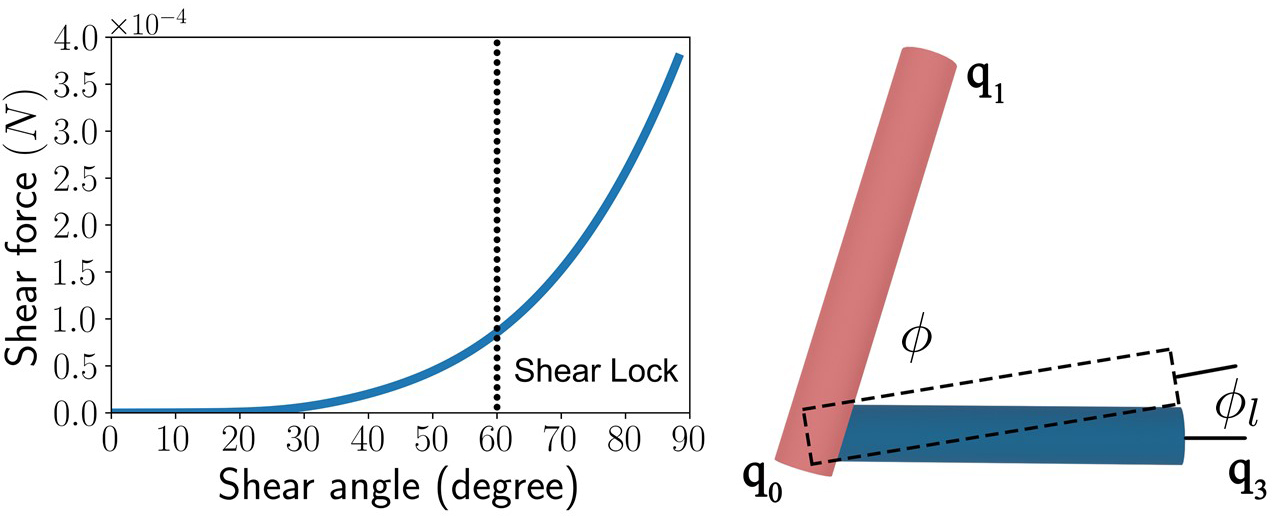}
    \vspace{-5pt}
    \caption{Shear force strength vs shear angle $\bar{\phi} - \phi$, (left) and graphical illustration (right).}
    \label{fig:shear}
\end{wrapfigure}

\textbf{Shear}. A shear force is generated when there is relative rotation between a warp and a weft at a crossing node~\citep{parsons_impact_2010}, which increases non-linearly when the shear angle increases~\citep{mohammed_shear_2000,peng_experimental_2004,Cao_Characterization_2008}. Previous differential models do not consider this type of forces. Therefore, we propose a new differentiable shear model. There are different stages when the shear angle increases~\citep{king_continuum_2005}. The shear force first grows almost linearly initially, then `shear lock' is triggered~\citep{wang_draping_1999} when the angle passes a threshold and the shear force starts to increase exponentially as the angle increases, producing highly non-linear behaviors. We therefore define the shear energy as a function of the shear angle $\bar{\phi}-\phi$ (Figure~\ref{fig:shear}): $\frac{1}{2} k_s L (\phi - \bar{\phi})^2$, 
where $\bar{\phi}=\frac{\uppi}{2}$ is the rest shear angle. $k_s = S \uppi R^2 (1 + F_n)$ is the shear stiffness and $S$ is the shear modulus. We embed the `shear lock' by boosting $k_s$ exponentially with $\gamma = (\sqrt{2L^2} - 2\sin\frac{\phi}{2}L)/R$ as long as it stays smaller than the lock threshold $\phi_l = 2\arcsin{\frac{R}{L}}$: $k_s$ equals to $S\uppi R^2 (1 + F_n)$ if $\phi > \phi_l$; and $S\uppi R^2 (1 + F_n) \gamma^c$ otherwise, where $c$ controls the increase rate of $k_s$ with respect to $\phi$ when `shear lock' occurs. Although $k_s$ has discontinuities within $[0, \frac{\phi}{2}]$, our new shear stiffness can be defined as:
\begin{equation}
    k_s = \frac{1}{2} (F_n + 1) S \uppi R^2 \Bigg( (1 + \gamma^c) + (1 - \gamma^c)
    \tanh\left( 
    \frac{\bar{\phi}^5(\phi-\phi_l)}
    {(\phi(\phi-\phi_l)(\phi-\bar{\phi}))^2 + \bar{\phi}^4 \sigma^2}
    \right) \Bigg)
\end{equation}
where $\sigma$ governs the transition smoothness between lock and no-lock. The smaller the $\sigma$ is, the smoother the transition is. The shear force (Figure~\ref{fig:shear}) derivation is in Appendix.

\textbf{Yarn-to-yarn collision}. The last internal force is the yarn-to-yarn collisions between parallel yarns. Although it is theoretically possible to use an existing approach~\citep{de_avila_belbute-peres_end--end_2018,liang_differentiable_2019}, it would require forming an optimization for all segments and therefore become prohibitively slow. Therefore, we introduce a new penalty energy, defined as a function of the nodes' distance in Eulerian coordinates for a warp segment $[\mathbf{q}_0,\mathbf{q}_1 ]$ (similar for a weft segment): $V_{0,1} 
    = \frac{1}{2} k_c L (\mbox{ReLU}(d - \Delta u))^2$
where $d=4R$ or $2R$ which is elaborated in Appendix.


\textbf{External forces and collisions}. Without loss of generality, we consider two external forces: gravity and wind force. Their impacts can be modeled by defining proper potential energies. Finally, after calculating all forces, the resultant force at every crossing node is the combined force of all segments that connect to that node. The cloth is simulated by solving Equation~\ref{eqn:linear}. We use bounding volume hierarchy \citep{tang_fast_2010} with continuous collision detection and non-rigid impact zones~\citep{harmon_robost_2008} to compute collision. Similar to~\citet{liang_differentiable_2019, wang_first_2020}, we form an optimization problem for continuous collision. Details can be found in Appendix.

\subsection{Derivatives of the simulator}
\label{sec:physicsSolve}
Now we have a fully differentiable simulator with parameters $\boldsymbol{\omega}$.  The $\boldsymbol{\omega}$ is the cloth physical parameters (stretching, bending, shearing, etc) when solving inverse problems, or the to be learned external forces in control experiments. Given a loss function $\mathcal{L}$, its gradient with respect to the parameters $\frac{\partial \mathcal{L}}{\partial \boldsymbol{\omega}}$ can help learn the right physics parameters via back-propagation. Implicit differentiation can be used to derive $\frac{\partial \mathcal{L}}{\partial \boldsymbol{\omega}}$ as detailed in Appendix. Finally, we define the loss function: $\mathcal{L}(\mathbf{q}, \hat{\mathbf{q}}) = \frac{1}{NT}\sum_{n=1}^{N}\sum_{t=1}^{T}
    \|\mathbf{q}_{n,t} - \hat{\mathbf{q}}_{n,t}\|^2_2$,
where $\mathbf{q}_{n,t}$ and $\hat{\mathbf{q}}_{n,t}$ are the ground-truth and predicted general position. $N$ and $T$ are the total number of nodes and simulation steps respectively. We use Stochastic Gradient Descent and run 70 epochs for training.



\textbf{Underconstrainedness Mitigation.} Learning physical parameters via solving an inverse problem is intrinsically under-constrained, leading to multiple solutions or implausible parameter values when fitting data, e.g. unconstrained learning leads to negative density. We mitigate this issue by incorporating prior knowledge. Instead of directly learning parameters $\boldsymbol{\omega}$, we set $\boldsymbol{\omega} = a\times \mbox{sigmoid}(y) + b$ where $a$ and $b$ are tunable scalars, and we learn $y$ instead. $a$ and $b$ essentially limit the range of $\boldsymbol{\omega}$. We can induce prior knowledge of parameter ranges such as yarn density ranges, because although the exact value is to be learned and not known \textit{a priori}, their ranges are available in practice. Our experiments demonstrate that this strategy effectively mitigates the multi-solution issue.

\section{Experiments}
\begin{wrapfigure}[11]{r}{8.0cm}
    \centering
    \includegraphics[trim=0 0 0 160, width=0.55\textwidth]{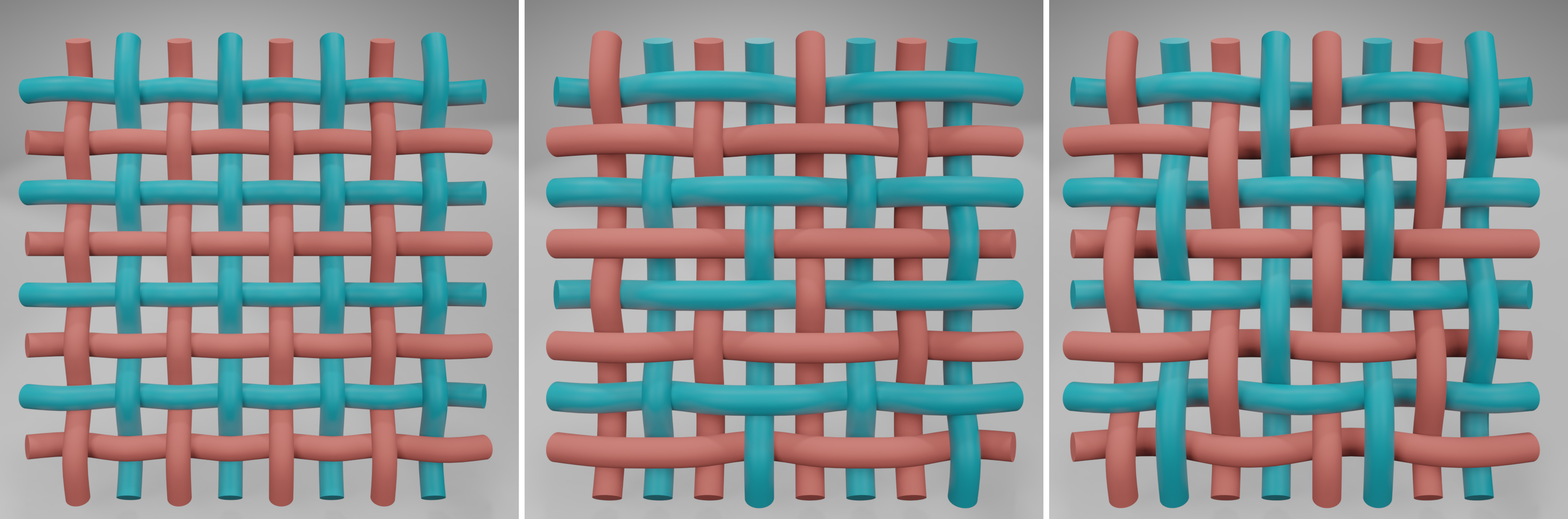}
    \caption{Woven patterns. Left-to-right: plain, satin, and twill. Teal and coral indicate different yarns.}
    \label{fig:woven_pattern}
\end{wrapfigure}

We employ a traditional indifferentiable yarn-level simulator~\citep{cirio_yarn_2014} to generate the ground-truth data, and build a dataset of fabrics with three types of yarns and three types of woven patterns. The yarns vary in density, elastic modulus, and bending modulus (Table \ref{tab:yarns}). The three woven patterns include plain, twill and satin. We use hybrid fabrics made from two types of yarns, and exhaustively combine three yarns with three woven patterns (Figure \ref{fig:woven_pattern}) to generate 
9 types of fabrics. We denote them as XXX-(X, X) where the prefix is the woven pattern and the numbers in the brackets are the yarns, e.g. Plain-(1, 2) means a plain pattern woven with Yarn1 and Yarn2. In our ground-truth data, we use a square piece of cloth hanging at its two corners and blown by wind with a constant magnitude. The simulation is conducted for 500 steps with $h=0.001s$. Training details are in Appendix.

\subsection{Learning physical parameters}
We first demonstrate our model's effectiveness in learning meaningful physical parameters, under various model sizes and different amounts of training data. 

\begin{table}[h]
    \centering
    \vspace{-0.3cm}
    \caption{Ground-truth parameters of three yarns.}
    \vspace{1ex}
    \begin{tabular}{lccc}
    \toprule
    Parameter/Yarn & Yarn1 &  Yarn2  & Yarn3 \\
    \midrule
    Density($kg/m$)  & 0.0020 & 0.0025 & 0.0024\\
    Stretch modulus($N/m$) & 500000  & 170000 & 120000\\
    Bending modulus($N/m$) & 0.00014 & 0.00011 & 0.00009 \\
    \bottomrule
    \end{tabular}
\label{tab:yarns}
\end{table}

\textbf{Learning capacity.} We first test whether meaningful physical parameters can be learned from cloths of different sizes. Small-size cloths tend to show low-frequency features, e.g. the general shape, as opposed to high-frequency features, such as wrinkles and buckling. We test our model on simulation data with sizes: 5$\times$5, 10$\times$10, 17$\times$17 and 25$\times$25, trained on the first 25 frames. Table~\ref{tab:all_parameters} shows that our model can effectively estimate yarn parameters with underlying physics models of different sizes. This has several implications. First, although cloth size does affect the overall dynamics of the motion in the ground-truth data (e.g. larger cloths have more wrinkles), it does not affect our model's learning capability. Second, since our model can reliably learn the yarn parameters on a small fraction of a cloth, it saves the computation of learning from large cloths, which improves the learning scalability. We can learn from small cloths and then scale to simulate large cloths. Further, inter-yarn parameters including shear $S$ and friction coefficient $\mu$ are highly correlated so that the model can easily end up learning only plausible parameter values rather than the true values. This is where we expect our model to suffer from the under-constrainedness problem as~\citet{liang_differentiable_2019}. Surprisingly, our model can learn the right parameters under different sizes. By introducing prior knowledge as aforementioned, the learned parameters are restricted within valid ranges. 

\vspace{-10pt}
\begin{table}[h]
    \caption{Inter/intra parameters learned on Plain-(1, 2), with ground-truth $S=1000Pa$, $\mu=0.5$. }
    \label{tab:all_parameters}
    \centering
    \begin{tabular}{ccccccc}
    \toprule
        Size & Shear $S$ & Friction $\mu$ & Yarn & Density & Stretch & Bend \\
        \midrule
         \multirow{2}*{$5\times5$} & \multirow{2}*{\num{949}} & \multirow{2}*{0.437}
         & 1 & \num{2.028e-3} & 479523 & \num{1.387e-4}\\
         & & & 2 & \num{2.450e-3} & 172928 & \num{1.112e-4}\\
         \multirow{2}*{$10\times10$} & \multirow{2}*{\num{932}} & \multirow{2}*{0.455}
         & 1 & \num{1.991e-3} & 484719 & \num{1.325e-4}\\
         & & & 2 & \num{2.448e-3} & 173843 & \num{1.026e-4}\\
         \multirow{2}*{$17\times17$} & \multirow{2}*{947} & \multirow{2}*{0.402}
         & 1 & \num{1.969e-3} & 505421 & \num{1.323e-4}\\
         & & & 2 & \num{2.440e-3} & 171304 & \num{1.034e-4}\\
         \multirow{2}*{$25\times25$} & \multirow{2}*{913} & \multirow{2}*{0.380}
         & 1 & \num{2.069e-3} & 510215 & \num{1.488e-4}\\
         & & & 2 & \num{2.443e-3} & 173920 & \num{1.201e-4}\\
         \bottomrule
    \end{tabular}
\end{table}

\textbf{Data efficiency.} Data efficiency is crucial as obtaining the ground-truth data can be expensive. Precise 3D geometry capture of real cloths is difficult and time-consuming, while simulation of high-res cloths is prohibitively slow. We further investigate the data efficiency by varying the amount of training data. We gradually increase the training data from the first 5 frames to the first 25 frames. Table~\ref{tab:frames} shows that our model has high data efficiency. It can learn reasonably well from as few as the first 5 frames. The benefits are two-fold. First, our model needs just a few frames to train, making it highly applicable. The second benefit is bigger but less obvious. The first 5 frames (from a static pose) normally contains little dynamics as the cloth just starts to move. This indicates that our model only requires a few frames of low-dynamics motions. This eases real-world measurements on cloth because no large motions are needed. This also saves time if simulation data is used, as small time step size is usually demanded in high-dynamic motion simulations.

\begin{table}[h]
\vspace{-10pt}
    \caption{Plain-(1,2) learnt parameters on different training data. Left: Yarn1, Right:Yarn2.}
    \label{tab:frames}
    \centering
    \begin{tabular}{ccccccc}
    \toprule
        Frames & Density & Stretch & Bend & Density & Stretch & Bend \\
        \midrule
         5 & \num{2.030e-3} & 494301 & \num{1.357e-4} & \num{2.450e-3} & 169597 & \num{1.130e-4} \\
         10 & \num{2.037e-3} & 491717 & \num{1.379e-4} & \num{2.443e-3} & 169543 & \num{1.130e-4}\\
         25 & \num{2.038e-3} & 491873 & \num{1.367e-4} & \num{2.447e-3} & 167217 & \num{1.096e-4} \\
         \bottomrule
    \end{tabular}
\end{table}

All simulations can be found in the supplementary video. We also include simulations with collisions and simulations on large cloths using parameters learnt on small cloths. More results and details are in Appendix.

\subsection{Comparisons}
\subsubsection{Prediction \& Data Efficiency}

To our best knowledge, there is no similar fine-grained DPM in the literature. The closest method is a general sheet model~\citep{liang_differentiable_2019}, so we compare our model with theirs. We employ their settings, and use a 17$\times$17 model and 50 frames simulation data, with 5, 10 and 25 frames for training and the whole 50 for testing. Since the two methods model cloths at different levels of granularity, their physical parameters are not directly comparable. We therefore compare their Mean Squared Error (MSE). We also include a traditional parameter estimation method based on Bayesian Optimization (BO) \citep{snoek_2012_practicalBO} combined with a yarn-level simulator~\citep{cirio_yarn_2014} as another baseline. In BO, we randomly select 5 initial points and use the expected improvement \citep{jones_1998_efficient} as the acquisition function. As the learning process of differentiable simulation consists of forward simulation and backward simulation, training 70 epochs can be considered as running 140 simulations. Therefore, we run 140 iterations when using BO. Moreover, we impose the same parameter ranges in the BO as we did in our model. 

\begin{table}[t]
\vspace{-15pt}
    \caption{Testing errors ($\times10^{-6}$) of our model (left) and \citep{liang_differentiable_2019} (middle) and BO (right) trained on 5, 10 and 25 frames generated by yarn-level simulator~\citep{cirio_yarn_2014}.}
    \label{tab:yarn_vs_sheet_vs_bo}
    \centering
    \begin{tabular}{ccc|cc|cc}
    \toprule
    Fabrics/Frames & 5 & 25 & 5 & 25 & 5 & 25\\
    \midrule
    Plain-(1,2) & \num{1.152e-4} & \num{3.962e-5} & 1.461 & 0.4124 & \num{0.512} & \num{0.109} \\
    Plain-(1,3) & \num{1.516e-4} & \num{3.555e-5} & 1.608 & 0.4567 & \num{1.280} & \num{0.738} \\
    Plain-(2,3) & \num{5.233e-4} & \num{2.117e-5} & 1.952 & 0.2294 & \num{28.19} & \num{18.16} \\
    \bottomrule
    \end{tabular}
\end{table}

\begin{figure}
 \centering
 \includegraphics[width=\textwidth]{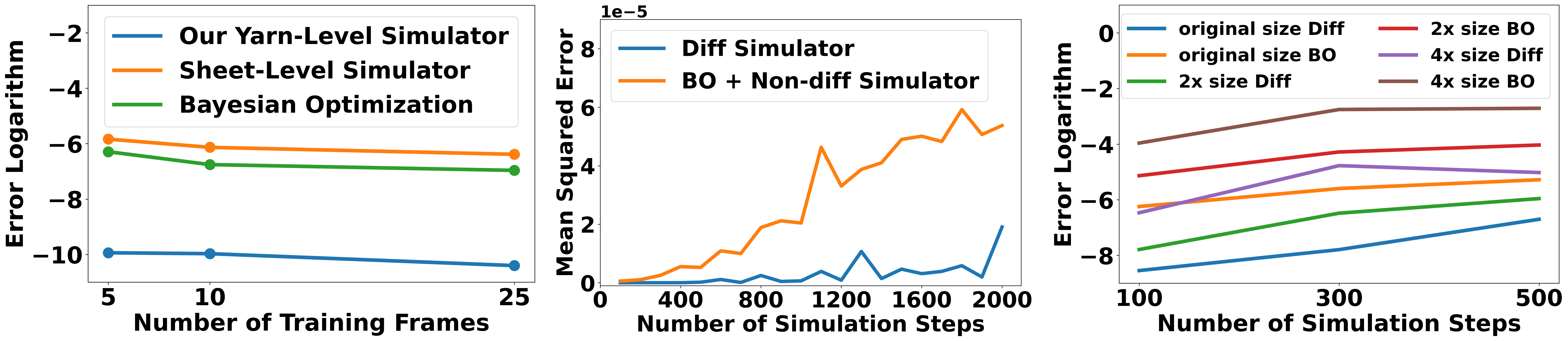}
 \vspace{-15pt}
 \caption{Simulation errors: data efficiency (left), long (middle) and big cloths simulation (right).}
 \vspace{-15pt}
 \label{fig:accumulatedErrors}
\end{figure}
 
\begin{wrapfigure}[17]{r}{0.35\textwidth}
    \centering
    \vspace{-15pt}
    \includegraphics[width=0.32\textwidth]{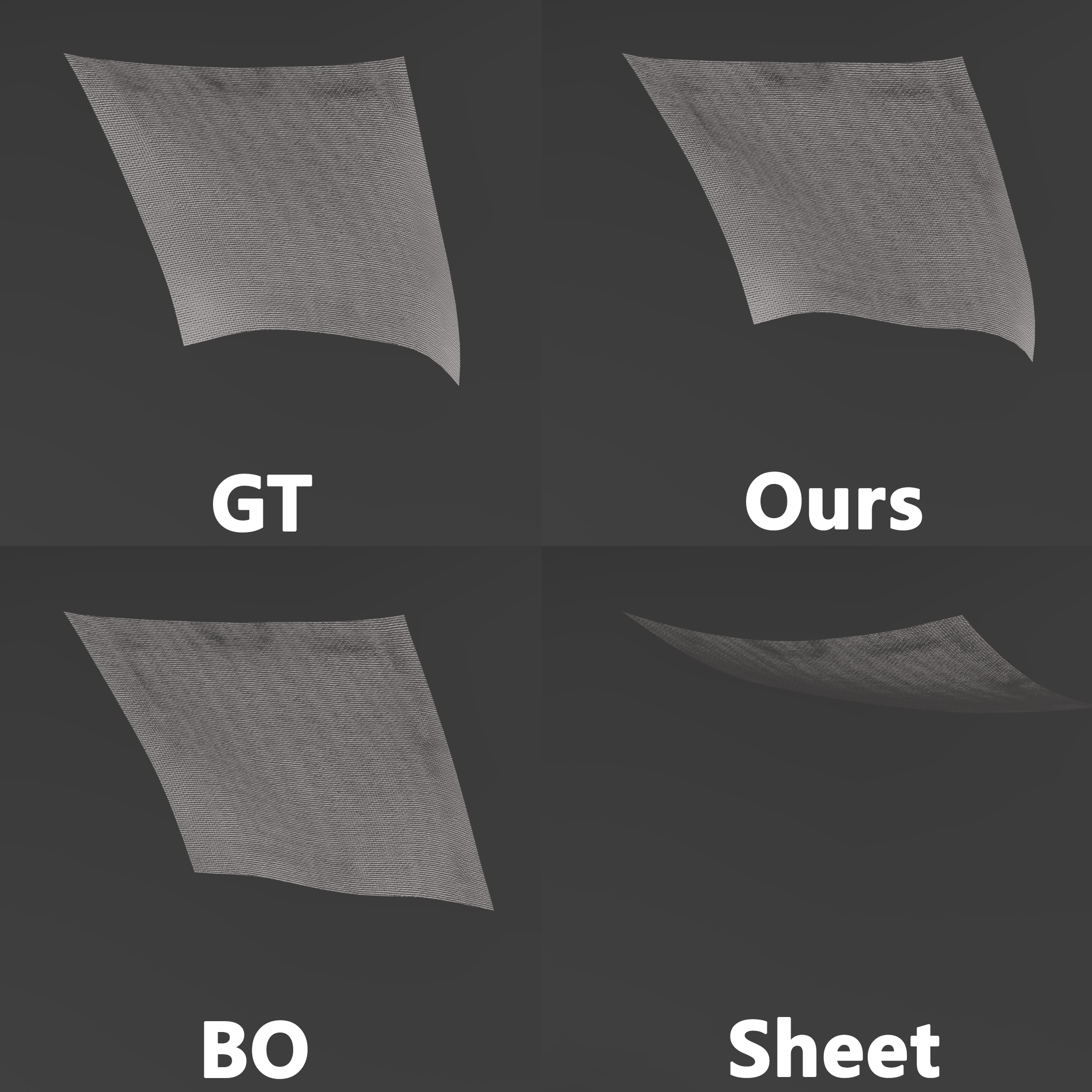}
    \caption{Simulation snapshots of the same step. The parameters estimated by our model is visually closest to the ground truth.}
    \label{fig:main_visual_com}
\end{wrapfigure}
From Table~\ref{tab:yarn_vs_sheet_vs_bo}, our model uses data more efficiently than \citep{liang_differentiable_2019} and BO. From training on 5 frames to 25 frames, our model reduces the error by as much as 96\% on Plain-(2, 3), while the largest improvements by the sheet-level model and BO optimization are 88\% on Plain-(2, 3) and 78\% on Plain-(1,2) respectively. Moreover, as shown in Figure \ref{fig:accumulatedErrors} left, our error on 5 frames is already several magnitudes smaller than the baselines. Further reducing it requires the model to be able to learn subtle dynamics very well. \citet{liang_differentiable_2019} essentially treats fabrics as a sheet. Since the simulation is from a yarn-level simulator~\citep{cirio_yarn_2014} which contains rich dynamics, the sheet model cannot precisely capture the subtle dynamics caused by individual yarns and their interactions. Further, the model granularity difference has more profound impact than just prediction. Being able to learn yarn parameters has immediate benefits for manufacturing and design, in terms of providing guidance on the choices of yarns and woven patterns. In addition, although BO can sometimes perform slightly better than the sheet model benefiting from a yarn-level simulator, its optimization process is not as efficient as ours. We only show results on Plain here and refer the readers to Appendix for Satin, Twill, and video comparisons. 


 \textbf{Error significance.} The MSE errors in Table~\ref{tab:yarn_vs_sheet_vs_bo} seem to be small, this is because the cloth is small and only simulated for a short period of time. But the results suggest errors in parameter estimation, which are amplified when the cloth is larger and simulated for a longer time. To demonstrate this, first, we run forward simulations for 2000 steps with parameters learned by our model and BO. Second, we show the compound influence using the parameters estimated by our model, \citet{liang_differentiable_2019}, and BO, and simulate a $17 \times 17$ cloth for 500 steps in the original size, 2 times size, and 4 times size. Fig.~\ref{fig:accumulatedErrors} middle-right show both results, which demonstrates the importance of  accurate parameter estimation. The errors of BO and \citet{liang_differentiable_2019} quickly become several times higher than our model when we scale the size and simulation time. We also show a visual comparison in Figure~\ref{fig:main_visual_com} and refer the readers to Appendix for more results.


\subsubsection{Control Learning}

We also show that our model can facilitate control learning. We design a task with a cloth placed on a table and aims to learn forces applied onto the four corners of the cloth to throw it into a box next to the table. The forces are only applied in the first 5 frames. We use our model to learn a sequence of forces which can throw the cloth into the box and compare it with a reinforcement learning baseline model: PPO \citep{schulman_2017_proximal}. In addition, we also present a variant of our model by appending two fully-connected layers after our model output (Ours + FC). We use the center of the box's bottom as the target location. When training our model, we use the $l_2$ distance between the cloth center of mass and the target position as the loss. When training PPO, we use the same $l_2$ distance for the reward.
\begin{wrapfigure}[14]{r}{0.32\textwidth}
    \centering
    \includegraphics[width=0.3\textwidth]{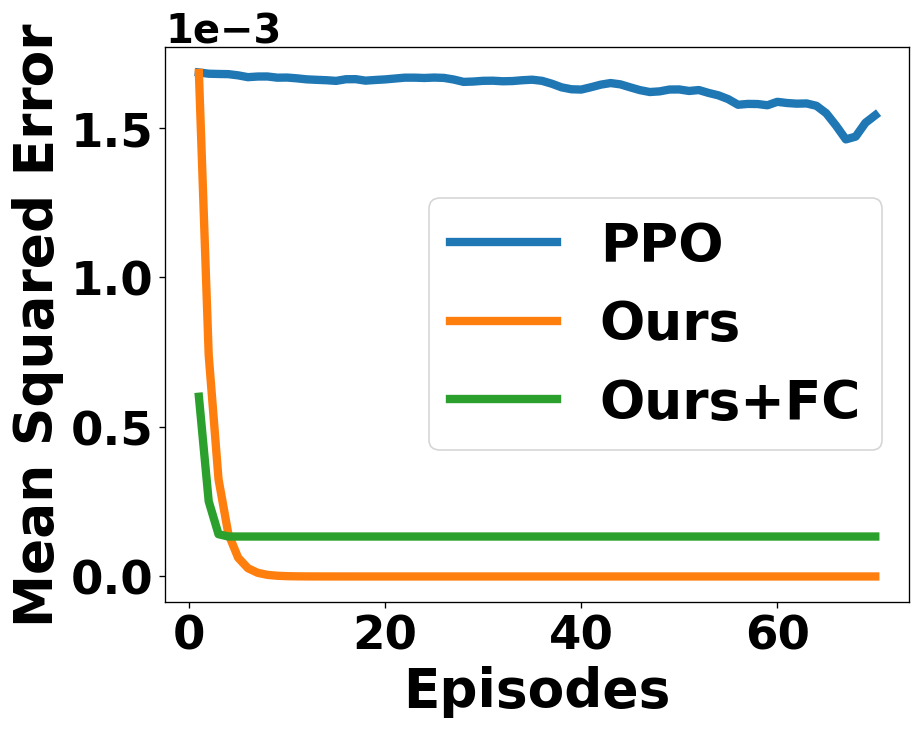}
    \caption{The MSE errors against epochs. Our approach learns faster than PPO.}
    \label{fig:control}
\end{wrapfigure}

The result shows both our model and Ours+FC can quickly learn the forces to throw the cloth into the box. By contrast, PPO is model-free and much slower because it needs to sample in a huge action space. By contrast, the full differentiability of our model enables a quicker search for effective control forces. More details can be found in Appendix.

In a broader context, there are also model-free methods~\citep{Yang_2017_learning,pfaff_learning_2020} which can also learn physics. The differences between our model and theirs are: 1. model explicability. The parameters that our model learns are interpretable and have physical meanings, so that it can guide manufacturing and design. 2. data efficiency. The data efficiency is much higher in our method. Our model can use as few as 5 frames for learning while model-free methods typically require hundreds to thousands.

\section{Discussion and Conclusion}
Our method is model-based, which requires domain knowledge and cannot simply `plug and play' on data as model-free methods~\citet{pfaff_learning_2020}. However, strong inductive biases from domain knowledge are necessary for differentiable physics to be applied in applications, because the model behaviour needs to be explainable in such applications, and cannot be merely black-box regression. Representative application domains include fabric manufacture/design and computer graphics, where both simulation and inverse problems need to be solved. 
Albeit focused on cloth, our model can be readily extended to general composite materials with mesh structures, e.g. from metal/plastic nets to buildings. In addition, our model can be embedded as a layer into a neural network, which helps learning control policies for cloth manipulation. Further, our model potentially enables a synergy between empirical physics modeling and deep learning, where our model can serve as a deterministic physics layer and other layers can incorporate non-linearity such as high-frequency dynamics in the system~\citep{shen_high_2008}. Finally, our modeling of general forces such as friction and shear contributes to differentiable physical modeling in a wider range, given the universal presence of such forces in the real world.

To our best knowledge, we proposed the first yarn-level differentiable fabric simulator, in the pursuit of fine-grained DPMs capable of incorporating domain knowledge. Through comprehensive evaluation, our model can effectively solve inverse problems, provide high data efficiency and facilitate control. We investigated differentiable modeling of common forces such as friction and shear, which provides a foundation for future attempts on fine-grained differentiable physics modeling. In future, we will pursue other composite materials such as metal meshes. Also, we will explore more complex dynamics such as buckling and permanent damages.

\bibliography{iclr2022_conference}
\bibliographystyle{iclr2022_conference}

\appendix
\section{Appendix}
All simulations are available in the accompanied videos:\url{ https://youtu.be/pCB8AD9R4Dk}
\subsection{Training Details}

\begin{figure}[ht]
    \centering
    \includegraphics[width=\textwidth]{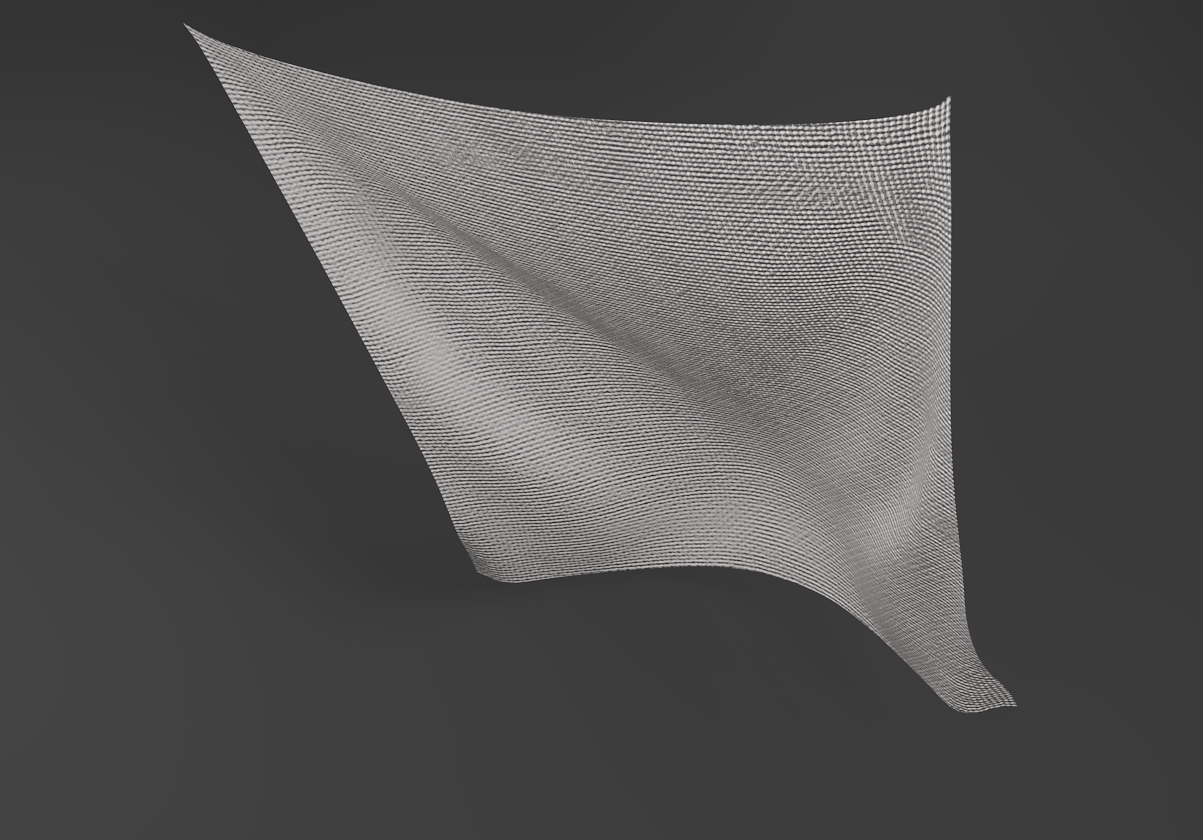}
    \caption{A piece of square cloth blown by constant magnitude wind.}
    \label{fig:blownCloth}
\end{figure}

Our ground-truth data is simulated with a piece of cloth hanging at its two corners, blown by a wind with a constant magnitude (Figure~\ref{fig:blownCloth}). 
The simulation is conducted with a time step h = 0.001. In all experiments, we use Stochastic Gradient Descent and run 70 epochs for training, except in XXX-(1,3) where we trained our model for 90 epochs. The training is conducted on a machine with Intel(R) Xeon(R) Silver 4216 CPU, 187G memory, NVIDIA TITAN RTX graphics card on Linux. The main factors of training speed are the cloth size and the training data size. In our experiments, the training takes approximately 68, 133, and 328 seconds per epoch on a $17 \times 17$ cloth with training data containing 5, 10, and 25 frames respectively.  The training per epoch takes approximately 13, 106, 328, and 1310 seconds with 25 training frames, on a $5 \times 5$, $10 \times 10$, $17\times17$, and $25\times25$ cloth respectively.

\begin{figure}
    \centering
    \includegraphics[width=\textwidth]{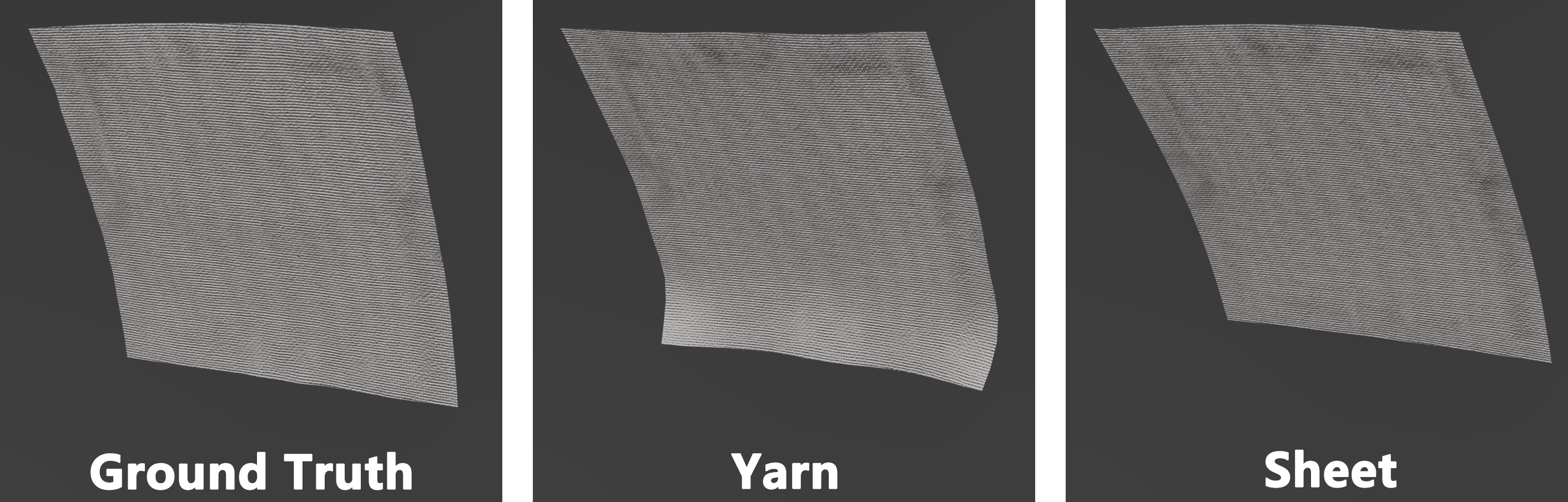}
    \caption{The visual learning results of the differentiable sheet-level simulator~\citep{liang_differentiable_2019} and our model learns on the data generate by \citep{narain_2012_adaptive}}
    \label{fig:visual_train_sheet}
\end{figure}

\textbf{Additional experiments.} Further, we also conduct comparisons on the data simulated under the same settings by a sheet-level simulator~\citep{narain_2012_adaptive}, which tends to be stiffer. This is to compare the performance when the ground-truth does not contain the same level of subtle dynamics. Since there is no Eulerian coordinates in the sheet-level simulation, we only use Lagragian coordinates in the loss function. The visual comparison is in Figure~\ref{fig:visual_train_sheet} and the prediction errors are shown in Table~\ref{tab:app_sheet_vs_yarn}. Our model can learn comparable results on 5 frames, and better results on 10 and 25 frames. The slightly worse 5-frame result is mainly because the first 5 frames contain small dynamics and therefore is insufficient for our model to learn the overall stiffness of the cloth. However, when 10 and 25 frames are given, the learning is significantly improved and even outperforms \citet{liang_differentiable_2019}. Also, since there is no woven pattern information in the ground truth, we examine our model across the three woven patterns, all giving more accurate predictions. Overall, the comparisons show our model has higher prediction accuracy regardless the granularity of the underlying physics model.

\textbf{Parameters.} We induce prior knowledge to limit the parameter learning within valid ranges, so that the multi-solution problem, also met by existing methods, can be mitigated. All cloths we used are made of two types of yarns. We use the same range, $d\in[0.001, 0.003]$, $b\in[0.00005, 0.00018]$, $S\in[0, 1200]$ and $\mu\in[0, 1.0]$ for both yarns, where $d$, $b$, $S$ and $\mu$ are the density, bending modulus, shear modulus and friction coefficient respectively. We use $s_1\in[0, 800000]$ and $s_2\in[0, 300000]$ for the stretching for both yarns. For other coefficients, we use $k_f=1000$ and $d_f=1000$ in the friction force, $c=3$ and $\sigma=0.6$ in the shear force, $k_c=1$ in yarn-to-yarn collision in all experiments.

When training our model on the data generated by a sheet-level cloth simulator~\citep{narain_2012_adaptive}, we use a pure woven cloth made of one type of yarn. This is because it is not possible to specify multiple yarn behaviors in a sheet simulator, so we use a pure yarn cloth for generating the ground truth. The cloth parameters are from the `white-dots-on-black' cloth in \citet{wang_2011_data} which is 100 percent polyester. To learn from it, we employ all three woven patterns in our model as there is no prior knowledge about the woven pattern of the `white-dots-on-black' cloth. We also fix the friction coefficient $\mu=0.5$ and impose the ranges on parameters shown in Table~\ref{tab:train_by_sheet}. Finally, we would like to point it out in real-world applications, information such as woven patterns and yarn materials are easily available so that the ranges of parameter values such as density, bending and stretching can be obtained. Although the knowledge of shearing and friction cannot be easily acquired, the ranges we use are general enough.

\begin{table}[tb]
    \centering
    \caption{Testing errors ($\times10^{-6}$) of our model and \citep{liang_differentiable_2019} trained on 5, 10 and 25 frames generated by~\citep{narain_2012_adaptive}.}
    \label{tab:app_sheet_vs_yarn}
    \vspace{1ex}
    \begin{tabular}{cccc}
    \toprule
    fabrics/frames & 5 & 10 & 25 \\
    \midrule
    Plain-(1,2) & \num{6.702} & \num{1.167} & \num{0.496} \\
    Satin-(1,2) & \num{7.972} & \num{1.225} & \num{0.624} \\
    Twill-(1,2) & \num{8.218} & \num{1.772} & \num{0.776} \\
    \citep{liang_differentiable_2019}&\num{4.098} & \num{4.752} & \num{1.716} \\
    \bottomrule
    \end{tabular}
\end{table}


Note that in all experiments, the prior knowledge we induce is only a weak prior, i.e. using the same general ranges for multiple experiments across different woven patterns, so that the learning success still lies in our model's ability to infer the right parameter values.

\paragraph{Parameter Initialization.} The material estimation results are affected by initialization. To test if our model can learn stably, we report the mean and the standard deviation of multiple experiments with different parameter initial values. The initial values of the physical parameters are randomly selected from a range of $\pm10 \%$ of the average of the two yarns. For instance, in learning the stretch in Plain-(1,2), we only know the ranges of the stretching parameters Y1 and Y2 of Yarn1 and Yarn2 but not the exact values. Therefore, when initializing Y1 and Y2, we randomly sample values from a range of $\pm10 \%$ of the mean stretch stiffness of the Yarn1 and Yarn2, $[\mbox{mean(Y1, Y2)} \times 0.9, \mbox{mean(Y1, Y2)} \times 1.1]$ for initialization. The results of the 5 repetitions are shown Table \ref{tab:rand_ini_1} and Table \ref{tab:rand_ini_2}. Given that the standard deviations are small, it shows that our model can stably learn reasonable parameter values.

\paragraph{Different Force Magnitude.} To evaluate the influence of the wind force, we conduct experiments using 5N, 10N, and 15N wind force to blow a piece of $17 \times 17$ Plain-(1,2) cloth. The learning result is shown in the Table \ref{tab:diff_wind} which demonstrate wind force strength has ignorable influence on the learned parameters.

\begin{table}[tb]
    \centering
    \caption{Learning cloth parameters with different initial values (part one).}
    \begin{tabular}{ccc}
    \toprule
        Size & Shear $S$ & Friction $\mu$ \\
        \midrule
         $5\times5$ & $1011.79\pm6.12$ & $0.39\pm0.08$\\
         $10\times10$ & $983.41\pm6.84$ & $0.44\pm0.03$\\
         $17\times17$ & $962.29\pm8.99$ & $0.47\pm0.06$\\
         \bottomrule
    \end{tabular}
    \label{tab:rand_ini_1}
\end{table}

\begin{table}[tb]
    \centering
    \caption{Learning cloth parameters with different initial values (part two).}
    \begin{tabular}{ccccc}
    \toprule
        Size & Yarn & Density & Stretch & Bend \\
        \midrule
         \multirow{2}*{$5\times5$}
         & 1 & \num{1.98e-3}$\pm$\num{3.00e-5} & $498595\pm8862$ & \num{1.37e-4}$\pm$\num{1.41e-6}\\
         &  2 & \num{2.45e-3}$\pm$\num{4.81e-5} & $186710\pm3776$ & \num{1.11e-4}$\pm$\num{4.78e-6}\\
         \multirow{2}*{$10\times10$}
         & 1 & \num{2.03e-3}$\pm$\num{5.04e-5} & $542375\pm7099$ & \num{1.44e-4}$\pm$\num{2.08e-6}\\
         & 2 & \num{2.47e-3}$\pm$\num{4.73e-5} & $180032\pm1848$ & \num{1.05e-4}$\pm$\num{8.18e-6}\\
          \multirow{2}*{$17\times17$}
         & 1 & \num{2.00e-3}$\pm$\num{6.66e-5} & $519993\pm3175$ & \num{1.43e-4}$\pm$\num{5.55e-6}\\
         & 2 & \num{2.45e-3}$\pm$\num{5.04e-5} & $176232\pm1514$ & \num{1.19e-4}$\pm$\num{6.50e-6}\\
         \bottomrule
    \end{tabular}
    \label{tab:rand_ini_2}
\end{table}

\begin{table}[H]
    \centering
    \caption{Learning cloth physical parameters with different wind force.}
    \begin{tabular}{ccccccc}
    \toprule
        Wind & Shear $S$ & Friction $\mu$ & Yarn & Density & Stretch & Bend \\
        \midrule
         \multirow{2}*{5} & \multirow{2}*{\num{947}} & \multirow{2}*{0.402}
         & 1 & \num{1.969e-3} & 505421 & \num{1.323e-4}\\
         & & & 2 & \num{2.440e-3} & 171304 & \num{1.034e-4}\\
         \multirow{2}*{10} & \multirow{2}*{\num{942}} & \multirow{2}*{0.520}
         & 1 & \num{2.026e-3} & 494109 & \num{1.311e-4}\\
         & & & 2 & \num{2.441e-3} & 168267 & \num{1.049e-4}\\
         \multirow{2}*{15} & \multirow{2}*{934} & \multirow{2}*{0.586}
         & 1 & \num{2.029e-3} & 487918 & \num{1.341e-4}\\
         & & & 2 & \num{2.437e-3} & 167601 & \num{1.066e-4}\\
         \bottomrule
    \end{tabular}
    \label{tab:diff_wind}
\end{table}

\begin{figure}[bt]
    \centering
    \includegraphics[width=\textwidth]{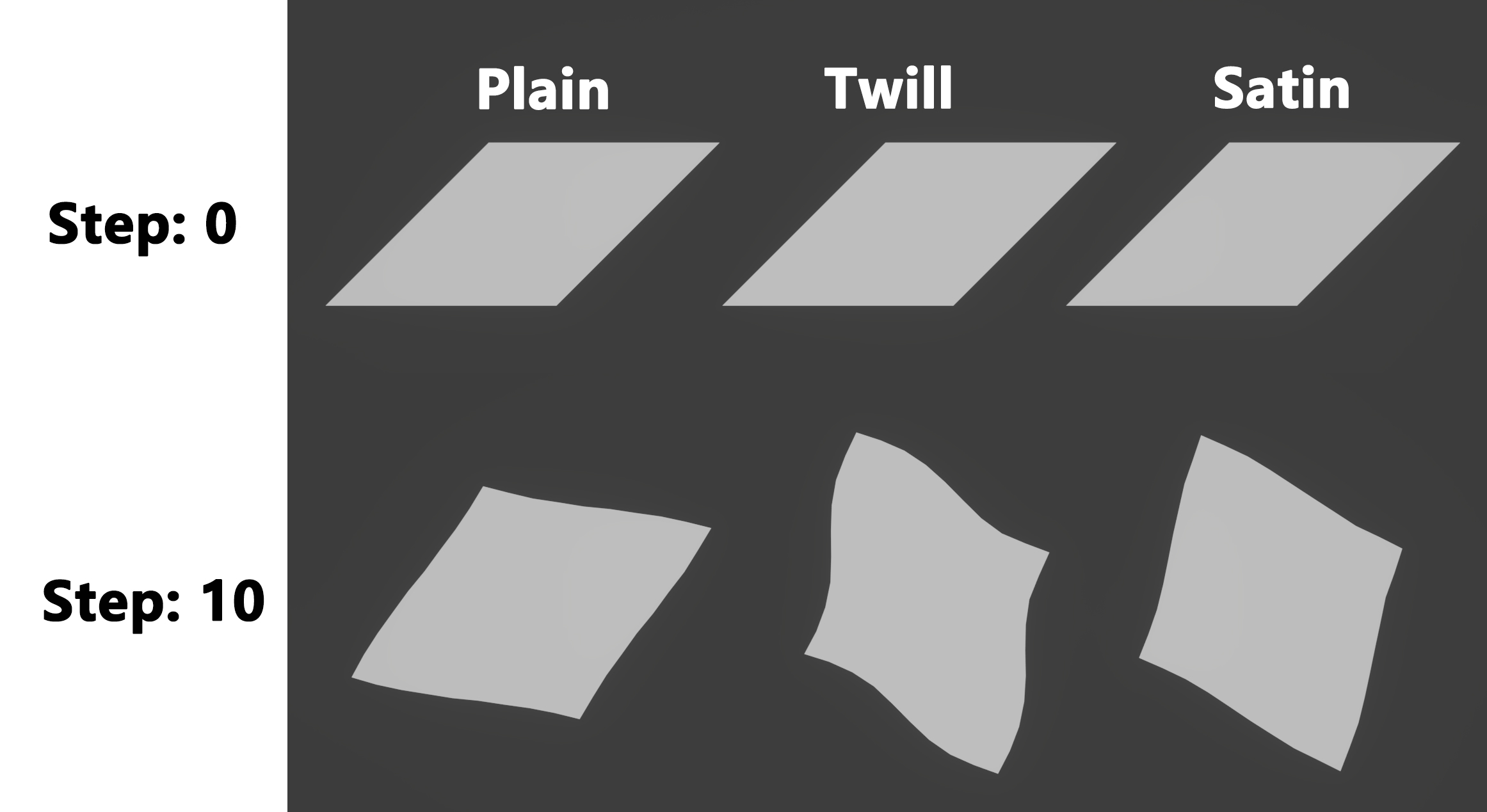}
    \caption{Three pieces of cloth woven in different patterns show different dynamics.}
    \label{fig:dyn_woven_pattern}
\end{figure}

\paragraph{Influence of Woven Patterns.} The investigation on different woven patterns is crucial as they affect the cloth dynamics significantly. To show this, we conducted simulations of three pieces of cloths with the same parameters, but with different woven patterns. We shear three pieces of cloth then release them. The Figure \ref{fig:dyn_woven_pattern} shows three pieces of cloth in the initial state and 10 steps later. There are obvious differences after merely 10 steps. This demonstrates woven patterns have considerable influences on the overall mechanical properties.

\begin{table}[bt]
    \centering
    \caption{Cloth parameters' initial values and ranges when ground-truth generated by sheet-level cloth simulator\citep{narain_2012_adaptive}}
    \label{tab:train_by_sheet}
    \begin{tabular}{ccccc}
    \toprule
        Name &  Density($kg/m$) & Stretch($N/m$) & Bend($N/m$) & Shear($N/m$) \\
        \midrule
        Value &  0.004 & 1e6 & 0.0001 & 20000\\
        Upper limit & 0.008 & 2e6 & 0.0002 & 30000 \\
        Lower limit & 0.001 & 0 & 0 & 0\\
        \bottomrule
    \end{tabular}
\end{table}



\begin{table}[bt]
    \caption{Testing errors ($\times10^{-6}$) of our model (left) and \citep{liang_differentiable_2019} (right) trained on 5, 10 and 25 frames. Ground-truth generated by a yarn-level simulator~\citep{cirio_yarn_2014}.}
    \label{tab:yarn_vs_sheet}
    \centering
    \begin{tabular}{cccc|ccc}
    \toprule
    fabrics/frames & 5 & 10 & 25 & 5 & 10 & 25 \\
    \midrule
    Plain-(1,2) & \num{1.152e-4} & \num{1.068e-4} & \num{3.962e-5} & \num{1.462} & 0.7375 & 0.4124 \\
    Plain-(1,3) & \num{1.516e-4} & \num{1.268e-4} & \num{3.555e-5} & \num{1.608} & 0.7906 & 0.4567 \\
    Plain-(2,3) & \num{5.233e-4} & \num{1.291e-4} & \num{2.117e-5} & \num{1.952} & 0.5999 & 0.2294 \\
    \hline
    Satin-(1,2) & \num{1.134e-4} & \num{1.070e-4} & \num{4.285e-5} & \num{1.466} & 0.7405 & 0.4146 \\
    Satin-(1,3) & \num{1.551e-4} & \num{1.355e-4} & \num{4.362e-5} & \num{1.624} & 0.8004 & 0.4445 \\
    Satin-(2,3) & \num{6.254e-4} & \num{1.355e-4} & \num{4.413e-5} & \num{2.128} & 0.5949 & 0.2265 \\
    \hline
    Twill-(1,2) & \num{1.130e-4} & \num{1.068e-4} & \num{4.208e-5} & \num{1.472} & 0.7451 & 0.4160 \\
    Twill-(1,3) & \num{1.550e-4} & \num{1.349e-4} & \num{4.200e-5} & \num{1.633} & 0.8059 & 0.4577 \\
    Twill-(2,3) & \num{6.470e-4} & \num{1.352e-4} & \num{4.938e-5} & \num{2.181} & 0.5994 & 0.2278 \\
    \bottomrule
    \end{tabular}
\end{table}

\subsection{Visual results}

\begin{figure}[bt]
    \centering
    \includegraphics[width=\textwidth]{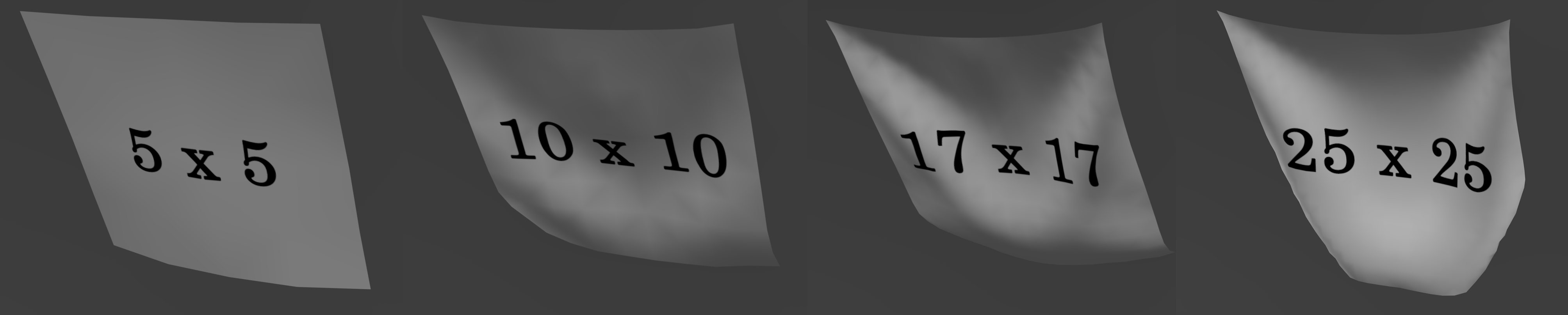}
    \caption{The visual results of our model learning on different cloth sizes. From left to right: $5\times5$, $10\times10$, $17\times17$ and $25\times25$.}
    \label{fig:multiresolution}
\end{figure}

Here we show some snapshots of our model on cloths of different sizes in Figure~\ref{fig:multiresolution}.  As expected, small cloths tend to show low dynamics and appear to be more `rigid'. Bigger cloths tend to have more subtle dynamics such as wrinkles, even under the same external impact, i.e. gravity and wind with a constant magnitude. More visual results can be found in the supplementary video.

\begin{figure}
    \centering
    \includegraphics[width=\textwidth]{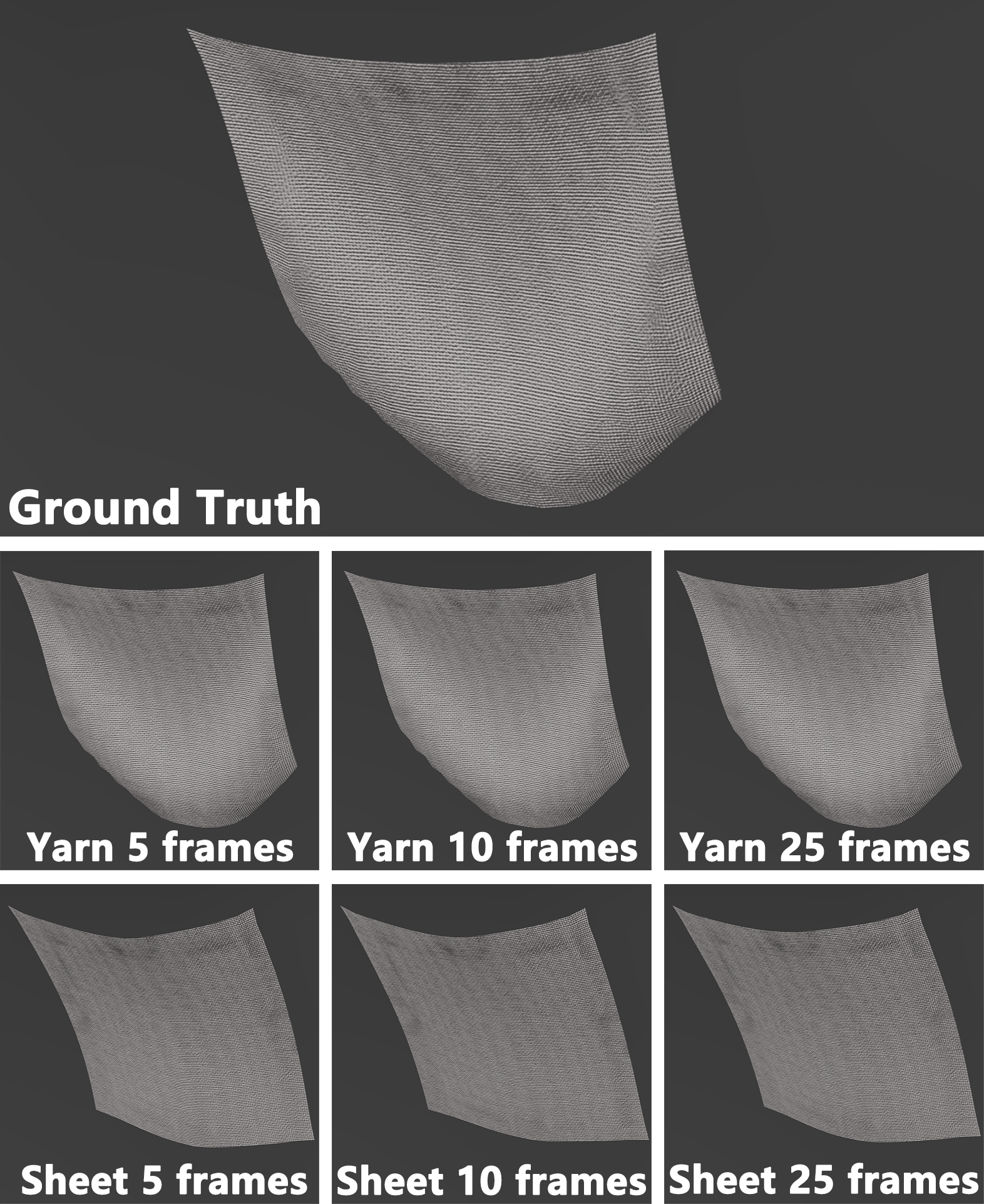}
    \caption{The visual results of Plain-(1, 2) ground-truth, our model, and sheet-level model trained with different number of frames. The snapshots are the 133$th$ frame of the simulations after learning.}
    \label{fig:comparison}
\end{figure}

\begin{figure}[tb]
    \centering
    \includegraphics[width=\textwidth]{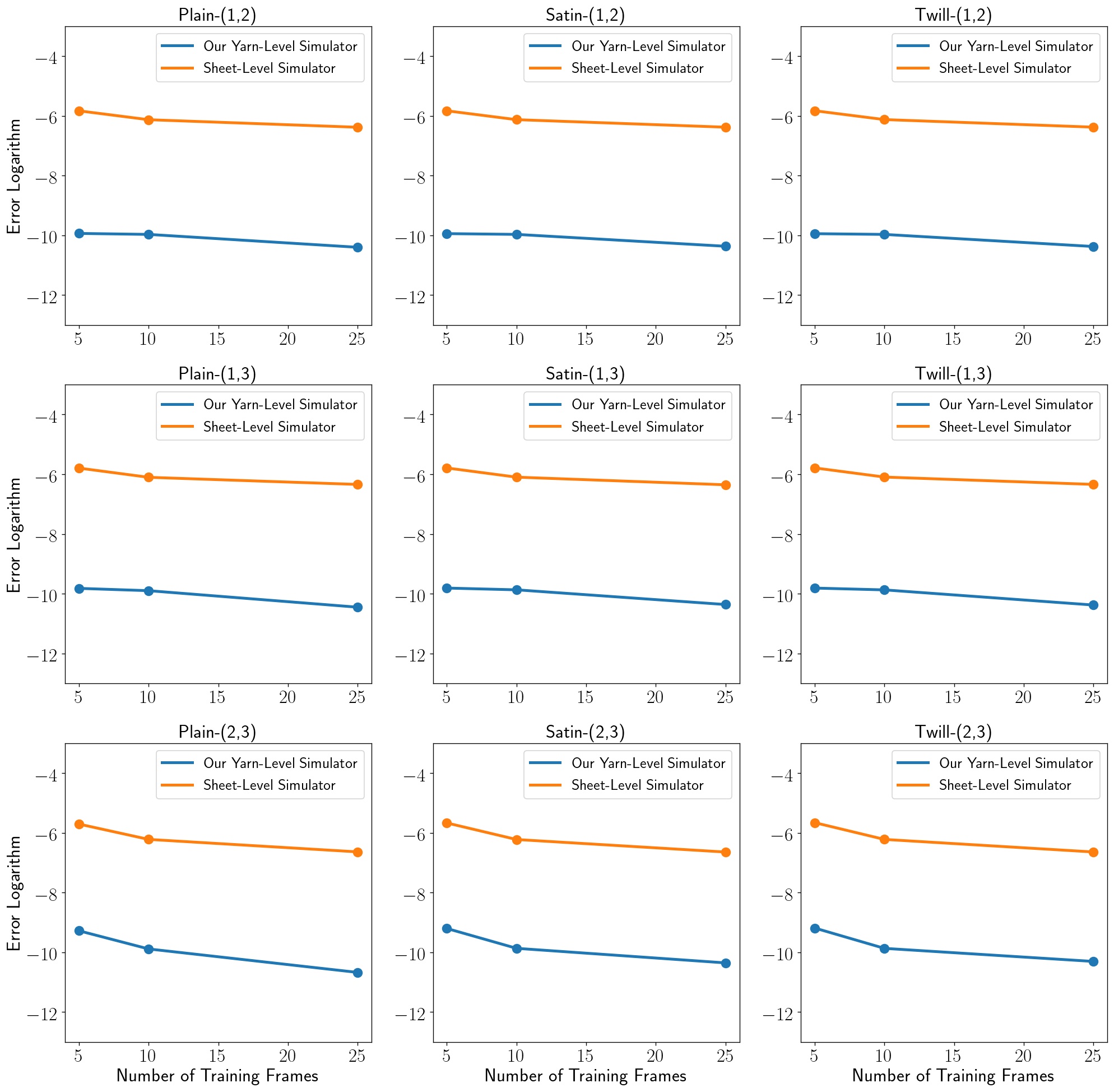}
    \caption{Prediction error logarithm vs training data.}
    \label{fig:9Error}
\end{figure}

\begin{table}[tb]
    \centering
    \caption{Learned parameters by Bayesian Optimization on different kinds of fabrics.}
    \begin{tabular}{ccccccc}
    \toprule
    Frames & Density & Stretch & Bend & Density & Stretch & Bend \\
    \midrule
    5  & \num{2.483e-3} & 647270 & \num{0.636e-4} & \num{2.125e-3} & 270641 & \num{1.576e-4} \\
    10 & \num{2.176e-3} & 577235 & \num{0.798e-4} & \num{2.264e-3} & 217144 & \num{1.542e-4} \\
    25 & \num{2.328e-3} & 537434 & \num{1.687e-4} & \num{2.097e-3} & 249896 & \num{0.976e-4} \\
    \midrule
    5  & \num{2.202e-3} & 605289 & \num{1.403e-4} & \num{2.349e-3} & 272153 & \num{0.868e-4} \\
    10 & \num{1.669e-3} & 257877 & \num{1.582e-4} & \num{2.635e-3} & 268451 & \num{0.529e-4} \\
    25 & \num{1.454e-3} & 315715 & \num{1.213e-4} & \num{2.950e-3} & 23702  & \num{1.656e-4} \\
    \midrule
    5  & \num{2.514e-3} & 250093 & \num{1.611e-4} & \num{2.363e-3} & 20371  & \num{0.985e-4} \\
    10 & \num{2.964e-3} & 164021 & \num{0.524e-4} & \num{2.255e-3} & 49648  & \num{1.225e-4} \\
    25 & \num{2.414e-3} & 73734  & \num{0.890e-4} & \num{2.436e-3} & 267452 & \num{1.113e-4} \\
    \bottomrule
    \end{tabular}
    \label{tab:bo_learned_parameteres}
\end{table}

\subsection{Yarn-level versus Sheet-level}
A full comparison between our model and \citep{liang_differentiable_2019} is shown in Table~\ref{tab:yarn_vs_sheet}, where a yarn-level simulator~\citep{cirio_2016_yarn} is used to generate the ground-truth. We exhaustively conduct comparisons using all combinations of yarns and woven patterns. We can see that our model is consistently better than \citep{liang_differentiable_2019} by large margins. Visually, we show snapshots in Figure~\ref{fig:comparison}. The sheet model results are in general more rigid and do not contain as much subtle dynamics as ours do, across different training frame numbers. Since 5, 10 and 25 frames contain different amounts of information on (subtle) motion dynamics, Figure~\ref{fig:comparison} shows that there is a lack of granularity in the sheet model when capturing subtle dynamics compared with ours.

Further, we also show the plots on the data efficiency in Figure~\ref{fig:9Error}, under all 9 yarn-woven pattern combinations, across different amounts of training data. In all settings, our data efficiency is significantly higher. By extrapolation, it would take a large number of extra training frames for the sheet-level model to achieve similar accuracy. More comparisons are also available in the supplementary video.

\subsection{Our model versus Bayesian Optimization}
Table \ref{tab:BO_error} shows the testing errors of the Bayesian Optimization. Although the MSE errors are small, the learned parameters are far from the ground truth (shown in the Table \ref{tab:bo_learned_parameteres}), which is somewhat surprising. After examining the results, we find that Bayesian Optimization suffers from the multi-solution problem so that it merely gives a set of working parameters instead of the true parameters. In other words, although the prediction error is low, physically speaking, the learned parameters are far from the true materials. This happens even when we use the same parameter ranges as in our model. This is an intrinsic property of Bayesian optimization which is based on sampling, and therefore difficult to avoid during learning.

\begin{table}[H]
    \caption{Testing error ($\times10^{-6}$) of Bayesian Optimization with yarn-level simulator \citep{cirio_2016_yarn} learned on 5, 10, and 25 frames.}
    \centering
    \begin{tabular}{cccc}
    \toprule
    Fabrics/Frames & 5 & 10 & 25  \\
    \midrule
    Plain-(1,2) & \num{0.512} & \num{0.176} & \num{0.109} \\
    Plain-(1,3) & \num{1.280} & \num{1.269} & \num{0.738} \\
    Plain-(2,3) & \num{28.19} & \num{19.22} & \num{18.16} \\
    \bottomrule
    \end{tabular}
    \label{tab:BO_error}
\end{table}

\subsection{Control Experiment Setting}
\begin{figure}[H]
    \centering
    \includegraphics[width=\textwidth]{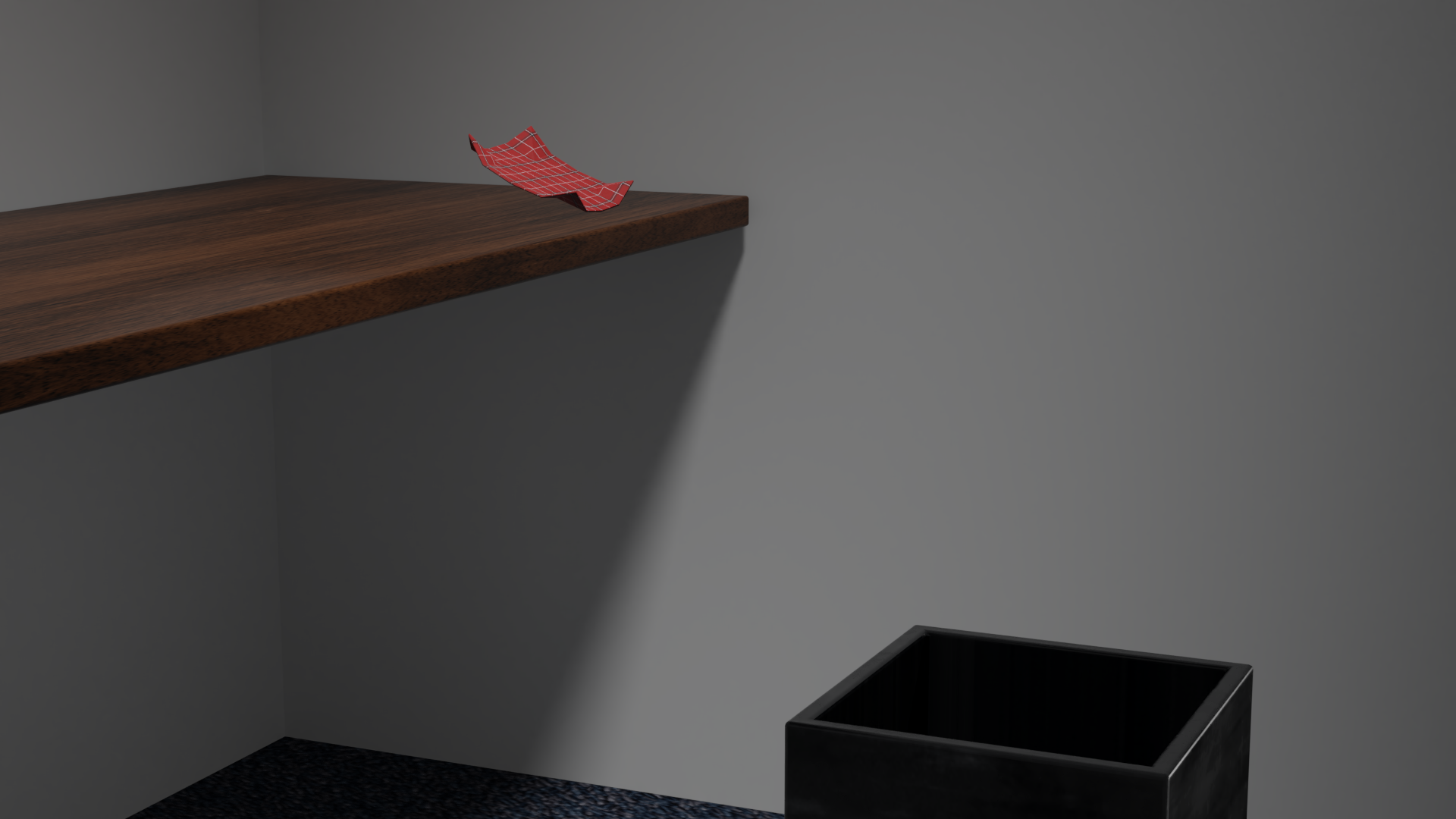}
    \caption{A square cloth is thrown from the table into the black box by four forces applied on the four corners of the cloth.}
    \label{fig:control_scene}
\end{figure}

The control experiment scenario is illustrated in the Figure \ref{fig:control_scene}. 

\subsection{Significant Error in Visual}
\begin{figure}[tb]
    \centering
    \includegraphics[width=0.9\textwidth]{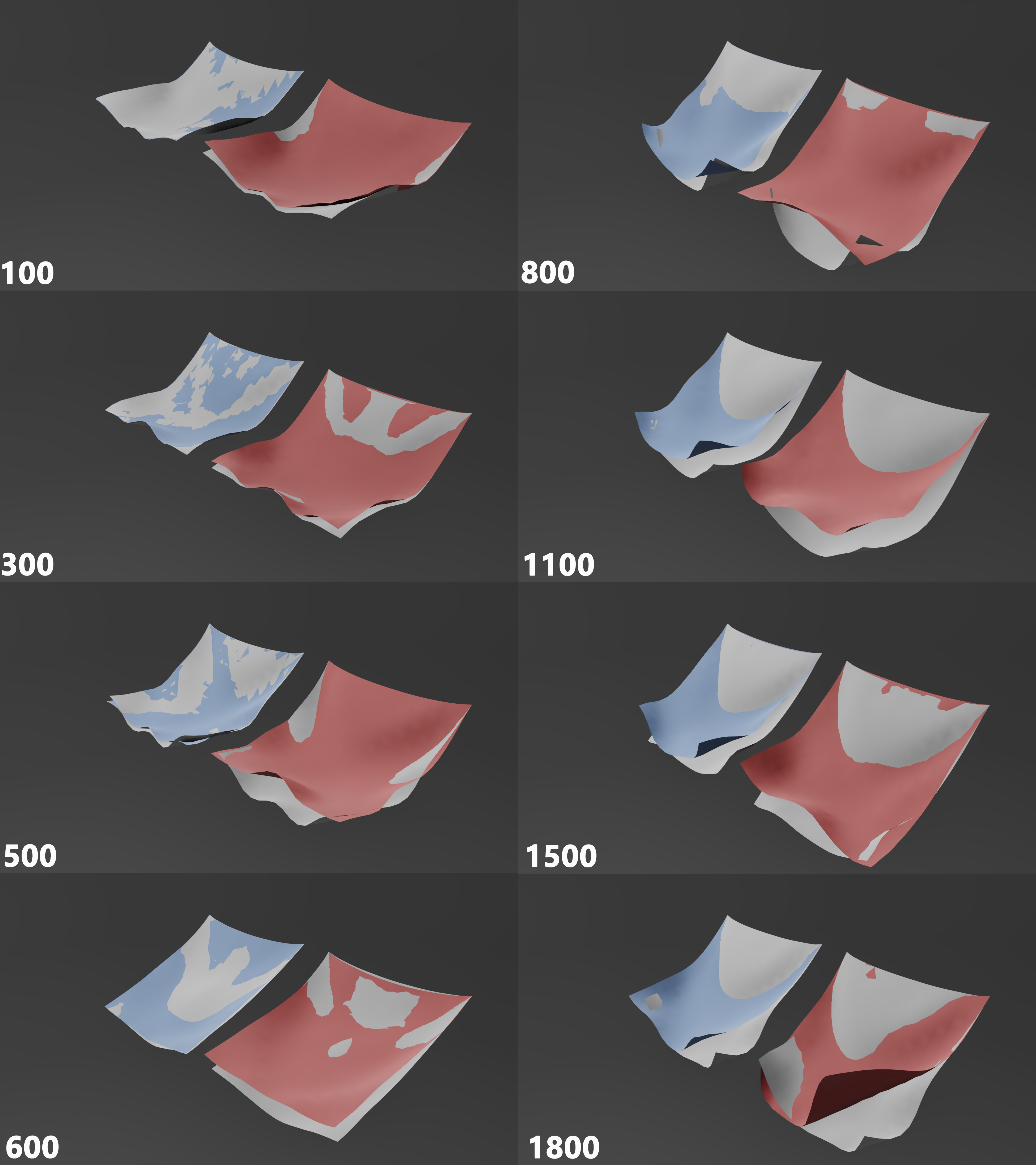}
    \caption{Visual differences in long simulations. The grey cloth is ground truth. The blue cloth and the red cloth are simulated with the parameters learned by our model and BO. The blue cloth shows smaller visual differences than the red one.}
    \label{fig:error_time}
\end{figure}

\begin{figure}[tb]
    \centering
    \includegraphics[width=0.8\textwidth]{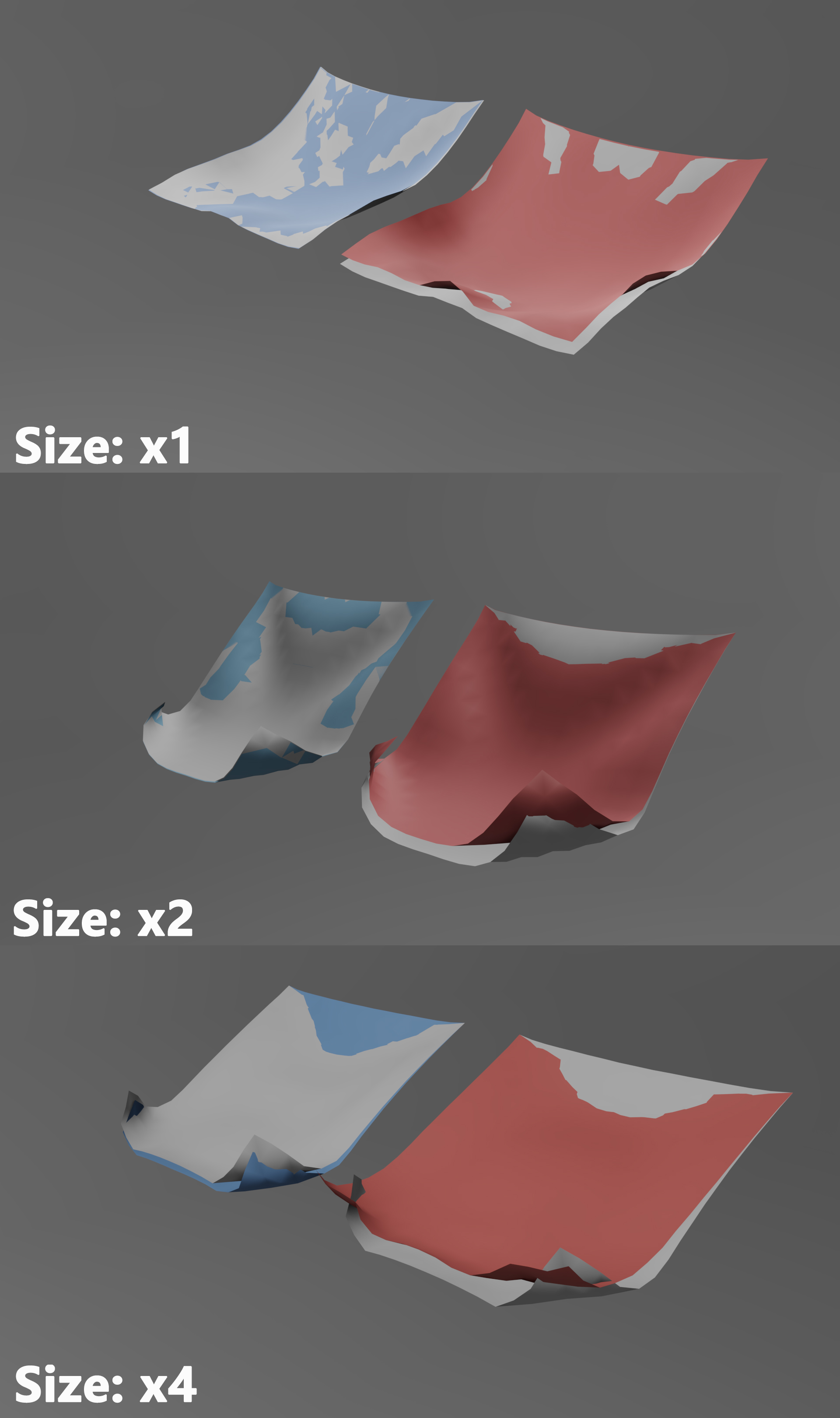}
    \caption{Visual differences on larger cloths and long simulation (500 steps). The grey cloth is ground truth. The blue cloth and the red cloth are simulated with the parameters learned by our model and BO. The blue cloth shows smaller differences than the red one.}
    \label{fig:error_space}
\end{figure}

We discussed the significance of the small error in physics-based simulation. Figure \ref{fig:error_time} and Figure \ref{fig:error_space} visually prove our explanations in the main paper: the error accumulates over time and increases with increasing cloth size.

\newpage

\newpage

\section{Differentiable Yarn-level Cloth Simulator}
In this section, we give the full details of our model and mathematical derivation. 
\subsection{Intro yarn force models}
\label{sec:app_representation}

\begin{wrapfigure}[12]{r}{0.4\textwidth}
    \centering
    \includegraphics[trim=0 0 0 110, width=0.3\textwidth]{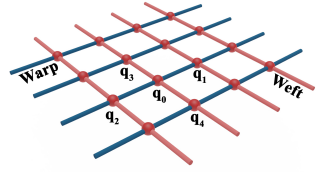}
    \caption{Blue and red rods denote warps and wefts respectively. $\mathbf{q}$s are the crossing nodes.}
    \label{fig:app_grid}
\end{wrapfigure}

Representing the cloth as in Figure~\ref{fig:app_grid}, we employ an EoL discretization~\citep{sueda_large_2011} and denote the spatial positions of crossing nodes in Lagrangian coordinates and represent the contact sliding movement in Eulerian coordinates, $\mathbf{q}_i \equiv (\mathbf{x}_i,u_i,v_i)$ where $\mathbf{x}_i\in \mathbb{R}^3$ implies crossing node $i$’s spatial position and $(u_i,v_i)$ the node’s position in the material frame. The two end points of yarns are taken as special crossing nodes as they do not contact with other yarns and therefore have no Eulerian terms, i.e. $\mathbf{q}_j \equiv \mathbf{x}_i$. Therefore, on a $r (rows) \times c (columns)$ cloth, there are $ (r-2) \times (c-2)$ crossing nodes with five Degrees of Freedom (DoFs) and $ 2r + 2c - 4$ crossing nodes with three DoFs. Every two neighboring crossing nodes on the same warp/weft delimit a warp/weft segment. A warp segment whose two end points are $\mathbf{q}_0$ and $\mathbf{q}_1$ is denoted as $\mathbf{[q_0,q_1]}$ and its position is $(\mathbf{x}_0,\mathbf{x}_1,u_0, u_1)$ (shown in \ref{fig:app_grid}). This way, a woven cloth is discretized into crossing nodes and segments which are the primitive units of the simulated cloth. Every segment is assumed to be straight so that linear interpolation can be employed on the segment, e.g. the
spatial position of a point in the segment $\mathbf{[q_0,q_1]}$ is $\mathbf{x}(u) = \frac{u - u_0}{\Delta u} \mathbf{x}_0 + \frac{u_1 - u}{\Delta u} \mathbf{x}_1$, where $u$ is the point's position in Eulerian coordinates and $\Delta u = u_1 - u_0$ is the length of the segment. We use $L$ to denote the distances between neighbor yarns and $R$ to denote the yarn radius. 

\subsection{System Equation for Simulation}
\label{sec:app_systemEquation}
A cloth’s state at time $t$, $\mathcal{S}_{(t)}=\{\mathcal{Q}_{(t)},\dot{\mathcal{Q}}_{(t)}\}$, includes all its crossing nodes’ positions $\mathcal{Q} = \{\mathbf{q}_i |i=1,2,\dots,N\}$ and velocities $\dot{\mathcal{Q}}=\{\dot{\mathbf{q}}_i | i=1,2, \dots,N\}$, where $N$ is the number of crossing nodes. Knowing the states, then we can calculate the internal and external forces:
\begin{equation}
\label{eqn:app_Lag}
    \mathbf{F} = \mathbf{M}\ddot{\mathbf{q}} = \frac{\partial T}{\partial \mathbf{q}} - \frac{\partial V}{\partial \mathbf{q}} - \dot{\mathbf{M}}\dot{\mathbf{q}}
\end{equation}
where $\mathbf{q}$, $\dot{\mathbf{q}}$, and $\ddot{\mathbf{q}}$ are the nodes general position, velocity, and acceleration respectively, with a dimension $l = 3 \times r \times c + 2 \times (r-2) \times (c-2)$. $\mathbf{M} \in \mathbb{R}^{l \times l}$ is the general mass matrix. The model assumes mass is distributed homogeneously in one segment, so the mass matrix of a warp segment $[\mathbf{q}_0,\mathbf{q}_1 ]$ is 
\begin{equation}
    \label{eqn:segment_mass}
    \mathbf{M}_{0,1} 
    = \frac{1}{6}\Delta u \rho
    \begin{pmatrix}
    2\mathbf{I}_3 & \mathbf{I}_3 & -2\mathbf{w} & -\mathbf{w} \\
    \mathbf{I}_3 & 2\mathbf{I}_3 & -\mathbf{w} & -2\mathbf{w} \\
    -2\mathbf{w}^{\top} & -\mathbf{w}^{\top} & 2\mathbf{w}^{\top}\mathbf{w} & \mathbf{w}^{\top}\mathbf{w} \\
    -\mathbf{w}^{\top} & -2\mathbf{w}^{\top} & \mathbf{w}^{\top}\mathbf{w} & 2\mathbf{w^{\top}w}\\
    \end{pmatrix}
\end{equation}
where $\mathbf{w} = \frac{\mathbf{x}_1 - \mathbf{x}_0}{\Delta u}$, and $\rho$ is yarn density. 
$T$ and $V$ are the kinetic and potential energy respectively. As the partial derivative of energy with respect to position is force, the right hand terms in Equation \ref{eqn:app_Lag} are inertia, conservative forces, and part of the time derivative of $\mathbf{M}\dot{\mathbf{q}}$. Non-conservative forces are added to the right side of the equation. 

We employ implicit Euler for stability in large steps~\citep{baraff_large_1998}. Given the acceleration $\ddot{\mathbf{q}} = \mathbf{M}^{-1} \mathbf{F}$ and the change of speed over time step $h$, $\Delta \dot{\mathbf{q}}$ can be approximated by $\Delta \dot{\mathbf{q}} = h \mathbf{M}^{-1} \mathbf{F}_{(t+1)}$, where $\mathbf{F}_{(t+1)}$ is the force at $t+1$ that can be approximated by first-order Taylor expansion $\mathbf{F}_{(t+1)} = \mathbf{F}_{(t)} + \frac{\partial \mathbf{F}_{(t)}}{\partial \mathbf{q}} \Delta \mathbf{q} + \frac{\partial \mathbf{F}_{(t)}}{\partial \dot{\mathbf{q}}} \Delta \dot{\mathbf{q}}$, where $\mathbf{F}_{(t)}$ is the force at $t$ which can be computed by Equation \ref{eqn:app_Lag}. Then node positions at $t+1$ are $\mathbf{q}_{(t+1)} = \mathbf{q}_{(t)} + h (\dot{\mathbf{q}}_{(t)} + \Delta \dot{\mathbf{q}})$.
Finally, we have the system equation for simulation:
\begin{equation}
\label{eqn:app_linear}
    \left(\mathbf{M} - \frac{\partial \mathbf{F}_{(t)}}{\partial \mathbf{q}}h^2 - \frac{\partial \mathbf{F}_{(t)}}{\partial \dot{\mathbf{q}}}h\right)\dot{\mathbf{q}}_{(t+1)} = h \left(\mathbf{F}_{(t)} - \frac{\partial \mathbf{F}_{(t)}}{\partial \dot{\mathbf{q}}} \right) + \mathbf{M} \dot{\mathbf{q}}_{(t)}
\end{equation}

To solve Equation~\ref{eqn:app_linear}, we explain every term including the general mass matrix $\mathbf{M}$ and every force contained in $\mathbf{F}_{(t)}$ below.

\subsection{General Mass Matrix}
The mass matrix of a warp segment $[\mathbf{q}_0,\mathbf{q}_1 ]$ is 
\begin{equation}
    \mathbf{M}_{0,1} 
    = \frac{1}{6}\Delta u \rho
    \begin{pmatrix}
    2\mathbf{I}_3 & \mathbf{I}_3 & -2\mathbf{w} & -\mathbf{w} \\
    \mathbf{I}_3 & 2\mathbf{I}_3 & -\mathbf{w} & -2\mathbf{w} \\
    -2\mathbf{w}^{\top} & -\mathbf{w}^{\top} & 2\mathbf{w}^{\top}\mathbf{w} & \mathbf{w}^{\top}\mathbf{w} \\
    -\mathbf{w}^{\top} & -2\mathbf{w}^{\top} & \mathbf{w}^{\top}\mathbf{w} & 2\mathbf{w^{\top}w}\\
    \end{pmatrix}
\end{equation}
where $\Delta u = u_1 - u_0$ is the distance between the two nodes in Eulerian coordinates, $\mathbf{w} = \frac{\mathbf{x}_1 - \mathbf{x}_0}{\Delta u}$, and $\rho$ is yarn's linear density. The partial derivatives of general mass matrix with respect to nodes' position is 
\begin{equation}
\label{eq:spatialDerivative_mass}
    \begin{pmatrix}
        \frac{\partial \mathbf{M}_{0,1}}{\partial \mathbf{x}_0} &
        \frac{\partial \mathbf{M}_{0,1}}{\partial \mathbf{x}_1} &
        \frac{\partial \mathbf{M}_{0,1}}{\partial u_0} &
        \frac{\partial \mathbf{M}_{0,1}}{\partial u_1} 
    \end{pmatrix}^{\top}
\end{equation}
As $\mathbf{x}_0$ and $\mathbf{x}_1$ are vectors:
\begin{equation}
    \frac{\partial \mathbf{M}_{0,1}}{\partial \mathbf{x}_0} =
    \begin{pmatrix}
        \frac{\partial \mathbf{M}_{0,1}}{\partial \mathbf{x}_0^{(1)}} \\
        \frac{\partial \mathbf{M}_{0,1}}{\partial \mathbf{x}_0^{(2)}} \\
        \frac{\partial \mathbf{M}_{0,1}}{\partial \mathbf{x}_0^{(3)}} 
    \end{pmatrix}
    \mbox{and}\;
        \frac{\partial \mathbf{M}_{0,1}}{\partial \mathbf{x}_1} =
    \begin{pmatrix}
        \frac{\partial \mathbf{M}_{0,1}}{\partial \mathbf{x}_1^{(1)}} \\
        \frac{\partial \mathbf{M}_{0,1}}{\partial \mathbf{x}_1^{(2)}} \\
        \frac{\partial \mathbf{M}_{0,1}}{\partial \mathbf{x}_1^{(3)}} 
    \end{pmatrix}
\end{equation}
The component $\frac{\partial \mathbf{M}_{0,1} }{\partial \mathbf{x}_0^{(1)}}$ is
\begin{equation}
    \frac{\partial \mathbf{M}_{0,1} }{\partial \mathbf{x}_0^{(1)}}
    = \frac{1}{6} \Delta u \rho
    \begin{pmatrix}
    \mathbf{0} & \mathbf{0} & 
    -2 \frac{\partial \mathbf{w}}{\partial \mathbf{x}_0^{(1)}} & 
    -\frac{\partial \mathbf{w}}{\partial \mathbf{x}_0^{(1)}} \\
    \mathbf{0} & \mathbf{0} & 
    - \frac{\partial \mathbf{w}}{\partial \mathbf{x}_0^{(1)}} & 
    -2\frac{\partial \mathbf{w}}{\partial \mathbf{x}_0^{(1)}} \\
    -2 \frac{\partial \mathbf{w}^{\top}}{\partial \mathbf{x}_0^{(1)}} & 
    -\frac{\partial \mathbf{w}^{\top}}{\partial \mathbf{x}_0^{(1)}} & 
    2 \frac{\partial \mathbf{w}^{\top}\mathbf{w}}{\partial \mathbf{x}_0^{(1)}} & 
    \frac{\partial \mathbf{w}^{\top}\mathbf{w}}{\partial \mathbf{x}_0^{(1)}} \\
    -\frac{\partial \mathbf{w}^{\top}}{\partial \mathbf{x}_0^{(1)}} & 
    -2 \frac{\partial \mathbf{w}^{\top}}{\partial \mathbf{x}_0^{(1)}} & 
    \frac{\partial \mathbf{w}^{\top}\mathbf{w}}{\partial \mathbf{x}_0^{(1)}} & 
    2 \frac{\partial \mathbf{w}^{\top}\mathbf{w}}{\partial \mathbf{x}_0^{(1)}}
    \end{pmatrix} \nonumber
\end{equation}

and $\frac{\partial \mathbf{M}_{0,1} }{\partial \mathbf{x}_0^{(2)}}$, $\frac{\partial \mathbf{M}_{0,1} }{\partial \mathbf{x}_0^{(3)}}$, $\frac{\partial \mathbf{M}_{0,1} }{\partial \mathbf{x}_1^{(1)}}$, $\frac{\partial \mathbf{M}_{0,1} }{\partial \mathbf{x}_1^{(2)}}$ and $\frac{\partial \mathbf{M}_{0,1} }{\partial \mathbf{x}_1^{(3)}}$ are in a similar form as $\frac{\partial \mathbf{M}_{0,1} }{\partial \mathbf{x}_0^{(1)}}$. In each term, we have:

\begin{align}
    \frac{\partial \mathbf{w}}{\partial \mathbf{x}_0^{(1)}} = 
    - \begin{pmatrix}
        \frac{1}{\Delta u} \\ 0 \\ 0
    \end{pmatrix}
    \mbox{,}\;
    \frac{\partial \mathbf{w}}{\partial \mathbf{x}_0^{(2)}} = 
    - \begin{pmatrix}
        0 \\ \frac{1}{\Delta u} \\ 0
    \end{pmatrix}
    \mbox{, and}\;
    \frac{\partial \mathbf{w}}{\partial \mathbf{x}_0^{(3)}} = 
    - \begin{pmatrix}
        0 \\ 0 \\ \frac{1}{\Delta u}
    \end{pmatrix} \nonumber \\ 
    \frac{\partial \mathbf{w}}{\partial \mathbf{x}_1^{(1)}} = 
    \begin{pmatrix}
         \frac{1}{\Delta u} \\ 0 \\ 0
    \end{pmatrix}
    \mbox{,}\;
    \frac{\partial \mathbf{w}}{\partial \mathbf{x}_1^{(2)}} = 
    \begin{pmatrix}
        0 \\ \frac{1}{\Delta u} \\ 0
    \end{pmatrix}
    \mbox{, and}\;
    \frac{\partial \mathbf{w}}{\partial \mathbf{x}_1^{(3)}} = 
    \begin{pmatrix}
        0 \\ 0 \\ \frac{1}{\Delta u}
    \end{pmatrix} \nonumber \\ 
    \frac{\partial \mathbf{w}^{\top}\mathbf{w}}{\partial \mathbf{x}_0^{(1)}} =
    \frac{\partial \mathbf{w}^{\top}}{\partial \mathbf{x}_0^{(1)}} \mathbf{w} +
    \mathbf{w}^{\top}\frac{\partial \mathbf{w}}{\partial \mathbf{x}_0^{(1)}} =
    -2\frac{\mathbf{x}_1^{(1)} - \mathbf{x}_0^{(1)}}{\Delta u^2} \nonumber
\end{align}
where $\frac{\partial \mathbf{w}^{\top}\mathbf{w}}{\partial \mathbf{x}_0^{(2)}}$, $\frac{\partial \mathbf{w}^{\top}\mathbf{w}}{\partial \mathbf{x}_0^{(3)}}$, $\frac{\partial \mathbf{w}^{\top}\mathbf{w}}{\partial \mathbf{x}_1^{(1)}}$, $\frac{\partial \mathbf{w}^{\top}\mathbf{w}}{\partial \mathbf{x}_1^{(2)}}$ and $\frac{\partial \mathbf{w}^{\top}\mathbf{w}}{\partial \mathbf{x}_1^{(3)}}$ have a similar form as $\frac{\partial \mathbf{w}^{\top}\mathbf{w}}{\partial \mathbf{x}_0^{(1)}}$.

Unsurprisingly, we can find that
\begin{equation}
    \frac{\partial \mathbf{M}_{0,1}}{\partial \mathbf{x}_0^{(1)}} = -\frac{\partial \mathbf{M}_{0,1}}{\partial \mathbf{x}_1^{(1)}}
    \;\mbox{,}\;
    \frac{\partial \mathbf{M}_{0,1}}{\partial \mathbf{x}_0^{(2)}} = -\frac{\partial \mathbf{M}_{0,1}}{\partial \mathbf{x}_1^{(2)}}
    \;\mbox{and}\;
    \frac{\partial \mathbf{M}_{0,1}}{\partial \mathbf{x}_0^{(3)}} = -\frac{\partial \mathbf{M}_{0,1}}{\partial \mathbf{x}_1^{(3)}} \nonumber
\end{equation}
After deriving the partial derivatives of $\mathbf{M}_{0,1}$ with respect to the Lagrangian coordinates, we give its partial derivatives with respect to Eulerian coordinates:
\begin{align}
    \frac{\partial \mathbf{M}_{0,1} }{\partial u_0}
    = &-\frac{1}{6} \rho
    \begin{pmatrix}
    2\mathbf{I}_3 & \mathbf{I}_3 & -2\mathbf{w} & -\mathbf{w} \\
    \mathbf{I}_3 & 2\mathbf{I}_3 & -\mathbf{w} & -2\mathbf{w} \\
    -2\mathbf{w}^{\top} & -\mathbf{w}^{\top} & 2\mathbf{w}^{\top}\mathbf{w} & \mathbf{w}^{\top}\mathbf{w} \\
    -\mathbf{w}^{\top} & -2\mathbf{w}^{\top} & \mathbf{w}^{\top}\mathbf{w} & 2\mathbf{w^{\top}w}\\
    \end{pmatrix} \notag \\
    &+ \frac{1}{6} \Delta u \rho
    \begin{pmatrix}
    \mathbf{0} & \mathbf{0} & -2 \frac{\partial \mathbf{w}}{\partial u_0} & -\frac{\partial \mathbf{w}}{\partial u_0} \\
    \mathbf{0} & \mathbf{0} & - \frac{\partial \mathbf{w}}{\partial u_0} & -2\frac{\partial \mathbf{w}}{\partial u_0} \\
    -2 \frac{\partial \mathbf{w}^{\top}}{\partial u_0} & -\frac{\partial \mathbf{w}^{\top}}{\partial u_0} & 2 \frac{\partial \mathbf{w}^{\top}\mathbf{w}}{\partial u_0} & \frac{\partial \mathbf{w}^{\top}\mathbf{w}}{\partial u_0} \\
    -\frac{\partial \mathbf{w}^{\top}}{\partial u_0} & -2 \frac{\partial \mathbf{w}^{\top}}{\partial u_0} & \frac{\partial \mathbf{w}^{\top}\mathbf{w}}{\partial u_0} & 2 \frac{\partial \mathbf{w}^{\top}\mathbf{w}}{\partial u_0}
    \end{pmatrix}
\end{align}

where $\frac{\partial \mathbf{M}_{0,1} }{\partial u_1}$ has a similar form as $\frac{\partial \mathbf{M}_{0,1} }{\partial u_0}$ and:
\begin{equation}
    \frac{\partial \mathbf{w}}{\partial u_0} 
    = \frac{\mathbf{x}_1 - \mathbf{x}_0}{\Delta u^2} 
    = \frac{\mathbf{w}}{\Delta u}
    \;\mbox{and}\;
    \frac{\partial \mathbf{w}}{\partial u_1} 
    = -\frac{\mathbf{x}_1 - \mathbf{x}_0}{\Delta u^2}
    = -\frac{\mathbf{w}}{\Delta u} \nonumber
\end{equation}

\begin{equation}
    \frac{\partial \mathbf{w}^{\top}\mathbf{w}}{\partial u_0}
    = \frac{\partial \mathbf{w}^{\top}}{\partial u_0} \mathbf{w}
    + \mathbf{w}^{\top} \frac{\partial \mathbf{w}}{\partial u_0}
    = 2 \frac{\mathbf{w}^{\top}\mathbf{w}}{\Delta u} \nonumber
\end{equation}

\begin{equation}
    \frac{\partial \mathbf{w}^{\top}\mathbf{w}}{\partial u_1}
    = \frac{\partial \mathbf{w}^{\top}}{\partial u_1} \mathbf{w}
    + \mathbf{w}^{\top} \frac{\partial \mathbf{w}}{\partial u_1}
    = -2 \frac{\mathbf{w}^{\top}\mathbf{w}}{\Delta u} \nonumber
\end{equation}
Likewise, we can find
\begin{equation}
    \frac{\partial \mathbf{M}_{0,1} }{\partial u_0} = -\frac{\partial \mathbf{M}_{0,1} }{\partial u_1} \nonumber
\end{equation}
So far, we have given the full details of $\mathbf{M}_{0,1}$' s partial derivatives with respect to positions in Equation~\ref{eq:spatialDerivative_mass}. Now we give its time derivative:
\begin{align}
    \dot{\mathbf{M}}_{0,1} &= 
    \frac{1}{6} \rho (\dot{u}_1 - \dot{u}_0)
    \begin{pmatrix}
        2\mathbf{I}_3 & \mathbf{I}_3 & -2\mathbf{w} & -\mathbf{w} \\
        \mathbf{I}_3 & 2\mathbf{I}_3 & -\mathbf{w} & -2\mathbf{w} \\
        -2\mathbf{w}^{\top} & -\mathbf{w}^{\top} & 2\mathbf{w}^{\top}\mathbf{w} & \mathbf{w}^{\top}\mathbf{w} \\
        -\mathbf{w}^{\top} & -2\mathbf{w}^{\top} & \mathbf{w}^{\top}\mathbf{w} & 2\mathbf{w^{\top}w}\\
    \end{pmatrix} \nonumber\\
    & + \frac{1}{6} \rho \Delta u
    \begin{pmatrix}
    \mathbf{0} & \mathbf{0} & 
    -2 \frac{\partial \mathbf{w}}{\partial t} &
    -\frac{\partial \mathbf{w}}{\partial t} \\
    \mathbf{0} & \mathbf{0} & 
    - \frac{\partial \mathbf{w}}{\partial t} &
    -2\frac{\partial \mathbf{w}}{\partial t} \\
    -2 \frac{\partial \mathbf{w}^{\top}}{\partial t} &
    -\frac{\partial \mathbf{w}^{\top}}{\partial t} &
    2 \frac{\partial \mathbf{w}^{\top}\mathbf{w}}{\partial t} &
    \frac{\partial \mathbf{w}^{\top}\mathbf{w}}{\partial t} \\
    -\frac{\partial \mathbf{w}^{\top}}{\partial t} &
    -2\frac{\partial \mathbf{w}^{\top}}{\partial t} &
    \frac{\partial \mathbf{w}^{\top}\mathbf{w}}{\partial t} &
    2\frac{\partial \mathbf{w}^{\top}\mathbf{w}}{\partial t}
    \end{pmatrix}
\end{align}
where
\begin{align}
    \frac{\partial \mathbf{w}}{\partial t} = 
    \frac{\partial}{\partial t} \frac{\mathbf{x}_1 - \mathbf{x}_0}{\Delta u} =
    \frac{(\dot{\mathbf{x}}_1 - \dot{\mathbf{x}}_0)\Delta u - (\mathbf{x}_1 - \mathbf{x}_0)(\dot{u}_1 - \dot{u}_0)}{\Delta u^2} \nonumber \\
    \frac{\partial \mathbf{w}^{\top}\mathbf{w}}{\partial t} = 
    \frac{\partial \mathbf{w}^{\top}}{\partial t} \mathbf{w} + 
    \mathbf{w}^{\top} \frac{\partial \mathbf{w}}{\partial t} \nonumber
\end{align}

In addition, the derivatives of $\dot{\mathbf{M}}_{0,1}\dot{\mathbf{q}}_{0,1}$ with respect to the nodes' positions are:
\begin{align}
\label{eq:spatialDerivatives_MdotQdot}
    \frac{\partial \dot{\mathbf{M}}_{0,1}\dot{\mathbf{q}}_{0,1}}{\partial \mathbf{x}_0} =
    \begin{pmatrix}
        \frac{\partial \dot{\mathbf{M}}_{0,1}}{\partial \mathbf{x}_0^{(1)}}\dot{\mathbf{q}}_{0,1} \\
        \frac{\partial \dot{\mathbf{M}}_{0,1}}{\partial \mathbf{x}_0^{(2)}}\dot{\mathbf{q}}_{0,1} \\
        \frac{\partial \dot{\mathbf{M}}_{0,1}}{\partial \mathbf{x}_0^{(3)}}\dot{\mathbf{q}}_{0,1}
    \end{pmatrix} \mbox{, }
    \frac{\partial \dot{\mathbf{M}}_{0,1}\dot{\mathbf{q}}_{0,1}}{\partial \mathbf{x}_1} =
    \begin{pmatrix}
        \frac{\partial \dot{\mathbf{M}}_{0,1}}{\partial \mathbf{x}_1^{(1)}}\dot{\mathbf{q}}_{0,1} \\
        \frac{\partial \dot{\mathbf{M}}_{0,1}}{\partial \mathbf{x}_1^{(2)}}\dot{\mathbf{q}}_{0,1} \\
        \frac{\partial \dot{\mathbf{M}}_{0,1}}{\partial \mathbf{x}_1^{(3)}}\dot{\mathbf{q}}_{0,1}
    \end{pmatrix} \nonumber \\
    \frac{\partial \dot{\mathbf{M}}_{0,1}\dot{\mathbf{q}}_{0,1}}{\partial u_0} =
    \frac{\partial \dot{\mathbf{M}}_{0,1}}{\partial u_0}\dot{\mathbf{q}}_{0,1}
    \mbox{, }
    \frac{\partial \dot{\mathbf{M}}_{0,1}\dot{\mathbf{q}}_{0,1}}{\partial u_1} = 
    \frac{\partial \dot{\mathbf{M}}_{0,1}}{\partial u_1}\dot{\mathbf{q}}_{0,1}
\end{align}



The components in Equation~\ref{eq:spatialDerivatives_MdotQdot} are:
\begin{align}
    \frac{\partial \dot{\mathbf{M}}_{0,1}}{\partial \mathbf{x}_0^{(1)}} &= 
    \frac{1}{6} \rho (\dot{u}_1 - \dot{u}_0) 
    \begin{pmatrix}
        \mathbf{0} & \mathbf{0} & 
        -2 \frac{\partial \mathbf{w}}{\partial \mathbf{x}_0^{(1)}} & 
        -\frac{\partial \mathbf{w}}{\partial \mathbf{x}_0^{(1)}} \\
        \mathbf{0} & \mathbf{0} & 
        - \frac{\partial \mathbf{w}}{\partial \mathbf{x}_0^{(1)}} & 
        -2\frac{\partial \mathbf{w}}{\partial \mathbf{x}_0^{(1)}} \\
        -2 \frac{\partial \mathbf{w}^{\top}}{\partial \mathbf{x}_0^{(1)}} & 
        -\frac{\partial \mathbf{w}^{\top}}{\partial \mathbf{x}_0^{(1)}} & 
        2 \frac{\partial \mathbf{w}^{\top}\mathbf{w}}{\partial \mathbf{x}_0^{(1)}} & 
        \frac{\partial \mathbf{w}^{\top}\mathbf{w}}{\partial \mathbf{x}_0^{(1)}} \\
        -\frac{\partial \mathbf{w}^{\top}}{\partial \mathbf{x}_0^{(1)}} & 
        -2 \frac{\partial \mathbf{w}^{\top}}{\partial \mathbf{x}_0^{(1)}} & 
        \frac{\partial \mathbf{w}^{\top}\mathbf{w}}{\partial \mathbf{x}_0^{(1)}} & 
        2 \frac{\partial \mathbf{w}^{\top}\mathbf{w}}{\partial \mathbf{x}_0^{(1)}}
    \end{pmatrix} \nonumber \\
     &+ \frac{1}{6} \rho \Delta u
    \begin{pmatrix}
        \mathbf{0} & \mathbf{0} & 
        -2 \frac{\partial^2 \mathbf{w}}{\partial t \partial \mathbf{x}_0^{(1)}} &
        -\frac{\partial^2 \mathbf{w}}{\partial t \partial \mathbf{x}_0^{(1)}} \\
        \mathbf{0} & \mathbf{0} & 
        - \frac{\partial^2 \mathbf{w}}{\partial t \partial \mathbf{x}_0^{(1)}} &
        -2\frac{\partial^2 \mathbf{w}}{\partial t \partial \mathbf{x}_0^{(1)}} \\
        -2 \frac{\partial^2 \mathbf{w}^{\top}}{\partial t \partial \mathbf{x}_0^{(1)}} &
        -\frac{\partial^2 \mathbf{w}^{\top}}{\partial t \partial \mathbf{x}_0^{(1)}} &
        2 \frac{\partial^2 \mathbf{w}^{\top}\mathbf{w}}{\partial t \partial \mathbf{x}_0^{(1)}} &
        \frac{\partial^2 \mathbf{w}^{\top}\mathbf{w}}{\partial t \partial \mathbf{x}_0^{(1)}} \\
        -\frac{\partial^2 \mathbf{w}^{\top}}{\partial t \partial \mathbf{x}_0^{(1)}} &
        -2\frac{\partial^2 \mathbf{w}^{\top}}{\partial t \partial \mathbf{x}_0^{(1)}} &
        \frac{\partial^2 \mathbf{w}^{\top}\mathbf{w}}{\partial t \partial \mathbf{x}_0^{(1)}} &
        2\frac{\partial^2 \mathbf{w}^{\top}\mathbf{w}}{\partial t \partial \mathbf{x}_0^{(1)}} \nonumber
    \end{pmatrix}
\end{align}
and 
\begin{align}
    \frac{\partial \dot{\mathbf{M}}_{0,1}}{\partial u_0} &=
    \frac{1}{6} \rho (\dot{u}_1 - \dot{u}_0) 
    \begin{pmatrix}
        \mathbf{0} & \mathbf{0} & -2 \frac{\partial \mathbf{w}}{\partial u_0} & -\frac{\partial \mathbf{w}}{\partial u_0} \\
        \mathbf{0} & \mathbf{0} & - \frac{\partial \mathbf{w}}{\partial u_0} & -2\frac{\partial \mathbf{w}}{\partial u_0} \\
        -2 \frac{\partial \mathbf{w}^{\top}}{\partial u_0} & -\frac{\partial \mathbf{w}^{\top}}{\partial u_0} & 2 \frac{\partial \mathbf{w}^{\top}\mathbf{w}}{\partial u_0} & \frac{\partial \mathbf{w}^{\top}\mathbf{w}}{\partial u_0} \\
        -\frac{\partial \mathbf{w}^{\top}}{\partial u_0} & -2 \frac{\partial \mathbf{w}^{\top}}{\partial u_0} & \frac{\partial \mathbf{w}^{\top}\mathbf{w}}{\partial u_0} & 2 \frac{\partial \mathbf{w}^{\top}\mathbf{w}}{\partial u_0}
    \end{pmatrix} \nonumber\\ 
    &- \frac{1}{6} \rho
    \begin{pmatrix}
        \mathbf{0} & \mathbf{0} & 
        -2 \frac{\partial \mathbf{w}}{\partial t} &
        -\frac{\partial \mathbf{w}}{\partial t} \\
        \mathbf{0} & \mathbf{0} & 
        - \frac{\partial \mathbf{w}}{\partial t} &
        -2\frac{\partial \mathbf{w}}{\partial t} \\
        -2 \frac{\partial \mathbf{w}^{\top}}{\partial t} &
        -\frac{\partial \mathbf{w}^{\top}}{\partial t} &
        2 \frac{\partial \mathbf{w}^{\top}\mathbf{w}}{\partial t} &
        \frac{\partial \mathbf{w}^{\top}\mathbf{w}}{\partial t} \\
        -\frac{\partial \mathbf{w}^{\top}}{\partial t} &
        -2\frac{\partial \mathbf{w}^{\top}}{\partial t} &
        \frac{\partial \mathbf{w}^{\top}\mathbf{w}}{\partial t} &
        2\frac{\partial \mathbf{w}^{\top}\mathbf{w}}{\partial t}
    \end{pmatrix} \nonumber\\
    & + \frac{1}{6} \Delta u \rho
    \begin{pmatrix}
        \mathbf{0} & \mathbf{0} & 
        -2 \frac{\partial^2 \mathbf{w}}{\partial t \partial u_0} &
        -\frac{\partial^2 \mathbf{w}}{\partial t \partial u_0} \\
        \mathbf{0} & \mathbf{0} & 
        - \frac{\partial^2 \mathbf{w}}{\partial t \partial u_0} &
        -2\frac{\partial^2 \mathbf{w}}{\partial t \partial u_0} \\
        -2 \frac{\partial^2 \mathbf{w}^{\top}}{\partial t \partial u_0} &
        -\frac{\partial^2 \mathbf{w}^{\top}}{\partial t \partial u_0} &
        2 \frac{\partial^2 \mathbf{w}^{\top}\mathbf{w}}{\partial t \partial u_0} &
        \frac{\partial^2 \mathbf{w}^{\top}\mathbf{w}}{\partial t \partial u_0} \\
        -\frac{\partial^2 \mathbf{w}^{\top}}{\partial t \partial u_0} &
        -2\frac{\partial^2 \mathbf{w}^{\top}}{\partial t \partial u_0} &
        \frac{\partial^2 \mathbf{w}^{\top}\mathbf{w}}{\partial t \partial u_0} &
        2\frac{\partial^2 \mathbf{w}^{\top}\mathbf{w}}{\partial t \partial u_0}\nonumber
    \end{pmatrix}
\end{align}

where
\begin{equation}
    \frac{\partial^2 \mathbf{w}}{\partial t \partial \mathbf{x}_0^{(1)}} = 
    \begin{pmatrix}
        \frac{\dot{u}_1 - \dot{u}_0}{\Delta u^2} \\ 0 \\ 0
    \end{pmatrix}
    \;\mbox{,}\;
    \frac{\partial^2 \mathbf{w}}{\partial t \partial \mathbf{x}_0^{(2)}} = 
    \begin{pmatrix}
        0 \\ \frac{\dot{u}_1 - \dot{u}_0}{\Delta u^2} \\ 0
    \end{pmatrix}
    \;\mbox{, and}\;
    \frac{\partial^2 \mathbf{w}}{\partial t \partial \mathbf{x}_0^{(2)}} = 
    \begin{pmatrix}
        0 \\ 0 \\ \frac{\dot{u}_1 - \dot{u}_0}{\Delta u^2}
    \end{pmatrix} \nonumber
\end{equation}
\begin{equation}
    \frac{\partial^2 \mathbf{w}}{\partial t \partial \mathbf{x}_1^{(1)}} = 
    -\begin{pmatrix}
        \frac{\dot{u}_1 - \dot{u}_0}{\Delta u^2} \\ 0 \\ 0
    \end{pmatrix}
    \;\mbox{,}\;
    \frac{\partial^2 \mathbf{w}}{\partial t \partial \mathbf{x}_1^{(2)}} = 
    -\begin{pmatrix}
        0 \\ \frac{\dot{u}_1 - \dot{u}_0}{\Delta u^2} \\ 0
    \end{pmatrix}
    \;\mbox{, and}\;
    \frac{\partial^2 \mathbf{w}}{\partial t \partial \mathbf{x}_1^{(2)}} = 
    -\begin{pmatrix}
        0 \\ 0 \\ \frac{\dot{u}_1 - \dot{u}_0}{\Delta u^2}
    \end{pmatrix} \nonumber
\end{equation}

\begin{equation}
    \frac{\partial^2 \mathbf{w}}{\partial t \partial u_0} = 
    \frac{\dot{\mathbf{x}}_1 - \dot{\mathbf{x}}_1}{(u_1 - u_0)^2} - 
    \frac{2(\mathbf{x}_1 - \mathbf{x}_0)(\dot{u}_1 - \dot{u}_0)}{(u_1 - u_0)^3} \nonumber
\end{equation}
\begin{equation}
    \frac{\partial^2 \mathbf{w}}{\partial t \partial u_1} = 
    -\frac{\dot{\mathbf{x}}_1 - \dot{\mathbf{x}}_1}{(u_1 - u_0)^2} + 
    \frac{2(\mathbf{x}_1 - \mathbf{x}_0)(\dot{u}_1 - \dot{u}_0)}{(u_1 - u_0)^3} \nonumber
\end{equation}

\begin{equation}
    \frac{\partial^2 \mathbf{w}^{\top}\mathbf{w}}{\partial t \partial \mathbf{x}_0^{(1)}} = 
    \frac{\partial \mathbf{w}^{\top}}{\partial \mathbf{x}_1^{(1)}} \frac{\partial \mathbf{w}}{\partial t} +
    \mathbf{w}^{\top} \frac{\partial^2 \mathbf{w}}{\partial t \partial \mathbf{x}_1^{(1)}} +
    \frac{\partial^2 \mathbf{w}^{\top}}{\partial t \partial \mathbf{x}_1^{(1)}} \mathbf{w} +
    \frac{\partial \mathbf{w}^{\top}}{\partial t} \frac{\partial \mathbf{w}}{\partial \mathbf{x}_1^{(1)}} \nonumber
\end{equation}

\begin{equation}
    \frac{\partial^2 \mathbf{w}^{\top}\mathbf{w}}{\partial t \partial u_0} = 
    \frac{\partial \mathbf{w}^{\top}}{\partial u_0} \frac{\partial \mathbf{w}}{\partial t} +
    \mathbf{w}^{\top} \frac{\partial^2 \mathbf{w}}{\partial t \partial u_0} +
    \frac{\partial^2 \mathbf{w}^{\top}}{\partial t \partial u_0} \mathbf{w} +
    \frac{\partial \mathbf{w}^{\top}}{\partial t} \frac{\partial \mathbf{w}}{\partial u_0} \nonumber
\end{equation}

The derivatives of $\dot{\mathbf{M}}_{0,1}\dot{\mathbf{q}}_{0,1}$ with respect to the nodes' velocities are:
\begin{equation}
    \frac{\partial \dot{\mathbf{M}}_{0,1}\dot{\mathbf{q}}_{0,1}}{\partial \dot{\mathbf{x}}_0} =
    \begin{pmatrix}
        \frac{\partial \dot{\mathbf{M}}_{0,1}\dot{\mathbf{q}}_{0,1}}{\partial \dot{\mathbf{x}}_0^{(1)}} \\
        \frac{\partial \dot{\mathbf{M}}_{0,1}\dot{\mathbf{q}}_{0,1}}{\partial \dot{\mathbf{x}}_0^{(2)}} \\
        \frac{\partial \dot{\mathbf{M}}_{0,1}\dot{\mathbf{q}}_{0,1}}{\partial \dot{\mathbf{x}}_0^{(3)}}
    \end{pmatrix} =
    \begin{pmatrix}
        \frac{\partial \dot{\mathbf{M}}_{0,1}}{\partial \dot{\mathbf{x}}_0^{(1)}} \dot{\mathbf{q}}_{0,1} +
        \dot{\mathbf{M}}_{0,1} \frac{\partial \dot{\mathbf{q}}_{0,1}}{\partial \dot{\mathbf{x}}_0^{(1)}} \\
        \frac{\partial \dot{\mathbf{M}}_{0,1}}{\partial \dot{\mathbf{x}}_0^{(2)}} \dot{\mathbf{q}}_{0,1} +
        \dot{\mathbf{M}}_{0,1} \frac{\partial \dot{\mathbf{q}}_{0,1}}{\partial \dot{\mathbf{x}}_0^{(2)}} \\
        \frac{\partial \dot{\mathbf{M}}_{0,1}}{\partial \dot{\mathbf{x}}_0^{(3)}}\dot{\mathbf{q}}_{0,1} +
        \dot{\mathbf{M}}_{0,1} \frac{\partial \dot{\mathbf{q}}_{0,1}}{\partial \dot{\mathbf{x}}_0^{(3)}} \nonumber
    \end{pmatrix}
\end{equation}

\begin{equation}
    \frac{\partial \dot{\mathbf{M}}_{0,1}\dot{\mathbf{q}}_{0,1}}{\partial \dot{\mathbf{x}}_1} =
    \begin{pmatrix}
        \frac{\partial \dot{\mathbf{M}}_{0,1}\dot{\mathbf{q}}_{0,1}}{\partial \dot{\mathbf{x}}_1^{(1)}} \\
        \frac{\partial \dot{\mathbf{M}}_{0,1}\dot{\mathbf{q}}_{0,1}}{\partial \dot{\mathbf{x}}_1^{(2)}} \\
        \frac{\partial \dot{\mathbf{M}}_{0,1}\dot{\mathbf{q}}_{0,1}}{\partial \dot{\mathbf{x}}_1^{(3)}}
    \end{pmatrix} = 
    \begin{pmatrix}
        \frac{\partial \dot{\mathbf{M}}_{0,1}}{\partial \dot{\mathbf{x}}_1^{(1)}} \dot{\mathbf{q}}_{0,1} +
        \dot{\mathbf{M}}_{0,1} \frac{\partial \dot{\mathbf{q}}_{0,1}}{\partial \dot{\mathbf{x}}_1^{(1)}} \\
        \frac{\partial \dot{\mathbf{M}}_{0,1}}{\partial \dot{\mathbf{x}}_1^{(2)}} \dot{\mathbf{q}}_{0,1} +
        \dot{\mathbf{M}}_{0,1} \frac{\partial \dot{\mathbf{q}}_{0,1}}{\partial \dot{\mathbf{x}}_1^{(2)}} \\
        \frac{\partial \dot{\mathbf{M}}_{0,1}}{\partial \dot{\mathbf{x}}_1^{(3)}}\dot{\mathbf{q}}_{0,1} +
        \dot{\mathbf{M}}_{0,1} \frac{\partial \dot{\mathbf{q}}_{0,1}}{\partial \dot{\mathbf{x}}_1^{(3)}}
    \end{pmatrix} \nonumber
\end{equation}

\begin{equation}
    \frac{\partial \dot{\mathbf{M}}_{0,1}\dot{\mathbf{q}}_{0,1}}{\partial \dot{u}_0} =
    \frac{\partial \dot{\mathbf{M}}_{0,1}}{\partial \dot{u}_0}\dot{\mathbf{q}}_{0,1} + 
    \dot{\mathbf{M}}_{0,1} \frac{\partial \dot{\mathbf{q}}_{0,1}}{\partial \dot{u}_0} \nonumber
\end{equation}

\begin{equation}
    \frac{\partial \dot{\mathbf{M}}_{0,1}\dot{\mathbf{q}}_{0,1}}{\partial \dot{u}_1} = 
    \frac{\partial \dot{\mathbf{M}}_{0,1}}{\partial \dot{u}_1}\dot{\mathbf{q}}_{0,1} +
    \dot{\mathbf{M}}_{0,1} \frac{\partial \dot{\mathbf{q}}_{0,1}}{\partial \dot{u}_1}
\end{equation}
where
\begin{equation}
    \frac{\partial \dot{\mathbf{M}}_{0,1}}{\partial \dot{\mathbf{x}}_1^{(1)}} = 
    \frac{1}{6} \Delta u \rho
    \begin{pmatrix}
        \mathbf{0} & \mathbf{0} & 
        -2 \frac{\partial^2 \mathbf{w}}{\partial t \partial \mathbf{x}_0^{(1)}} &
        -\frac{\partial^2 \mathbf{w}}{\partial t \partial \mathbf{x}_0^{(1)}} \\
        \mathbf{0} & \mathbf{0} & 
        - \frac{\partial^2 \mathbf{w}}{\partial t \partial \mathbf{x}_0^{(1)}} &
        -2\frac{\partial^2 \mathbf{w}}{\partial t \partial \mathbf{x}_0^{(1)}} \\
        -2 \frac{\partial^2 \mathbf{w}^{\top}}{\partial t \partial \mathbf{x}_0^{(1)}} &
        -\frac{\partial^2 \mathbf{w}^{\top}}{\partial t \partial \mathbf{x}_0^{(1)}} &
        2 \frac{\partial^2 \mathbf{w}^{\top}\mathbf{w}}{\partial t \partial \mathbf{x}_0^{(1)}} &
        \frac{\partial^2 \mathbf{w}^{\top}\mathbf{w}}{\partial t \partial \mathbf{x}_0^{(1)}} \\
        -\frac{\partial^2 \mathbf{w}^{\top}}{\partial t \partial \mathbf{x}_0^{(1)}} &
        -2\frac{\partial^2 \mathbf{w}^{\top}}{\partial t \partial \mathbf{x}_0^{(1)}} &
        \frac{\partial^2 \mathbf{w}^{\top}\mathbf{w}}{\partial t \partial \mathbf{x}_0^{(1)}} &
        2\frac{\partial^2 \mathbf{w}^{\top}\mathbf{w}}{\partial t \partial \mathbf{x}_0^{(1)}}
    \end{pmatrix} \nonumber
\end{equation}

\begin{align}
    \frac{\partial \dot{\mathbf{M}}_{0,1}}{\partial \dot{u}_0} &= 
    - \frac{1}{6} \rho 
        \begin{pmatrix}
        2\mathbf{I}_3 & \mathbf{I}_3 & -2\mathbf{w} & -\mathbf{w} \\
        \mathbf{I}_3 & 2\mathbf{I}_3 & -\mathbf{w} & -2\mathbf{w} \\
        -2\mathbf{w}^{\top} & -\mathbf{w}^{\top} & 2\mathbf{w}^{\top}\mathbf{w} & \mathbf{w}^{\top}\mathbf{w} \\
        -\mathbf{w}^{\top} & -2\mathbf{w}^{\top} & \mathbf{w}^{\top}\mathbf{w} & 2\mathbf{w^{\top}w}\\
    \end{pmatrix} \nonumber\\
    & + 
    \frac{1}{6} \Delta u \rho 
    \begin{pmatrix}
        \mathbf{0} & \mathbf{0} & 
        -2 \frac{\partial^2 \mathbf{w}}{\partial t \partial \dot{u}_0} &
        -\frac{\partial^2 \mathbf{w}}{\partial t \partial \dot{u}_0} \\
        \mathbf{0} & \mathbf{0} & 
        - \frac{\partial^2 \mathbf{w}}{\partial t \partial \dot{u}_0} &
        -2\frac{\partial^2 \mathbf{w}}{\partial t \partial \dot{u}_0} \\
        -2 \frac{\partial^2 \mathbf{w}^{\top}}{\partial t \partial \dot{u}_0} &
        -\frac{\partial^2 \mathbf{w}^{\top}}{\partial t \partial \dot{u}_0} &
        2 \frac{\partial^2 \mathbf{w}^{\top}\mathbf{w}}{\partial t \partial \dot{u}_0} &
        \frac{\partial^2 \mathbf{w}^{\top}\mathbf{w}}{\partial t \partial \dot{u}_0} \\
        -\frac{\partial^2 \mathbf{w}^{\top}}{\partial t \partial \dot{u}_0} &
        -2\frac{\partial^2 \mathbf{w}^{\top}}{\partial t \partial \dot{u}_0} &
        \frac{\partial^2 \mathbf{w}^{\top}\mathbf{w}}{\partial t \partial \dot{u}_0} &
        2\frac{\partial^2 \mathbf{w}^{\top}\mathbf{w}}{\partial t \partial \dot{u}_0}
    \end{pmatrix}\nonumber
\end{align}

\begin{equation}
    \frac{\partial \dot{\mathbf{q}}_{0,1}}{\partial \dot{\mathbf{x}}_0^{(1)}} =
    \begin{pmatrix}
        1 \\ 0 \\ 0 \\ 0 \\ 0 \\ 0 \\ 0 \\ 0
    \end{pmatrix}
    \;\mbox{and}\;
    \frac{\partial \dot{\mathbf{q}}_{0,1}}{\partial u_0} =
    \begin{pmatrix}
        0 \\ 0 \\ 0 \\ 1 \\ 0 \\ 0 \\ 0 \\ 0
    \end{pmatrix}\nonumber
\end{equation}

\subsection{Inertia}
Kinetic energy is computed segment-wise, e.g. for a segment $[\mathbf{q_0, q_1}]$:
\begin{equation}
    T_{0,1}
    = \frac{1}{2} \dot{\mathbf{q}}_{0,1}^{\top} \mathbf{M}_{0,1} \dot{\mathbf{q}}_{0,1} 
    = \frac{1}{2} 
    \begin{pmatrix}
    \dot{\mathbf{x}}_0^{\top} & \dot{\mathbf{x}}_1^{\top} & \dot{u}_0 & \dot{u}_1
    \end{pmatrix}
    \mathbf{M}_{0,1}
    \begin{pmatrix}
    \dot{\mathbf{x}}_0 \\ \dot{\mathbf{x}}_1 \\ \dot{u}_0  \\ \dot{u}_1 
    \end{pmatrix}
\end{equation}
Its derivatives with respect to each node's position is the node's inertia: 
\begin{equation}
    \frac{\partial T_{0,1}}{\partial \mathbf{q}_{0,1}} =
    \begin{pmatrix}
        \mathbf{F}_{\mathbf{x}_0} \\
        \mathbf{F}_{\mathbf{x}_1} \\
        \mathbf{F}_{u_0} \\
        \mathbf{F}_{u_1}
    \end{pmatrix}
\end{equation}
\begin{align}
    \mathbf{F}_{\mathbf{x}_0} = \frac{\partial T_{0,1}}{\partial \mathbf{x}_0} = \frac{1}{2}\dot{\mathbf{q}}_{0,1}^{\top} \frac{\partial \mathbf{M}_{0,1}}{\partial \mathbf{x}_0} \dot{\mathbf{q}}_{0,1} \nonumber\\
    \mathbf{F}_{\mathbf{x}_1} = \frac{\partial T_{0,1}}{\partial \mathbf{x}_1} = \frac{1}{2}\dot{\mathbf{q}}_{0,1}^{\top} \frac{\partial \mathbf{M}_{0,1}}{\partial \mathbf{x}_1} \dot{\mathbf{q}}_{0,1} \nonumber\\
    F_{u_0} = \frac{\partial T_{0,1}}{\partial u_0} = \frac{1}{2}\dot{\mathbf{q}}_{0,1}^{\top} \frac{\partial \mathbf{M}_{0,1}}{\partial u_0} \dot{\mathbf{q}}_{0,1} \nonumber\\
    F_{u_1} = \frac{\partial T_{0,1}}{\partial u_1} = \frac{1}{2}\dot{\mathbf{q}}_{0,1}^{\top} \frac{\partial \mathbf{M}_{0,1}}{\partial u_1} \dot{\mathbf{q}}_{0,1}\nonumber
\end{align}
where $\mathbf{F}_{x_0}$ and $\mathbf{F}_{u_0}$ are the inertia of $\mathbf{q}_0$ in Lagrangian and Eulerian coordinates respectively. Similarly, $\mathbf{F}_{x_1}$ and $\mathbf{F}_{u_1}$ are the inertia of $\mathbf{q}_1$. The derivative of the forces with respect to positions are:
\begin{equation}
    \frac{\partial^2 T_{0,1}}{\partial \mathbf{q}_{0,1} \partial  \mathbf{q}_{0,1}} =
    \begin{pmatrix}
        \frac{\partial \mathbf{F}_{\mathbf{x}_0}}{\partial \mathbf{x}_0} &
        \frac{\partial \mathbf{F}_{\mathbf{x}_0}}{\partial \mathbf{x}_1} &
        \frac{\partial \mathbf{F}_{\mathbf{x}_0}}{\partial u_0} &
        \frac{\partial \mathbf{F}_{\mathbf{x}_0}}{\partial u_1} \\
        \frac{\partial \mathbf{F}_{\mathbf{x}_1}}{\partial \mathbf{x}_0} &
        \frac{\partial \mathbf{F}_{\mathbf{x}_1}}{\partial \mathbf{x}_1} &
        \frac{\partial \mathbf{F}_{\mathbf{x}_1}}{\partial u_0} &
        \frac{\partial \mathbf{F}_{\mathbf{x}_1}}{\partial u_1} \\
        \frac{\partial \mathbf{F}_{u_0}}{\partial \mathbf{x}_0} &
        \frac{\partial \mathbf{F}_{u_0}}{\partial \mathbf{x}_1} &
        \frac{\partial \mathbf{F}_{u_0}}{\partial u_0} &
        \frac{\partial \mathbf{F}_{u_0}}{\partial u_1}\\
        \frac{\partial \mathbf{F}_{u_1}}{\partial \mathbf{x}_0} &
        \frac{\partial \mathbf{F}_{u_1}}{\partial \mathbf{x}_1} &
        \frac{\partial \mathbf{F}_{u_1}}{\partial u_0} &
        \frac{\partial \mathbf{F}_{u_1}}{\partial u_1}\\
    \end{pmatrix}
\end{equation}
The derivative of the force in Lagrangian coordinates with respect to Lagrangian coordinates is
\begin{align}
    \frac{\partial \mathbf{F}_{\mathbf{x}_0}}{\partial \mathbf{x}_0} &= 
    \frac{1}{2}
    \dot{\mathbf{q}}_{0,1}^{\top} 
    \frac{\partial^2 \mathbf{M}_{0,1}}{\partial \mathbf{x}_0 \partial \mathbf{x}_0} \dot{\mathbf{q}}_{0,1} \nonumber \\
    & = \frac{1}{2}
    \begin{pmatrix}
        \dot{\mathbf{q}}_{0,1}^{\top} 
        \frac{\partial^2 \mathbf{M}_{0,1}}{\partial \mathbf{x}_0^{(1)} \partial \mathbf{x}_0^{(1)}} 
        \dot{\mathbf{q}}_{0,1} &
        \dot{\mathbf{q}}_{0,1}^{\top} 
        \frac{\partial^2 \mathbf{M}_{0,1}}{\partial \mathbf{x}_0^{(1)} \partial \mathbf{x}_0^{(2)}} 
        \dot{\mathbf{q}}_{0,1} &
        \dot{\mathbf{q}}_{0,1}^{\top} 
        \frac{\partial^2 \mathbf{M}_{0,1}}{\partial \mathbf{x}_0^{(1)} \partial \mathbf{x}_0^{(3)}} 
        \dot{\mathbf{q}}_{0,1} & \\
        \dot{\mathbf{q}}_{0,1}^{\top} 
        \frac{\partial^2 \mathbf{M}_{0,1}}{\partial \mathbf{x}_0^{(2)} \partial \mathbf{x}_0^{(1)}} 
        \dot{\mathbf{q}}_{0,1} &
        \dot{\mathbf{q}}_{0,1}^{\top} 
        \frac{\partial^2 \mathbf{M}_{0,1}}{\partial \mathbf{x}_0^{(2)} \partial \mathbf{x}_0^{(2)}} 
        \dot{\mathbf{q}}_{0,1} &
        \dot{\mathbf{q}}_{0,1}^{\top} 
        \frac{\partial^2 \mathbf{M}_{0,1}}{\partial \mathbf{x}_0^{(2)} \partial \mathbf{x}_0^{(3)}} 
        \dot{\mathbf{q}}_{0,1} & \\
        \dot{\mathbf{q}}_{0,1}^{\top} 
        \frac{\partial^2 \mathbf{M}_{0,1}}{\partial \mathbf{x}_0^{(3)} \partial \mathbf{x}_0^{(1)}} 
        \dot{\mathbf{q}}_{0,1} &
        \dot{\mathbf{q}}_{0,1}^{\top} 
        \frac{\partial^2 \mathbf{M}_{0,1}}{\partial \mathbf{x}_0^{(3)} \partial \mathbf{x}_0^{(2)}} 
        \dot{\mathbf{q}}_{0,1} &
        \dot{\mathbf{q}}_{0,1}^{\top} 
        \frac{\partial^2 \mathbf{M}_{0,1}}{\partial \mathbf{x}_0^{(3)} \partial \mathbf{x}_0^{(3)}} 
        \dot{\mathbf{q}}_{0,1} & \\
    \end{pmatrix} \nonumber
\end{align}
$\frac{\partial \mathbf{F}_{\mathbf{x}_0}}{\partial \mathbf{x}_1}$, $\frac{\partial \mathbf{F}_{\mathbf{x}_1}}{\partial \mathbf{x}_0}$ and $\frac{\partial \mathbf{F}_{\mathbf{x}_1}}{\partial \mathbf{x}_1}$ are in similar forms as $\frac{\partial \mathbf{F}_{\mathbf{x}_0}}{\partial \mathbf{x}_0}$. Also, the derivative of the force in Lagrangian coordinate with respect to Eulerian coordinates is:

\begin{equation}
    \frac{\partial \mathbf{F}_{\mathbf{x}_0}}{\partial u_0} = 
    \frac{1}{2}
    \dot{\mathbf{q}}_{0,1}^{\top} 
    \frac{\partial^2 \mathbf{M}_{0,1}}{\partial \mathbf{x}_0 \partial u_0} 
    \dot{\mathbf{q}}_{0,1} \nonumber
\end{equation}
$\frac{\partial \mathbf{F}_{\mathbf{x}_0}}{\partial u_1}$, $\frac{\partial \mathbf{F}_{\mathbf{x}_1}}{\partial u_0}$ and $\frac{\partial \mathbf{F}_{\mathbf{x}_1}}{\partial u_1}$ are in similar forms as $\frac{\partial \mathbf{F}_{\mathbf{x}_0}}{\partial u_0}$. Correspondingly, the derivative of the force in Eulerian coordinates with respect to Lagrangian coordinates is:



\begin{equation}
    \frac{\partial \mathbf{F}_{u_0}}{\partial \mathbf{x}_0} = 
    \frac{1}{2}
    \dot{\mathbf{q}}_{0,1}^{\top} 
    \frac{\partial^2 \mathbf{M}_{0,1}}{\partial u_0 \partial \mathbf{x}_0} 
    \dot{\mathbf{q}}_{0,1} \nonumber
\end{equation}
$\frac{\partial \mathbf{F}_{u_0}}{\partial \mathbf{x}_1}$, $\frac{\partial \mathbf{F}_{u_1}}{\partial \mathbf{x}_0}$ and $\frac{\partial \mathbf{F}_{u_1}}{\partial \mathbf{x}_1}$ are in similar forms as $\frac{\partial \mathbf{F}_{u_0}}{\partial \mathbf{x}_0}$. The derivative of the force in Eulerian coordinates with respect to Eulerian coordinates is:



\begin{equation}
    \frac{\partial \mathbf{F}_{u_0}}{\partial u_0} = 
    \frac{1}{2}
    \dot{\mathbf{q}}_{0,1}^{\top} 
    \frac{\partial^2 \mathbf{M}_{0,1}}{\partial u_0 \partial u_0} 
    \dot{\mathbf{q}}_{0,1} \nonumber
\end{equation}
$\frac{\partial \mathbf{F}_{u_0}}{\partial u_1}$, $\frac{\partial \mathbf{F}_{u_1}}{\partial u_0}$ and $\frac{\partial \mathbf{F}_{u_1}}{\partial u_1}$ are in similar forms as $\frac{\partial \mathbf{F}_{u_0}}{\partial u_0}$.



Specially, the entries in $\frac{\partial \mathbf{F}_{\mathbf{x}_0}}{\partial \mathbf{x}_0}$ are:
\begin{equation}
    \frac{\partial^2 \mathbf{M}_{0,1} }{\partial \mathbf{x}_0^{(1)} \partial \mathbf{x}_0^{(1)}}
    = \frac{1}{6} \Delta u \rho
    \begin{pmatrix}
    \mathbf{0} & \mathbf{0} & \mathbf{0} & \mathbf{0} \\
    \mathbf{0} & \mathbf{0} & \mathbf{0} & \mathbf{0} \\
    \mathbf{0} & \mathbf{0} &
    2 \frac{\partial^2 \mathbf{w}^{\top}\mathbf{w}}{\partial \mathbf{x}_0^{(1)} \partial \mathbf{x}_0^{(1)}} & 
      \frac{\partial^2 \mathbf{w}^{\top}\mathbf{w}}{\partial \mathbf{x}_0^{(1)} \partial \mathbf{x}_0^{(1)}} \\
    \mathbf{0} & \mathbf{0} &
      \frac{\partial^2 \mathbf{w}^{\top}\mathbf{w}}{\partial \mathbf{x}_0^{(1)} \partial \mathbf{x}_0^{(1)}} & 
    2 \frac{\partial^2 \mathbf{w}^{\top}\mathbf{w}}{\partial \mathbf{x}_0^{(1)} \partial \mathbf{x}_0^{(1)}}
    \end{pmatrix} \nonumber
\end{equation}

\begin{equation}
    \frac{\partial^2 \mathbf{M}_{0,1} }{\partial \mathbf{x}_0^{(2)} \partial \mathbf{x}_0^{(2)}}
    = \frac{1}{6} \Delta u \rho
    \begin{pmatrix}
    \mathbf{0} & \mathbf{0} & \mathbf{0} & \mathbf{0} \\
    \mathbf{0} & \mathbf{0} & \mathbf{0} & \mathbf{0} \\
    \mathbf{0} & \mathbf{0} &
    2 \frac{\partial^2 \mathbf{w}^{\top}\mathbf{w}}{\partial \mathbf{x}_0^{(2)} \partial \mathbf{x}_0^{(2)}} & 
      \frac{\partial^2 \mathbf{w}^{\top}\mathbf{w}}{\partial \mathbf{x}_0^{(2)} \partial \mathbf{x}_0^{(2)}} \\
    \mathbf{0} & \mathbf{0} &
      \frac{\partial^2 \mathbf{w}^{\top}\mathbf{w}}{\partial \mathbf{x}_0^{(2)} \partial \mathbf{x}_0^{(2)}} & 
    2 \frac{\partial^2 \mathbf{w}^{\top}\mathbf{w}}{\partial \mathbf{x}_0^{(2)} \partial \mathbf{x}_0^{(2)}}
    \end{pmatrix}\nonumber
\end{equation}

\begin{equation}
    \frac{\partial^2 \mathbf{M}_{0,1} }{\partial \mathbf{x}_0^{(3)} \partial \mathbf{x}_0^{(3)}}
    = \frac{1}{6} \Delta u \rho
    \begin{pmatrix}
    \mathbf{0} & \mathbf{0} & \mathbf{0} & \mathbf{0} \\
    \mathbf{0} & \mathbf{0} & \mathbf{0} & \mathbf{0} \\
    \mathbf{0} & \mathbf{0} &
    2 \frac{\partial^2 \mathbf{w}^{\top}\mathbf{w}}{\partial \mathbf{x}_0^{(3)} \partial \mathbf{x}_0^{(3)}} & 
      \frac{\partial^2 \mathbf{w}^{\top}\mathbf{w}}{\partial \mathbf{x}_0^{(3)} \partial \mathbf{x}_0^{(3)}} \\
    \mathbf{0} & \mathbf{0} &
      \frac{\partial^2 \mathbf{w}^{\top}\mathbf{w}}{\partial \mathbf{x}_0^{(3)} \partial \mathbf{x}_0^{(3)}} & 
    2 \frac{\partial^2 \mathbf{w}^{\top}\mathbf{w}}{\partial \mathbf{x}_0^{(3)} \partial \mathbf{x}_0^{(3)}}
    \end{pmatrix}\nonumber
\end{equation}
where
\begin{equation}
    \frac{\partial^2 \mathbf{w}^{\top}\mathbf{w}}{\partial \mathbf{x}_0^{(1)} \partial \mathbf{x}_0^{(1)}} = 
    \frac{2}{\Delta u^2}
    \;\mbox{,}\;
    \frac{\partial^2 \mathbf{w}^{\top}\mathbf{w}}{\partial \mathbf{x}_0^{(2)} \partial \mathbf{x}_0^{(2)}} = 
    \frac{2}{\Delta u^2}
    \;\mbox{,and}\;
    \frac{\partial^2 \mathbf{w}^{\top}\mathbf{w}}{\partial \mathbf{x}_0^{(3)} \partial \mathbf{x}_0^{(3)}} = 
    \frac{2}{\Delta u^2}\nonumber
\end{equation}
The other components are
\begin{equation}
    \frac{\partial^2 \mathbf{M}_{0,1} }{\partial \mathbf{x}_0^{(1)} \partial \mathbf{x}_0^{(2)}} =
    \frac{\partial^2 \mathbf{M}_{0,1} }{\partial \mathbf{x}_0^{(1)} \partial \mathbf{x}_0^{(3)}} = \mathbf{0}\nonumber
\end{equation}

\begin{equation}
    \frac{\partial^2 \mathbf{M}_{0,1} }{\partial \mathbf{x}_0^{(2)} \partial \mathbf{x}_0^{(1)}} =
    \frac{\partial^2 \mathbf{M}_{0,1} }{\partial \mathbf{x}_0^{(2)} \partial \mathbf{x}_0^{(3)}} = \mathbf{0}\nonumber
\end{equation}

\begin{equation}
    \frac{\partial^2 \mathbf{M}_{0,1} }{\partial \mathbf{x}_0^{(3)} \partial \mathbf{x}_0^{(1)}} =
    \frac{\partial^2 \mathbf{M}_{0,1} }{\partial \mathbf{x}_0^{(3)} \partial \mathbf{x}_0^{(2)}} = \mathbf{0}\nonumber
\end{equation}

Moreover, as
\begin{equation}
    \frac{\partial^2 \mathbf{w}^{\top}\mathbf{w}}{\partial \mathbf{x}_0^{(1)} \partial \mathbf{x}_1^{(1)}} = 
    -\frac{2}{\Delta u^2}
    \;\mbox{,}\;
    \frac{\partial^2 \mathbf{w}^{\top}\mathbf{w}}{\partial \mathbf{x}_0^{(2)} \partial \mathbf{x}_1^{(2)}} = 
    -\frac{2}{\Delta u^2}
    \;\mbox{,and}\;
    \frac{\partial^2 \mathbf{w}^{\top}\mathbf{w}}{\partial \mathbf{x}_0^{(3)} \partial \mathbf{x}_1^{(3)}} = 
    -\frac{2}{\Delta u^2}\nonumber
\end{equation}

\begin{equation}
    \frac{\partial^2 \mathbf{M}_{0,1} }{\partial \mathbf{x}_0^{(1)} \partial \mathbf{x}_1^{(2)}} =
    \frac{\partial^2 \mathbf{M}_{0,1} }{\partial \mathbf{x}_0^{(1)} \partial \mathbf{x}_1^{(3)}} = \mathbf{0}\nonumber
\end{equation}

\begin{equation}
    \frac{\partial^2 \mathbf{M}_{0,1} }{\partial \mathbf{x}_0^{(2)} \partial \mathbf{x}_1^{(1)}} =
    \frac{\partial^2 \mathbf{M}_{0,1} }{\partial \mathbf{x}_0^{(2)} \partial \mathbf{x}_1^{(3)}} = \mathbf{0}\nonumber
\end{equation}

\begin{equation}
    \frac{\partial^2 \mathbf{M}_{0,1} }{\partial \mathbf{x}_0^{(3)} \partial \mathbf{x}_1^{(1)}} =
    \frac{\partial^2 \mathbf{M}_{0,1} }{\partial \mathbf{x}_0^{(3)} \partial \mathbf{x}_1^{(2)}} = \mathbf{0}\nonumber
\end{equation}
We can find that 
\begin{equation}
    \frac{\partial^2 \mathbf{M}_{0,1} }{\partial \mathbf{x}_0^{(1)} \partial \mathbf{x}_0^{(1)}} =
    - \frac{\partial^2 \mathbf{M}_{0,1} }{\partial \mathbf{x}_0^{(1)} \partial \mathbf{x}_1^{(1)}}\nonumber
\end{equation}

\begin{equation}
    \frac{\partial^2 \mathbf{M}_{0,1} }{\partial \mathbf{x}_0^{(2)} \partial \mathbf{x}_0^{(2)}} =
    - \frac{\partial^2 \mathbf{M}_{0,1} }{\partial \mathbf{x}_0^{(2)} \partial \mathbf{x}_1^{(2)}}\nonumber
\end{equation}

\begin{equation}
    \frac{\partial^2 \mathbf{M}_{0,1} }{\partial \mathbf{x}_0^{(3)} \partial \mathbf{x}_0^{(3)}} =
    - \frac{\partial^2 \mathbf{M}_{0,1} }{\partial \mathbf{x}_0^{(3)} \partial \mathbf{x}_1^{(3)}}\nonumber
\end{equation}
Therefore, 
\begin{equation}
    \frac{\partial \mathbf{F}_{\mathbf{x}_0}}{\partial \mathbf{x}_0}
    = -\frac{\partial \mathbf{F}_{\mathbf{x}_0}}{\partial \mathbf{x}_1}\nonumber
\end{equation}

\begin{equation}
    \frac{\partial \mathbf{F}_{\mathbf{x}_1}}{\partial \mathbf{x}_1}
    = -\frac{\partial \mathbf{F}_{\mathbf{x}_1}}{\partial \mathbf{x}_0}
    = \frac{\partial \mathbf{F}_{\mathbf{x}_0}}{\partial \mathbf{x}_0}\nonumber
\end{equation}
To compute the derivatives of the forces in Lagrangian coordinates with respect to Eulerian coordinates, we need to compute:
\begin{equation}
    \frac{\partial^2 \mathbf{M}_{0,1}}{\partial \mathbf{x}_0 \partial u_0}
    = \begin{pmatrix}
        \frac{\partial^2 \mathbf{M}_{0,1}}{\partial \mathbf{x}_0^{(1)} \partial u_0} \\
        \frac{\partial^2 \mathbf{M}_{0,1}}{\partial \mathbf{x}_0^{(2)} \partial u_0} \\
        \frac{\partial^2 \mathbf{M}_{0,1}}{\partial \mathbf{x}_0^{(3)} \partial u_0}
    \end{pmatrix}
    \;\mbox{and}\;
    \frac{\partial^2 \mathbf{M}_{0,1}}{\partial \mathbf{x}_0 \partial u_1}
    = \begin{pmatrix}
        \frac{\partial^2 \mathbf{M}_{0,1}}{\partial \mathbf{x}_0^{(1)} \partial u_1} \\
        \frac{\partial^2 \mathbf{M}_{0,1}}{\partial \mathbf{x}_0^{(2)} \partial u_1} \\
        \frac{\partial^2 \mathbf{M}_{0,1}}{\partial \mathbf{x}_0^{(3)} \partial u_1}
    \end{pmatrix}\nonumber
\end{equation}

\begin{equation}
    \frac{\partial^2 \mathbf{M}_{0,1}}{\partial \mathbf{x}_1 \partial u_0}
    = \begin{pmatrix}
        \frac{\partial^2 \mathbf{M}_{0,1}}{\partial \mathbf{x}_1^{(1)} \partial u_0} \\
        \frac{\partial^2 \mathbf{M}_{0,1}}{\partial \mathbf{x}_1^{(2)} \partial u_0} \\
        \frac{\partial^2 \mathbf{M}_{0,1}}{\partial \mathbf{x}_1^{(3)} \partial u_0}
    \end{pmatrix}
    \;\mbox{and}\;
    \frac{\partial^2 \mathbf{M}_{0,1}}{\partial \mathbf{x}_1 \partial u_1}
    = \begin{pmatrix}
        \frac{\partial^2 \mathbf{M}_{0,1}}{\partial \mathbf{x}_1^{(1)} \partial u_1} \\
        \frac{\partial^2 \mathbf{M}_{0,1}}{\partial \mathbf{x}_1^{(2)} \partial u_1} \\
        \frac{\partial^2 \mathbf{M}_{0,1}}{\partial \mathbf{x}_1^{(3)} \partial u_1}
    \end{pmatrix}\nonumber
\end{equation}
Take one component $\frac{\partial^2 \mathbf{M}_{0,1}}{\partial \mathbf{x}_0^{(1)} \partial u_0}$ as example:
\begin{align}
    \frac{\partial^2 \mathbf{M}_{0,1}}{\partial \mathbf{x}_0^{(1)} \partial u_0}
    &= - \frac{1}{6}
    \begin{pmatrix}
        \mathbf{0} & \mathbf{0} & 
        -2 \frac{\partial \mathbf{w}}{\partial \mathbf{x}_0^{(1)}} & 
        -\frac{\partial \mathbf{w}}{\partial \mathbf{x}_0^{(1)}} \\
        \mathbf{0} & \mathbf{0} & 
        - \frac{\partial \mathbf{w}}{\partial \mathbf{x}_0^{(1)}} & 
        -2\frac{\partial \mathbf{w}}{\partial \mathbf{x}_0^{(1)}} \\
        -2 \frac{\partial \mathbf{w}^{\top}}{\partial \mathbf{x}_0^{(1)}} & 
        -\frac{\partial \mathbf{w}^{\top}}{\partial \mathbf{x}_0^{(1)}} & 
        2 \frac{\partial \mathbf{w}^{\top}\mathbf{w}}{\partial \mathbf{x}_0^{(1)}} & 
        \frac{\partial \mathbf{w}^{\top}\mathbf{w}}{\partial \mathbf{x}_0^{(1)}} \\
        -\frac{\partial \mathbf{w}^{\top}}{\partial \mathbf{x}_0^{(1)}} & 
        -2 \frac{\partial \mathbf{w}^{\top}}{\partial \mathbf{x}_0^{(1)}} & 
        \frac{\partial \mathbf{w}^{\top}\mathbf{w}}{\partial \mathbf{x}_0^{(1)}} & 
        2 \frac{\partial \mathbf{w}^{\top}\mathbf{w}}{\partial \mathbf{x}_0^{(1)}}
    \end{pmatrix}\nonumber\\
    & + \frac{1}{6} \Delta u \rho
    \begin{pmatrix}
    \mathbf{0} & \mathbf{0} & 
    -2 \frac{\partial^2 \mathbf{w}}{\partial \mathbf{x}_0^{(1)} \partial u_0} & 
    -\frac{\partial^2 \mathbf{w}}{\partial \mathbf{x}_0^{(1)} \partial u_0} \\
    \mathbf{0} & \mathbf{0} & 
    - \frac{\partial^2 \mathbf{w}}{\partial \mathbf{x}_0^{(1)} \partial u_0} & 
    -2\frac{\partial^2 \mathbf{w}}{\partial \mathbf{x}_0^{(1)} \partial u_0} \\
    -2 \frac{\partial^2 \mathbf{w}^{\top}}{\partial \mathbf{x}_0^{(1)} \partial u_0} & 
    -\frac{\partial^2 \mathbf{w}^{\top}}{\partial \mathbf{x}_0^{(1)} \partial u_0} & 
    2 \frac{\partial^2 \mathbf{w}^{\top}\mathbf{w}}{\partial \mathbf{x}_0^{(1)} \partial u_0} & 
    \frac{\partial^2 \mathbf{w}^{\top}\mathbf{w}}{\partial \mathbf{x}_0^{(1)} \partial u_0} \\
    -\frac{\partial^2 \mathbf{w}^{\top}}{\partial \mathbf{x}_0^{(1)} \partial u_0} & 
    -2 \frac{\partial^2 \mathbf{w}^{\top}}{\partial \mathbf{x}_0^{(1)} \partial u_0} & 
    \frac{\partial^2 \mathbf{w}^{\top}\mathbf{w}}{\partial \mathbf{x}_0^{(1)} \partial u_0} & 
    2 \frac{\partial^2 \mathbf{w}^{\top}\mathbf{w}}{\partial \mathbf{x}_0^{(1)} \partial u_0}
    \end{pmatrix}\nonumber
\end{align}
in which 
\begin{equation}
    \frac{\partial^2 \mathbf{w}}{\partial \mathbf{x}_0^{(1)} \partial u_0} = 
    -\begin{pmatrix}
        \frac{1}{\Delta u^2} \\ 0 \\ 0
    \end{pmatrix}
    \;\mbox{,}\;
    \frac{\partial^2 \mathbf{w}}{\partial \mathbf{x}_0^{(2)} \partial u_0} = 
    -\begin{pmatrix}
        0 \\ \frac{1}{\Delta u^2} \\ 0
    \end{pmatrix}
    \;\mbox{,}\;
    \frac{\partial^2 \mathbf{w}}{\partial \mathbf{x}_0^{(3)} \partial u_0} = 
    -\begin{pmatrix}
        0 \\ 0 \\ \frac{1}{\Delta u^2}
    \end{pmatrix}\nonumber
\end{equation}

\begin{equation}
    \frac{\partial^2 \mathbf{w}}{\partial \mathbf{x}_0^{(1)} \partial u_1} = 
    \begin{pmatrix}
        \frac{1}{\Delta u^2} \\ 0 \\ 0
    \end{pmatrix}
    \;\mbox{,}\;
    \frac{\partial^2 \mathbf{w}}{\partial \mathbf{x}_0^{(2)} \partial u_1} = 
    \begin{pmatrix}
        0 \\ \frac{1}{\Delta u^2} \\ 0
    \end{pmatrix}
    \;\mbox{,}\;
    \frac{\partial^2 \mathbf{w}}{\partial \mathbf{x}_0^{(3)} \partial u_1} = 
    \begin{pmatrix}
        0 \\ 0 \\ \frac{1}{\Delta u^2}
    \end{pmatrix}\nonumber
\end{equation}

\begin{equation}
    \frac{\partial^2 \mathbf{w}}{\partial \mathbf{x}_1^{(1)} \partial u_0} = 
    -\begin{pmatrix}
        \frac{1}{\Delta u^2} \\ 0 \\ 0
    \end{pmatrix}
    \;\mbox{,}\;
    \frac{\partial^2 \mathbf{w}}{\partial \mathbf{x}_1^{(2)} \partial u_0} = 
    -\begin{pmatrix}
        0 \\ \frac{1}{\Delta u^2} \\ 0
    \end{pmatrix}
    \;\mbox{,}\;
    \frac{\partial^2 \mathbf{w}}{\partial \mathbf{x}_1^{(3)} \partial u_0} = 
    -\begin{pmatrix}
        0 \\ 0 \\ \frac{1}{\Delta u^2}
    \end{pmatrix}\nonumber
\end{equation}

\begin{equation}
    \frac{\partial^2 \mathbf{w}}{\partial \mathbf{x}_1^{(1)} \partial u_0} = 
    \begin{pmatrix}
        \frac{1}{\Delta u^2} \\ 0 \\ 0
    \end{pmatrix}
    \;\mbox{,}\;
    \frac{\partial^2 \mathbf{w}}{\partial \mathbf{x}_1^{(2)} \partial u_0} = 
    \begin{pmatrix}
        0 \\ \frac{1}{\Delta u^2} \\ 0
    \end{pmatrix}
    \;\mbox{,}\;
    \frac{\partial^2 \mathbf{w}}{\partial \mathbf{x}_1^{(3)} \partial u_0} = 
    \begin{pmatrix}
        0 \\ 0 \\ \frac{1}{\Delta u^2}
    \end{pmatrix}\nonumber
\end{equation}
It should be easy to compute 
$\frac{\partial^2 \mathbf{M}_{0,1}}{\partial \mathbf{x}_0 \partial u_0}$,
$\frac{\partial^2 \mathbf{M}_{0,1}}{\partial \mathbf{x}_0 \partial u_1}$,
$\frac{\partial^2 \mathbf{M}_{0,1}}{\partial \mathbf{x}_1 \partial u_0}$,
$\frac{\partial^2 \mathbf{M}_{0,1}}{\partial \mathbf{x}_1 \partial u_1}$ and then compute 
$\frac{\partial \mathbf{F}_{\mathbf{x}_0}}{\partial u_0}$,
$\frac{\partial \mathbf{F}_{\mathbf{x}_0}}{\partial u_1}$,
$\frac{\partial \mathbf{F}_{\mathbf{x}_1}}{\partial u_0}$,
$\frac{\partial \mathbf{F}_{\mathbf{x}_1}}{\partial u_1}$.
The derivatives of the forces in Eulerain coordinates with respect to Lagrangian coordinates are the transpose of the forces in Lagrangian coordinates with respect to Eulerian coordinates:
\begin{equation}
    \frac{\partial \mathbf{F}_{u_0}}{\partial {\mathbf{x}_0}} =
    \left(\frac{\partial \mathbf{F}_{\mathbf{x}_0}}{\partial u_0}\right)^{\top}
    \quad
    \frac{\partial \mathbf{F}_{u_0}}{\partial {\mathbf{x}_1}} =
    \left(\frac{\partial \mathbf{F}_{\mathbf{x}_1}}{\partial u_0}\right)^{\top}\nonumber
\end{equation}

\begin{equation}
    \frac{\partial \mathbf{F}_{u_1}}{\partial {\mathbf{x}_0}} =
    \left(\frac{\partial \mathbf{F}_{\mathbf{x}_0}}{\partial u_1}\right)^{\top}
    \quad
    \frac{\partial \mathbf{F}_{u_1}}{\partial {\mathbf{x}_1}} =
    \left(\frac{\partial \mathbf{F}_{\mathbf{x}_1}}{\partial u_1}\right)^{\top}\nonumber
\end{equation}
To compute the derivatives of the forces in Eluerian coordinates with respect to Eulerian coordinates, we need to compute:
$\frac{\partial^2 \mathbf{M}_{0,1}}{\partial u_0 \partial u_0}$,
$\frac{\partial^2 \mathbf{M}_{0,1}}{\partial u_0 \partial u_1}$,
$\frac{\partial^2 \mathbf{M}_{0,1}}{\partial u_1 \partial u_0}$, and
$\frac{\partial^2 \mathbf{M}_{0,1}}{\partial u_1 \partial u_1}$. 
For example, 
\begin{align}
    \frac{\partial^2 \mathbf{M}_{0,1}}{\partial u_0 \partial u_0} = 
    &-\frac{1}{6} \rho
    \begin{pmatrix}
        \mathbf{0} & \mathbf{0} & -2 \frac{\partial \mathbf{w}}{\partial u_0} & -\frac{\partial \mathbf{w}}{\partial u_0} \\
        \mathbf{0} & \mathbf{0} & - \frac{\partial \mathbf{w}}{\partial u_0} & -2\frac{\partial \mathbf{w}}{\partial u_0} \\
        -2 \frac{\partial \mathbf{w}^{\top}}{\partial u_0} & -\frac{\partial \mathbf{w}^{\top}}{\partial u_0} & 2 \frac{\partial \mathbf{w}^{\top}\mathbf{w}}{\partial u_0} & \frac{\partial \mathbf{w}^{\top}\mathbf{w}}{\partial u_0} \\
        -\frac{\partial \mathbf{w}^{\top}}{\partial u_0} & -2 \frac{\partial \mathbf{w}^{\top}}{\partial u_0} & \frac{\partial \mathbf{w}^{\top}\mathbf{w}}{\partial u_0} & 2 \frac{\partial \mathbf{w}^{\top}\mathbf{w}}{\partial u_0}
    \end{pmatrix} \nonumber\\
    &-\frac{1}{6} \rho
    \begin{pmatrix}
        \mathbf{0} & \mathbf{0} & -2 \frac{\partial \mathbf{w}}{\partial u_0} & -\frac{\partial \mathbf{w}}{\partial u_0} \\
        \mathbf{0} & \mathbf{0} & - \frac{\partial \mathbf{w}}{\partial u_0} & -2\frac{\partial \mathbf{w}}{\partial u_0} \\
        -2 \frac{\partial \mathbf{w}^{\top}}{\partial u_0} & -\frac{\partial \mathbf{w}^{\top}}{\partial u_0} & 2 \frac{\partial \mathbf{w}^{\top}\mathbf{w}}{\partial u_0} & \frac{\partial \mathbf{w}^{\top}\mathbf{w}}{\partial u_0} \\
        -\frac{\partial \mathbf{w}^{\top}}{\partial u_0} & -2 \frac{\partial \mathbf{w}^{\top}}{\partial u_0} & \frac{\partial \mathbf{w}^{\top}\mathbf{w}}{\partial u_0} & 2 \frac{\partial \mathbf{w}^{\top}\mathbf{w}}{\partial u_0}
    \end{pmatrix} \nonumber\\
    &+ \frac{1}{6} \Delta u \rho
    \begin{pmatrix}
        \mathbf{0} & \mathbf{0} & 
        -2 \frac{\partial^2 \mathbf{w}}{\partial u_0 \partial u_0} & 
        -\frac{\partial^2 \mathbf{w}}{\partial u_0 \partial u_0} \\
        \mathbf{0} & \mathbf{0} & 
        - \frac{\partial^2 \mathbf{w}}{\partial u_0 \partial u_0} &
        -2\frac{\partial^2 \mathbf{w}}{\partial u_0 \partial u_0} \\
        -2 \frac{\partial^2 \mathbf{w}^{\top}}{\partial u_0 \partial u_0} & 
        -\frac{\partial^2 \mathbf{w}^{\top}}{\partial u_0 \partial u_0} &
        2 \frac{\partial^2 \mathbf{w}^{\top}\mathbf{w}}{\partial u_0 \partial u_0} & 
        \frac{\partial^2 \mathbf{w}^{\top}\mathbf{w}}{\partial u_0 \partial u_0} \\
        -\frac{\partial^2 \mathbf{w}^{\top}}{\partial u_0 \partial u_0} & 
        -2 \frac{\partial^2 \mathbf{w}^{\top}}{\partial u_0 \partial u_0} & 
        \frac{\partial^2 \mathbf{w}^{\top}\mathbf{w}}{\partial u_0 \partial u_0} & 
        2 \frac{\partial^2 \mathbf{w}^{\top}\mathbf{w}}{\partial u_0 \partial u_0}
    \end{pmatrix}\nonumber
\end{align}
in which
\begin{equation}
    \frac{\partial^2 \mathbf{w}}{\partial u_0 \partial u_0} = 
    2\frac{\mathbf{w}}{\Delta u^2}
    \;\mbox{and}\;
    \frac{\partial^2 \mathbf{w}^{\top}\mathbf{w}}{\partial u_0 \partial u_0} = 
    6\frac{\mathbf{w}^{\top}\mathbf{w}}{\Delta u^2}\nonumber
\end{equation}
Similarly, we can compute
\begin{equation}
    \frac{\partial^2 \mathbf{w}}{\partial u_0 \partial u_1} = 
    -2\frac{\mathbf{w}}{\Delta u^2}
    \quad
    \frac{\partial^2 \mathbf{w}^{\top}\mathbf{w}}{\partial u_0 \partial u_1} = 
    -6\frac{\mathbf{w}^{\top}\mathbf{w}}{\Delta u^2}\nonumber
\end{equation}

\begin{equation}
    \frac{\partial^2 \mathbf{w}}{\partial u_1 \partial u_0} = 
    -2\frac{\mathbf{w}}{\Delta u^2}
    \quad
    \frac{\partial^2 \mathbf{w}^{\top}\mathbf{w}}{\partial u_1 \partial u_0} = 
    -6\frac{\mathbf{w}^{\top}\mathbf{w}}{\Delta u^2}\nonumber
\end{equation}

\begin{equation}
    \frac{\partial^2 \mathbf{w}}{\partial u_1 \partial u_1} = 
    2\frac{\mathbf{w}}{\Delta u^2}
    \quad
    \frac{\partial^2 \mathbf{w}^{\top}\mathbf{w}}{\partial u_1 \partial u_1} = 
    6\frac{\mathbf{w}^{\top}\mathbf{w}}{\Delta u^2}\nonumber
\end{equation}
to compute $\frac{\partial^2 \mathbf{M}_{0,1}}{\partial u_0 \partial u_1}$,
$\frac{\partial^2 \mathbf{M}_{0,1}}{\partial u_1 \partial u_0}$, and
$\frac{\partial^2 \mathbf{M}_{0,1}}{\partial u_1 \partial u_1}$. The derivatives of the inertia with respect to nodes' velocities are:
\begin{align}
    \frac{\partial \mathbf{F}_{\mathbf{x}_0}}{\partial \dot{\mathbf{x}}_0} =
    \begin{pmatrix}
        \frac{1}{2}
        \frac{\partial \dot{\mathbf{q}}_{0,1}^{\top} }{\partial \dot{\mathbf{x}}_0^{(1)}}
        \frac{\partial \mathbf{M}_{0,1}}{\partial \mathbf{x}_0} 
        \dot{\mathbf{q}}_{0,1} +
        \frac{1}{2}
        \dot{\mathbf{q}}_{0,1}^{\top}
        \frac{\partial \mathbf{M}_{0,1}}{\partial \mathbf{x}_0} 
        \frac{\partial \dot{\mathbf{q}}_{0,1} }{\partial \dot{\mathbf{x}}_0^{(1)}} \\
        \frac{1}{2}
        \frac{\partial \dot{\mathbf{q}}_{0,1}^{\top} }{\partial \dot{\mathbf{x}}_0^{(2)}}
        \frac{\partial \mathbf{M}_{0,1}}{\partial \mathbf{x}_0} 
        \dot{\mathbf{q}}_{0,1} +
        \frac{1}{2}
        \dot{\mathbf{q}}_{0,1}^{\top}
        \frac{\partial \mathbf{M}_{0,1}}{\partial \mathbf{x}_0} 
        \frac{\partial \dot{\mathbf{q}}_{0,1} }{\partial \dot{\mathbf{x}}_0^{(2)}} \\
        \frac{1}{2}
        \frac{\partial \dot{\mathbf{q}}_{0,1}^{\top} }{\partial \dot{\mathbf{x}}_0^{(3)}}
        \frac{\partial \mathbf{M}_{0,1}}{\partial \mathbf{x}_0} 
        \dot{\mathbf{q}}_{0,1} +
        \frac{1}{2}
        \dot{\mathbf{q}}_{0,1}^{\top}
        \frac{\partial \mathbf{M}_{0,1}}{\partial \mathbf{x}_0} 
        \frac{\partial \dot{\mathbf{q}}_{0,1} }{\partial \dot{\mathbf{x}}_0^{(3)}} \\
    \end{pmatrix}\nonumber
\end{align}

\begin{align}
    \frac{\partial \mathbf{F}_{\mathbf{x}_0}}{\partial \dot{\mathbf{x}}_1} =
    \begin{pmatrix}
        \frac{1}{2}
        \frac{\partial \dot{\mathbf{q}}_{0,1}^{\top} }{\partial \dot{\mathbf{x}}_1^{(1)}}
        \frac{\partial \mathbf{M}_{0,1}}{\partial \mathbf{x}_0} 
        \dot{\mathbf{q}}_{0,1} +
        \frac{1}{2}
        \dot{\mathbf{q}}_{0,1}^{\top}
        \frac{\partial \mathbf{M}_{0,1}}{\partial \mathbf{x}_0} 
        \frac{\partial \dot{\mathbf{q}}_{0,1} }{\partial \dot{\mathbf{x}}_1^{(1)}} \\
        \frac{1}{2}
        \frac{\partial \dot{\mathbf{q}}_{0,1}^{\top} }{\partial \dot{\mathbf{x}}_1^{(2)}}
        \frac{\partial \mathbf{M}_{0,1}}{\partial \mathbf{x}_0} 
        \dot{\mathbf{q}}_{0,1} +
        \frac{1}{2}
        \dot{\mathbf{q}}_{0,1}^{\top}
        \frac{\partial \mathbf{M}_{0,1}}{\partial \mathbf{x}_0} 
        \frac{\partial \dot{\mathbf{q}}_{0,1} }{\partial \dot{\mathbf{x}}_1^{(2)}} \\
        \frac{1}{2}
        \frac{\partial \dot{\mathbf{q}}_{0,1}^{\top} }{\partial \dot{\mathbf{x}}_1^{(3)}}
        \frac{\partial \mathbf{M}_{0,1}}{\partial \mathbf{x}_0} 
        \dot{\mathbf{q}}_{0,1} +
        \frac{1}{2}
        \dot{\mathbf{q}}_{0,1}^{\top}
        \frac{\partial \mathbf{M}_{0,1}}{\partial \mathbf{x}_0} 
        \frac{\partial \dot{\mathbf{q}}_{0,1} }{\partial \dot{\mathbf{x}}_1^{(3)}} \\
    \end{pmatrix}\nonumber
\end{align}

\begin{align}
    \frac{\partial \mathbf{F}_{\mathbf{x}_1}}{\partial \dot{\mathbf{x}}_0} =
    \begin{pmatrix}
        \frac{1}{2}
        \frac{\partial \dot{\mathbf{q}}_{0,1}^{\top} }{\partial \dot{\mathbf{x}}_0^{(1)}}
        \frac{\partial \mathbf{M}_{0,1}}{\partial \mathbf{x}_1} 
        \dot{\mathbf{q}}_{0,1} +
        \frac{1}{2}
        \dot{\mathbf{q}}_{0,1}^{\top}
        \frac{\partial \mathbf{M}_{0,1}}{\partial \mathbf{x}_1} 
        \frac{\partial \dot{\mathbf{q}}_{0,1} }{\partial \dot{\mathbf{x}}_0^{(1)}} \\
        \frac{1}{2}
        \frac{\partial \dot{\mathbf{q}}_{0,1}^{\top} }{\partial \dot{\mathbf{x}}_0^{(2)}}
        \frac{\partial \mathbf{M}_{0,1}}{\partial \mathbf{x}_1} 
        \dot{\mathbf{q}}_{0,1} +
        \frac{1}{2}
        \dot{\mathbf{q}}_{0,1}^{\top}
        \frac{\partial \mathbf{M}_{0,1}}{\partial \mathbf{x}_1} 
        \frac{\partial \dot{\mathbf{q}}_{0,1} }{\partial \dot{\mathbf{x}}_0^{(2)}} \\
        \frac{1}{2}
        \frac{\partial \dot{\mathbf{q}}_{0,1}^{\top} }{\partial \dot{\mathbf{x}}_0^{(3)}}
        \frac{\partial \mathbf{M}_{0,1}}{\partial \mathbf{x}_1} 
        \dot{\mathbf{q}}_{0,1} +
        \frac{1}{2}
        \dot{\mathbf{q}}_{0,1}^{\top}
        \frac{\partial \mathbf{M}_{0,1}}{\partial \mathbf{x}_1} 
        \frac{\partial \dot{\mathbf{q}}_{0,1} }{\partial \dot{\mathbf{x}}_0^{(3)}} \\
    \end{pmatrix}\nonumber
\end{align}

\begin{align}
    \frac{\partial \mathbf{F}_{\mathbf{x}_1}}{\partial \dot{\mathbf{x}}_1} =
    \begin{pmatrix}
        \frac{1}{2}
        \frac{\partial \dot{\mathbf{q}}_{0,1}^{\top} }{\partial \dot{\mathbf{x}}_1^{(1)}}
        \frac{\partial \mathbf{M}_{0,1}}{\partial \mathbf{x}_1} 
        \dot{\mathbf{q}}_{0,1} +
        \frac{1}{2}
        \dot{\mathbf{q}}_{0,1}^{\top}
        \frac{\partial \mathbf{M}_{0,1}}{\partial \mathbf{x}_1} 
        \frac{\partial \dot{\mathbf{q}}_{0,1} }{\partial \dot{\mathbf{x}}_1^{(1)}} \\
        \frac{1}{2}
        \frac{\partial \dot{\mathbf{q}}_{0,1}^{\top} }{\partial \dot{\mathbf{x}}_1^{(2)}}
        \frac{\partial \mathbf{M}_{0,1}}{\partial \mathbf{x}_1} 
        \dot{\mathbf{q}}_{0,1} +
        \frac{1}{2}
        \dot{\mathbf{q}}_{0,1}^{\top}
        \frac{\partial \mathbf{M}_{0,1}}{\partial \mathbf{x}_1} 
        \frac{\partial \dot{\mathbf{q}}_{0,1} }{\partial \dot{\mathbf{x}}_1^{(2)}} \\
        \frac{1}{2}
        \frac{\partial \dot{\mathbf{q}}_{0,1}^{\top} }{\partial \dot{\mathbf{x}}_1^{(3)}}
        \frac{\partial \mathbf{M}_{0,1}}{\partial \mathbf{x}_1} 
        \dot{\mathbf{q}}_{0,1} +
        \frac{1}{2}
        \dot{\mathbf{q}}_{0,1}^{\top}
        \frac{\partial \mathbf{M}_{0,1}}{\partial \mathbf{x}_1} 
        \frac{\partial \dot{\mathbf{q}}_{0,1} }{\partial \dot{\mathbf{x}}_1^{(3)}} \\
    \end{pmatrix}\nonumber
\end{align}

\begin{equation}
    \frac{\partial \mathbf{F}_{\mathbf{x}_0}}{\partial \dot{u}_0}
    = \frac{1}{2}
    \frac{\partial \dot{\mathbf{q}}_{0,1}^{\top} }{\partial \dot{u}_0}
    \frac{\partial \mathbf{M}_{0,1}}{\partial \mathbf{x}_0} 
    \dot{\mathbf{q}}_{0,1} +
    \frac{1}{2}
    \dot{\mathbf{q}}_{0,1}^{\top}
    \frac{\partial \mathbf{M}_{0,1}}{\partial \mathbf{x}_0} 
    \frac{\partial \dot{\mathbf{q}}_{0,1} }{\partial \dot{u}_0} \nonumber
\end{equation}

\begin{equation}
    \frac{\partial \mathbf{F}_{\mathbf{x}_1}}{\partial \dot{u}_0}
    = \frac{1}{2}
    \frac{\partial \dot{\mathbf{q}}_{0,1}^{\top} }{\partial \dot{u}_0}
    \frac{\partial \mathbf{M}_{0,1}}{\partial \mathbf{x}_1} 
    \dot{\mathbf{q}}_{0,1} +
    \frac{1}{2}
    \dot{\mathbf{q}}_{0,1}^{\top}
    \frac{\partial \mathbf{M}_{0,1}}{\partial \mathbf{x}_1} 
    \frac{\partial \dot{\mathbf{q}}_{0,1} }{\partial \dot{u}_0} \nonumber
\end{equation}

\begin{equation}
    \frac{\partial \mathbf{F}_{\mathbf{x}_0}}{\partial \dot{u}_1}
    = \frac{1}{2}
    \frac{\partial \dot{\mathbf{q}}_{0,1}^{\top} }{\partial \dot{u}_1}
    \frac{\partial \mathbf{M}_{0,1}}{\partial \mathbf{x}_0} 
    \dot{\mathbf{q}}_{0,1} +
    \frac{1}{2}
    \dot{\mathbf{q}}_{0,1}^{\top}
    \frac{\partial \mathbf{M}_{0,1}}{\partial \mathbf{x}_0} 
    \frac{\partial \dot{\mathbf{q}}_{0,1} }{\partial \dot{u}_1} \nonumber
\end{equation}

\begin{equation}
    \frac{\partial \mathbf{F}_{\mathbf{x}_1}}{\partial \dot{u}_1}
    = \frac{1}{2}
    \frac{\partial \dot{\mathbf{q}}_{0,1}^{\top} }{\partial \dot{u}_1}
    \frac{\partial \mathbf{M}_{0,1}}{\partial \mathbf{x}_1} 
    \dot{\mathbf{q}}_{0,1} +
    \frac{1}{2}
    \dot{\mathbf{q}}_{0,1}^{\top}
    \frac{\partial \mathbf{M}_{0,1}}{\partial \mathbf{x}_1} 
    \frac{\partial \dot{\mathbf{q}}_{0,1} }{\partial \dot{u}_1} \nonumber
\end{equation}

\subsection{Stretching}
Stretch force resists length changes of segments (with the rest length $\|\mathbf{w}\| = 1$). Therefore, the stretching energy is generated when the length changes. We compute the energy of segment $\mathbf{[q_0,q_1]}$, in a similar way as~\citep{loock_2001_virtual,spillmann_2007_corde} :
\begin{equation}
    V_{0,1} = \frac{1}{2} Y \pi R^2 \Delta u (\| \mathbf{w}\| - 1)^2
\end{equation}
where $Y$ is yarn's elastic modulus and $R$ is yarns' radius. The stretching forces at the two nodes are:
\begin{equation}
    \mathbf{F}_{x_1} = - \mathbf{F}_{x_0} = 
    -\frac{\partial V_{0,1}}{\partial \mathbf{x}_1} = 
    -Y \pi R^2 (\|\mathbf{w}\|-1) \mathbf{d}_{0,1}
\end{equation}
\begin{equation}
    \mathbf{F}_{u_1} = - \mathbf{F}_{u_0} = 
    -\frac{\partial V_{0,1}}{\partial u_1} = 
    \frac{1}{2} Y \pi R^2 (\|\mathbf{w}\|^2 - 1)
\end{equation}
where $\mathbf{d}_{0,1}$ is the unit vector points from $\mathbf{q}_0$ to $\mathbf{q}_1$, $\mathbf{d}_{0,1} = \frac{\mathbf{x}_1 - \mathbf{x}_0}{\|\mathbf{x}_1 - \mathbf{x}_0\|}$. The derivatives of the stretching forces with respect to nodes' positions are:
\begin{equation}
    \frac{\partial \mathbf{F}_{\mathbf{x}_1}}{\partial \mathbf{x}_1} =
    \frac{\partial \mathbf{F}_{\mathbf{x}_0}}{\partial \mathbf{x}_0} =
    - \frac{\partial \mathbf{F}_{\mathbf{x}_1}}{\partial \mathbf{x}_0} = 
    - \frac{\partial \mathbf{F}_{\mathbf{x}_0}}{\partial \mathbf{x}_1} = 
    Y \pi R^2(\frac{1}{l_1} \mathbf{P}_{0,1} - \frac{1}{\Delta u}\mathbf{I})
\end{equation}
\begin{equation}
    \frac{\partial F_{u_1}}{\partial u_1} =
    \frac{\partial F_{u_0}}{\partial u_0} =
    - \frac{\partial F_{u_1}}{\partial u_0} = 
    - \frac{\partial F_{u_0}}{\partial u_1} = 
    - Y \pi R^2 \frac{\|\mathbf{w}\|^2}{\Delta u}
\end{equation}
\begin{equation}
    \frac{\partial \mathbf{F}_{\mathbf{x}_1}}{\partial u_1} =
    \frac{\partial \mathbf{F}_{\mathbf{x}_0}}{\partial u_0} =
    - \frac{\partial \mathbf{F}_{\mathbf{x}_1}}{\partial u_0} = 
    - \frac{\partial \mathbf{F}_{\mathbf{x}_0}}{\partial u_1} = 
    Y \pi R^2 \frac{\|\mathbf{w}\|^2}{\Delta u} \mathbf{d}_{0,1}
\end{equation}
\begin{equation}
    \frac{\partial F_{u_1}}{\partial \mathbf{x}_1} =
    \frac{\partial F_{u_0}}{\partial \mathbf{x}_0} =
    - \frac{\partial F_{u_1}}{\partial \mathbf{x}_0} = 
    - \frac{\partial F_{u_0}}{\partial \mathbf{x}_1} = 
    \frac{Y \pi R^2}{\Delta u} \mathbf{w}^{\top}
\end{equation}
where $\mathbf{P}_{0,1} = \mathbf{I}_3 - \mathbf{d}_{0,1}\mathbf{d}_{0,1}^{\top}$
\subsection{Bending} 
We adopt the discrete differential geometry method \citep{sullivan_2008_curves} to define the curvature at the common crossing node of two adjacent segments. Bending energy is defined as the integration of bending energy density along the two segments. The bending energy on the two connected warp segments $[\mathbf{q}_2,\mathbf{q}_0 ]$ and $[\mathbf{q}_0,\mathbf{q}_1 ]$ is 
 \begin{equation}
    V_{2,0,1} = B \pi R^2 \frac{\theta^2}{u_1 - u_2}
 \end{equation}
where $B$ is yarn bending modulus and $\theta = \arcsin(-\mathbf{d}_{0,1}^{\top}\mathbf{d}_{0,2})$ is the angle between the two segments. Its derivatives with respective to the node position are the bending forces:
\begin{equation}
    \mathbf{F}_{\mathbf{x}_1} = - \frac{2B \pi R^2 \theta}{l_1 (u_1 - u_2)\sin{\theta}} \mathbf{P}_{0,1} \mathbf{d_{0,2}}
\end{equation}
\begin{equation}
    \mathbf{F}_{\mathbf{x}_2} = - \frac{(2B \pi R^2 \theta)}{l_2 (u_1 - u_2) \sin{\theta}} \mathbf{P}_{0,2} \mathbf{d_{0,1}}
\end{equation}
\begin{equation}
    \mathbf{F}_{\mathbf{x}_0} = - (\mathbf{F}_{\mathbf{x}_1} + \mathbf{F}_{\mathbf{x}_2})
\end{equation}
\begin{equation}
    \mathbf{F}_{u_1} = -\mathbf{F}_{u_2} = \frac{2B \pi R^2 \theta^2}{(u_1 - u_2)^2} 
\end{equation}
\begin{equation}
    \mathbf{F}_{u_0} = 0
\end{equation}
The derivatives of the bending forces with respected to the nodes' position are
\begin{align}
    \frac{\partial \mathbf{F}_{\mathbf{x}_1}}{\partial \mathbf{x}_1} =&
    \frac{2 B \pi R^2}{l_1^2(u_1 - u_0) \sin{\theta}} 
    \bigg( \theta
    \bigg(
    \mathbf{P}_{0,1} \mathbf{d}_{0,2} \mathbf{d}_{0,1}^{\top} +
    \frac{\cos{\theta}}{\sin^2{\theta}}\mathbf{P}_{0,1} \mathbf{d}_{0,2} \mathbf{d}_{0,2}^{\top} \mathbf{P}_{0,1} +
    \cos{\theta} \mathbf{P}_{0,1} \notag \\
    &+ \mathbf{d}_{0,1} \mathbf{d}_{0,2}^{\top} \mathbf{P}_{0,1}
    \bigg)
    -\frac{1}{\sin{\theta}} \mathbf{P}_{0,1} \mathbf{d}_{0,2} \mathbf{d}_{0,2}^{\top} \mathbf{P}_{0,1}
    \bigg)
\end{align}

\begin{align}
    \frac{\partial \mathbf{F}_{\mathbf{x}_2}}{\partial \mathbf{x}_2} =&
    \frac{2 B \pi R^2}{l_2^2(u_1 - u_0) \sin{\theta}} 
    \bigg( \theta
    \bigg(
    \mathbf{P}_{0,2} \mathbf{d}_{0,1} \mathbf{d}_{0,2}^{\top} +
    \frac{\cos{\theta}}{\sin^2{\theta}}\mathbf{P}_{0,2} \mathbf{d}_{0,1} \mathbf{d}_{0,1}^{\top} \mathbf{P}_{0,2} +
    \cos{\theta} \mathbf{P}_{0,2} \notag \\
    &+ \mathbf{d}_{0,2} \mathbf{d}_{0,1}^{\top} \mathbf{P}_{0,2}
    \bigg)
    -\frac{1}{\sin{\theta}} \mathbf{P}_{0,2} \mathbf{d}_{0,1} \mathbf{d}_{0,1}^{\top} \mathbf{P}_{0,2}
    \bigg)
\end{align}

\begin{equation}
    \frac{\partial \mathbf{F}_{\mathbf{x}_1}}{\partial \mathbf{x}_2} =
    - \frac{2 B \pi R^2}{l_2 l_1 (u_1 - u_2) \sin{\theta}}
    \bigg( \theta
    \bigg(
    \mathbf{P}_{0,1} - 
    \frac{\cos{\theta}}{\sin^2{\theta}} \mathbf{P}_{0,1} \mathbf{d}_{0,2} \mathbf{d}_{0,1}^{\top}
    \bigg) +
    \frac{1}{\sin{\theta}} \mathbf{P}_{0,1} \mathbf{d}_{0,2} \mathbf{d}_{0,1}^{\top}
    \bigg) \mathbf{P}_{0,2}
\end{equation}

\begin{equation}
    \frac{\partial \mathbf{F}_{\mathbf{x}_2}}{\partial \mathbf{x}_1} =
    - \frac{2 B \pi R^2}{l_1 l_2 (u_1 - u_2) \sin{\theta}}
    \bigg( \theta
    \bigg(
    \mathbf{P}_{0,2} - 
    \frac{\cos{\theta}}{\sin^2{\theta}} \mathbf{P}_{0,2} \mathbf{d}_{0,1} \mathbf{d}_{0,2}^{\top}
    \bigg) +
    \frac{1}{\sin{\theta}} \mathbf{P}_{0,2} \mathbf{d}_{0,1} \mathbf{d}_{0,2}^{\top}
    \bigg) \mathbf{P}_{0,1}
\end{equation}

\begin{equation}
    \frac{\partial \mathbf{F}_{\mathbf{x}_1}}{\partial \mathbf{x}_0} = 
    - \bigg(
    \frac{\partial \mathbf{F}_{\mathbf{x}_1}}{\partial \mathbf{x}_1} + 
    \frac{\partial \mathbf{F}_{\mathbf{x}_1}}{\partial \mathbf{x}_2}
    \bigg)
\end{equation}

\begin{equation}
    \frac{\partial \mathbf{F}_{\mathbf{x}_2}}{\partial \mathbf{x}_0} = 
    - \bigg(
    \frac{\partial \mathbf{F}_{\mathbf{x}_2}}{\partial \mathbf{x}_1} + 
    \frac{\partial \mathbf{F}_{\mathbf{x}_2}}{\partial \mathbf{x}_2}
    \bigg)
\end{equation}

\begin{equation}
    \frac{\partial \mathbf{F}_{\mathbf{x}_0}}{\partial \mathbf{x}_1} = 
    - \bigg(
    \frac{\partial \mathbf{F}_{\mathbf{x}_1}}{\partial \mathbf{x}_1} + 
    \frac{\partial \mathbf{F}_{\mathbf{x}_2}}{\partial \mathbf{x}_1}
    \bigg)
\end{equation}

\begin{equation}
    \frac{\partial \mathbf{F}_{\mathbf{x}_0}}{\partial \mathbf{x}_2} = 
    - \bigg(
    \frac{\partial \mathbf{F}_{\mathbf{x}_1}}{\partial \mathbf{x}_2} + 
    \frac{\partial \mathbf{F}_{\mathbf{x}_2}}{\partial \mathbf{x}_2}
    \bigg)
\end{equation}

\begin{equation}
    \frac{\partial \mathbf{F}_{\mathbf{x}_0}}{\partial \mathbf{x}_0} = 
    - \bigg(
    \frac{\partial \mathbf{F}_{\mathbf{x}_1}}{\partial \mathbf{x}_0} + 
    \frac{\partial \mathbf{F}_{\mathbf{x}_2}}{\partial \mathbf{x}_0}
    \bigg)
\end{equation}

\begin{equation}
    \frac{\partial F_{u_1}}{\partial u_1} = 
    \frac{\partial F_{u_2}}{\partial u_2} =
    \frac{\partial F_{u_1}}{\partial u_2} =
    \frac{\partial F_{u_2}}{\partial u_1} = 
    -\frac{2 B \pi R^2 \theta^2}{(u_1 - u_2)^2}
\end{equation}

\begin{equation}
    \frac{\partial \mathbf{F}_{\mathbf{x}_1}}{\partial u_1} = 
    - \frac{\partial \mathbf{F}_{\mathbf{x}_1}}{\partial u_2} = 
    \frac{2 B \pi R^2 \theta}{l_1 (u_1 - u_2)^2 \sin{\theta}} \mathbf{P}_{0,1} \mathbf{d}_{0,2}
\end{equation}

\begin{equation}
    \frac{\partial \mathbf{F}_{\mathbf{x}_2}}{\partial u_1} = 
    - \frac{\partial \mathbf{F}_{\mathbf{x}_2}}{\partial u_2} = 
    \frac{2 B \pi R^2 \theta}{l_2 (u_1 - u_2)^2 \sin{\theta}} \mathbf{P}_{0,2} \mathbf{d}_{0,1}
\end{equation}

\begin{equation}
    \frac{\partial \mathbf{F}_{\mathbf{x}_0}}{\partial u_1} = 
    - \frac{\partial \mathbf{F}_{\mathbf{x}_0}}{\partial u_2} = 
    - \bigg(
    \frac{\partial \mathbf{F}_{\mathbf{x}_1}}{\partial u_1} + 
    \frac{\partial \mathbf{F}_{\mathbf{x}_2}}{\partial u_1}
    \bigg)
\end{equation}

\begin{equation}
    \frac{\partial \mathbf{F}_{u_1}}{\partial \mathbf{x}_1} = 
    - \frac{\partial \mathbf{F}_{u_2}}{\partial \mathbf{x}_1} = 
    \frac{2 B \pi R^2 \theta}{l_1 (u_1 - u_2)^2 \sin{\theta}} \mathbf{d}_{0,2}^{\top} \mathbf{P}_{0,1} 
\end{equation}

\begin{equation}
    \frac{\partial \mathbf{F}_{u_1}}{\partial \mathbf{x}_2} = 
    - \frac{\partial \mathbf{F}_{u_2}}{\partial \mathbf{x}_2} = 
    \frac{2 B \pi R^2 \theta}{l_2 (u_1 - u_2)^2 \sin{\theta}} \mathbf{d}_{0,1}^{\top} \mathbf{P}_{0,2} 
\end{equation}

\begin{equation}
    \frac{\partial \mathbf{F}_{u_1}}{\partial \mathbf{x}_0} = 
    - \frac{\partial \mathbf{F}_{u_2}}{\partial \mathbf{x}_0} = 
    - \bigg(
    \frac{\partial \mathbf{F}_{u_1}}{\partial \mathbf{x}_1} + 
    \frac{\partial \mathbf{F}_{u_1}}{\partial \mathbf{x}_2}
    \bigg)
\end{equation}

\subsection{Slide Friction}
The slide friction at a crossing node $\mathbf{q}_0$ along warp $u$ direction is
\begin{equation}
\label{eq:app_friction}
    F_{Slide} = -\Big( 
    \frac{k_f\delta u - K(\delta u)\mu F_n}{2} K(\mu F_n - F_u)
    + \frac{k_f\delta u + K(\delta u)\mu F_n}{2} \Big) - d_f \dot{u}_0
\end{equation}
The derivative of friction force with respect to node position in Eulerian coordinate is
\begin{align}
    \frac{\partial F_{Slide}}{\partial u_0} =& 
    -\frac{k_f - ((1 - \tanh^2{\delta u})\mu F_n + \tanh{\delta u} \mu \frac{\partial F_n}{\partial u_0})}{2} \tanh{(\mu F_n - F_u)} \notag \\
    &- \frac{k_f\delta u - \tanh{\delta u} \mu F_n}{2}(1 - \tanh^2{(\mu F_n - F_u)})
    \left(\frac{\partial F_u}{\partial u_0} - \mu \frac{\partial F_n}{\partial u_0}\right) \notag \\
    &- \frac{k_f + (1 - \tanh^2{\delta u})\mu F_n + \tanh{\delta u} \mu \frac{\partial F_n}{\partial u_0}}{2}
\end{align}
The derivative of friction force with respect to node velocity in Eulerian coordinate is 
\begin{equation}
    \frac{\partial F_{Slide}}{\partial \dot{u}_0} = 
    \frac{k_f \delta u - \tanh{\delta u} \mu F_n}{2} 
    (1 - \tanh^2(\mu F_n - F_u)) 
    \frac{\partial F_u}{\partial \dot{u}_0} - d_f
\end{equation}

\subsection{Shearing}
The potential energy over the segments $[\mathbf{q}_0, \mathbf{q}_1]$ and $[\mathbf{q}_0, \mathbf{q}_3]$ caused by shearing deformation is 
\begin{equation}
    V_{1,0,3} = \frac{1}{2} k_s  L (\phi - \bar{\phi})^2
\end{equation}
\begin{equation}
    k_s = \frac{1}{2} (F_n + 1) S R^2 \Bigg( (1 + \gamma^c) + (1 - \gamma^c)
    \tanh\left( 
    \frac{\bar{\phi}^5(\phi-\phi_l)}
    {(\phi(\phi-\phi_l)(\phi-\bar{\phi}))^2 + \bar{\phi}^4 \sigma^2}
    \right) \Bigg) \nonumber
\end{equation}
The shear forces at those crossing nodes are
\begin{equation}
    \mathbf{F}_{\mathbf{x}_1} = 
    -\frac{\partial V_{1,0,3}}{\partial \mathbf{x}_1} = 
    -\frac{1}{2} \frac{\partial k_s}{\partial \mathbf{x}_1} L (\phi - \bar{\phi})^2
    + \frac{k_s L (\phi - \bar{\phi})}{l_1 \sin{\phi}} \mathbf{P}_{0,1} \mathbf{d}_{0,3}
\end{equation}
\begin{equation}
    \mathbf{F}_{\mathbf{x}_3} = 
    -\frac{\partial V_{1,0,3}}{\partial \mathbf{x}_3} = 
    -\frac{1}{2} \frac{\partial k_s}{\partial \mathbf{x}_3} L (\phi - \bar{\phi})^2
    + \frac{k_s L (\phi - \bar{\phi})}{l_3 \sin{\phi}} \mathbf{P}_{0,3} \mathbf{d}_{0,1}
\end{equation}
\begin{equation}
    \mathbf{F}_{\mathbf{x}_0} = - (\mathbf{F}_{\mathbf{x}_1} + \mathbf{F}_{\mathbf{x}_3})
\end{equation}
For the sake of simplicity, we define:
\begin{equation}
    g(\phi) =
    \frac{\bar{\phi}^5(\phi-\phi_l)}
    {(\phi(\phi-\phi_l)(\phi-\bar{\phi}))^2 + \bar{\phi}^4 \sigma^2}\nonumber
\end{equation}
\begin{equation}
    f(\phi) = \tanh{g(\phi)}\nonumber
\end{equation}

The numerator and denominator of $g(\phi)$ are
\begin{equation}
    g_{num}(\phi) = \bar{\phi}^5(\phi-\phi_l) \nonumber
\end{equation}
and 
\begin{equation}
    g_{den}(\phi) = (\phi(\phi-\phi_l)(\phi-\bar{\phi}))^2 + \bar{\phi}^4 \sigma^2 \nonumber
\end{equation}
Then, we have:
\begin{equation}
    \frac{\partial k_s}{\partial \mathbf{x}_3} =
    \frac{1}{2} (F_n + 1) S R^2 
    \left(c \gamma^{c-1} \frac{\partial \gamma}{\partial \mathbf{x}_3} -
    c \gamma^{c-1} \frac{\partial \gamma}{\partial \mathbf{x}_3} f(\phi) +
    (1-\gamma^{c}) (1 - f(\phi)^2) \frac{\partial g(\phi)}{\partial \mathbf{x}_3}\right)\nonumber
\end{equation}
\begin{equation}
    \frac{\partial k_s}{\partial \mathbf{x}_1} = 
    \frac{1}{2} (F_n + 1) S R^2 
    \left(c \gamma^{c-1} \frac{\partial \gamma}{\partial \mathbf{x}_1} -
    c \gamma^{c-1} \frac{\partial \gamma}{\partial \mathbf{x}_1} f(\phi) +
    (1-\gamma^{c}) (1 - f(\phi)^2) \frac{\partial g(\phi)}{\partial \mathbf{x}_1}\right)\nonumber
\end{equation}
where
\begin{equation}
    \frac{\partial \gamma}{\partial \mathbf{x}_1} = 
    - \frac{L}{R} \cos{\frac{\phi}{2}} \frac{\partial \phi}{\partial \mathbf{x}_1}
    \mbox{, }
    \frac{\partial \gamma}{\partial \mathbf{x}_3} = 
    - \frac{L}{R} \cos{\frac{\phi}{2}} \frac{\partial \phi}{\partial \mathbf{x}_3} 
    \mbox{, }\nonumber
\end{equation}

\begin{equation}
    \frac{\partial g(\phi)}{\partial \mathbf{x}_1} = 
    \frac{
        \frac{\partial g_{num}(\phi)}{\partial \mathbf{x}_1} g_{den}(\phi) -
        g_{num}(\phi) \frac{\partial g_{den}(\phi)}{\partial \mathbf{x}_1}
    }
    {g_{den}^2(\phi)}
    \mbox{,}\nonumber
\end{equation}

\begin{equation}
    \frac{\partial g(\phi)}{\partial \mathbf{x}_3} = 
    \frac{
        \frac{\partial g_{num}(\phi)}{\partial \mathbf{x}_3} g_{den}(\phi) -
        g_{num}(\phi) \frac{\partial g_{den}(\phi)}{\partial \mathbf{x}_3}
    }
    {g_{den}^2(\phi)}
    \mbox{.}\nonumber
\end{equation}
The terms 
$\frac{\partial g_{num}(\phi)}{\partial \mathbf{x}_1}$, 
$\frac{\partial g_{den}(\phi)}{\partial \mathbf{x}_1}$,
$\frac{\partial g_{num}(\phi)}{\partial \mathbf{x}_3}$, and
$\frac{\partial g_{den}(\phi)}{\partial \mathbf{x}_3}$ are:
\begin{equation}
    \frac{\partial g_{num}(\phi)}{\partial \mathbf{x}_1} = 
    \bar{\phi}^5 \frac{\partial \phi}{\partial \mathbf{x}_1} =
    - \bar{\phi}^5 \frac{\mathbf{P}_{0,1}\mathbf{d}_{0,3}}{l_1 \sin{\phi}}\nonumber
\end{equation}
\begin{equation}
    \frac{\partial g_{num}(\phi)}{\partial \mathbf{x}_3} = 
    \bar{\phi}^5 \frac{\partial \phi}{\partial \mathbf{x}_3} =
    - \bar{\phi}^5 \frac{\mathbf{P}_{0,3}\mathbf{d}_{0,1}}{l_3 \sin{\phi}}\nonumber
\end{equation}
\begin{equation}
    \frac{\partial g_{den}(\phi)}{\partial \mathbf{x}_1} =
    2(\phi(\phi - \phi_l)(\phi - \bar{\phi}))
    \left(
    \frac{\partial \phi}{\partial \mathbf{x}_1}(\phi-\phi_l) (\phi-\bar{\phi}) + 
    \phi \frac{\partial \phi}{\partial \mathbf{x}_1} (\phi-\bar{\phi}) + 
    \phi(\phi-\phi_l)\frac{\partial \phi}{\partial \mathbf{x}_1}
    \right)\nonumber
\end{equation}
\begin{equation}
    \frac{\partial g_{den}(\phi)}{\partial \mathbf{x}_3} =
    2(\phi(\phi - \phi_l)(\phi - \bar{\phi}))
    \left(
    \frac{\partial (\phi)}{\partial \mathbf{x}_3}(\phi-\phi_l) (\phi-\bar{\phi}) + 
    \phi \frac{\partial \phi}{\partial \mathbf{x}_3} (\phi-\bar{\phi}) + 
    \phi(\phi-\phi_l)\frac{\partial \phi}{\partial \mathbf{x}_3}
    \right)\nonumber
\end{equation}
The derivatives of the shear forces with respect to the nodes' positions in Lagrangian coordinate are:
\begin{equation}
    \frac{\partial \mathbf{F}_{\mathbf{x}_1}}{\partial \mathbf{x}_1} = 
    - \frac{1}{2} \frac{\partial^2 k_s}{\partial \mathbf{x}_1 \mathbf{x}_1} L (\phi - \bar{\phi})^2
    - L (\phi - \bar{\phi}) \frac{\partial k_s}{\partial \mathbf{x}_1} \frac{\partial \phi}{\partial \mathbf{x}_1} 
    + \frac{\partial }{\partial \mathbf{x}_1} \frac{k_s L (\phi - \bar{\phi})}{l_1 \sin{\phi}} \mathbf{P}_{0,1} \mathbf{d}_{0,3}\nonumber
\end{equation}

\begin{equation}
    \frac{\partial \mathbf{F}_{\mathbf{x}_3}}{\partial \mathbf{x}_3} = 
    - \frac{1}{2} \frac{\partial^2 k_s}{\partial \mathbf{x}_3 \mathbf{x}_3} L (\phi - \bar{\phi})^2
    - L (\phi - \bar{\phi}) \frac{\partial k_s}{\partial \mathbf{x}_3} \frac{\partial \phi}{\partial \mathbf{x}_3} 
    + \frac{\partial }{\partial \mathbf{x}_3} \frac{k_s L (\phi - \bar{\phi})}{l_3 \sin{\phi}} \mathbf{P}_{0,3} \mathbf{d}_{0,1}\nonumber
\end{equation}

\begin{equation}
    \frac{\partial \mathbf{F}_{\mathbf{x}_1}}{\partial \mathbf{x}_3} = 
    - \frac{1}{2} \frac{\partial^2 k_s}{\partial \mathbf{x}_1 \mathbf{x}_3} L (\phi - \bar{\phi})^2
    - L (\phi - \bar{\phi}) \frac{\partial k_s}{\partial \mathbf{x}_1} \frac{\partial \phi}{\partial \mathbf{x}_3} 
    - L (\phi - \bar{\phi}) \frac{\partial \phi}{\partial \mathbf{x}_1} \frac{\partial k_s}{\partial \mathbf{x}_3} 
    + \frac{\partial }{\partial \mathbf{x}_3} \frac{k_s L (\phi - \bar{\phi})}{l_1 \sin{\phi}} \mathbf{P}_{0,1} \mathbf{d}_{0,3}\nonumber
\end{equation}

\begin{equation}
    \frac{\partial \mathbf{F}_{\mathbf{x}_3}}{\partial \mathbf{x}_1} = 
    - \frac{1}{2} \frac{\partial^2 k_s}{\partial \mathbf{x}_3 \mathbf{x}_1} L (\phi - \bar{\phi})^2
    - L (\phi - \bar{\phi}) \frac{\partial k_s}{\partial \mathbf{x}_3} \frac{\partial \phi}{\partial \mathbf{x}_1} 
    - L (\phi - \bar{\phi}) \frac{\partial \phi}{\partial \mathbf{x}_3} \frac{\partial k_s}{\partial \mathbf{x}_1} 
    + \frac{\partial }{\partial \mathbf{x}_1} \frac{k_s L (\phi - \bar{\phi})}{l_3 \sin{\phi}} \mathbf{P}_{0,3} \mathbf{d}_{0,1}\nonumber
\end{equation}
where
\begin{align}
    &\frac{\partial }{\partial \mathbf{x}_1} \frac{k_s L (\phi - \bar{\phi})}{l_1 \sin{\phi}} \mathbf{P}_{0,1} \mathbf{d}_{0,3} =
    \frac{k_s L}{l_1^2 \sin{\phi}} 
    \bigg(
    \Big( \phi - \bar{\phi} \Big)
    \bigg(
    - \mathbf{P}_{0,1} \mathbf{d}_{0,3} \mathbf{d}_{0,1}^{\top} 
    + \frac{\cos{\phi}}{\sin^2{\phi}} \mathbf{P}_{0,1} \mathbf{d}_{0,3} \mathbf{d}_{0,3}^{\top} \mathbf{P}_{0,1} \notag \\
    &- \cos{\phi} \mathbf{P}_{0,1}
    - \mathbf{d}_{0,1} \mathbf{d}_{0,3}^{\top} \mathbf{P}_{0,1}
    \bigg) 
    - \frac{1}{\sin{\phi}} \mathbf{P}_{0,1} \mathbf{d}_{0,3} \mathbf{d}_{0,3}^{\top} \mathbf{P}_{0,1}
    \bigg) \nonumber
\end{align}

\begin{align}
    & \frac{\partial }{\partial \mathbf{x}_3} \frac{k_s L (\phi - \bar{\phi})}{l_3 \sin{\phi}} \mathbf{P}_{0,3} \mathbf{d}_{0,1} =
    \frac{k_s L}{l_3^2 \sin{\phi}} 
    \bigg(
    \Big( \phi - \bar{\phi} \Big)
    \bigg(
    - \mathbf{P}_{0,3} \mathbf{d}_{0,1} \mathbf{d}_{0,3}^{\top} 
    + \frac{\cos{\phi}}{\sin^2{\phi}} \mathbf{P}_{0,3} \mathbf{d}_{0,1} \mathbf{d}_{0,1}^{\top} \mathbf{P}_{0,3} \notag \\
    &- \cos{\phi} \mathbf{P}_{0,3}
    - \mathbf{d}_{0,3} \mathbf{d}_{0,1}^{\top} \mathbf{P}_{0,3}
    \bigg) 
    - \frac{1}{\sin{\phi}} \mathbf{P}_{0,3} \mathbf{d}_{0,1} \mathbf{d}_{0,1}^{\top} \mathbf{P}_{0,3}
    \bigg) \nonumber
\end{align}

\begin{align}
    &\frac{\partial }{\partial \mathbf{x}_3} \frac{k_s L (\phi - \bar{\phi})}{l_1 \sin{\phi}} \mathbf{P}_{0,1} \mathbf{d}_{0,3} =
    \frac{k_s L}{l_3 l_1 \sin{\phi}} 
    \bigg(
    \Big( \phi - \bar{\phi} \Big)
    \bigg(
    \frac{\cos{\phi}}{\sin^2{\phi}} \mathbf{P}_{0,1} \mathbf{d}_{0,3} \mathbf{d}_{0,1}^{\top} \mathbf{P}_{0,3}
    + \mathbf{P}_{0,1} \mathbf{P}_{0,3}
    \bigg) 
    \notag \\
    &- \frac{1}{\sin{\phi}} \mathbf{P}_{0,1} \mathbf{d}_{0,3} \mathbf{d}_{0,1}^{\top} \mathbf{P}_{0,3}
    \bigg) \nonumber
\end{align}

\begin{align}
    &\frac{\partial }{\partial \mathbf{x}_1} \frac{k_s L (\phi - \bar{\phi})}{l_3 \sin{\phi}} \mathbf{P}_{0,3} \mathbf{d}_{0,1} =
    \frac{k_s L}{l_1 l_3 \sin{\phi}} 
    \bigg(
    \Big( \phi - \bar{\phi} \Big)
    \bigg(
    \frac{\cos{\phi}}{\sin^2{\phi}} \mathbf{P}_{0,3} \mathbf{d}_{0,1} \mathbf{d}_{0,3}^{\top} \mathbf{P}_{0,1}
    + \mathbf{P}_{0,3} \mathbf{P}_{0,1}
    \bigg) 
    \notag \\
    &- \frac{1}{\sin{\phi}} \mathbf{P}_{0,3} \mathbf{d}_{0,1} \mathbf{d}_{0,3}^{\top} \mathbf{P}_{0,1}
    \bigg) \nonumber
\end{align}
Moreover, the other terms are
\begin{equation}
    \frac{\partial \mathbf{F}_{\mathbf{x}_1}}{\partial \mathbf{x}_0} = 
    - \left(
    \frac{\partial \mathbf{F}_{\mathbf{x}_1}}{\partial \mathbf{x}_1} +
    \frac{\partial \mathbf{F}_{\mathbf{x}_1}}{\partial \mathbf{x}_3}
    \right)
    \mbox{, }
    \frac{\partial \mathbf{F}_{\mathbf{x}_3}}{\partial \mathbf{x}_0} = 
    - \left(
    \frac{\partial \mathbf{F}_{\mathbf{x}_3}}{\partial \mathbf{x}_1} +
    \frac{\partial \mathbf{F}_{\mathbf{x}_3}}{\partial \mathbf{x}_3}
    \right)\nonumber
\end{equation}

\begin{equation}
    \frac{\partial \mathbf{F}_{\mathbf{x}_0}}{\partial \mathbf{x}_1} = 
    - \left(
    \frac{\partial \mathbf{F}_{\mathbf{x}_1}}{\partial \mathbf{x}_1} +
    \frac{\partial \mathbf{F}_{\mathbf{x}_3}}{\partial \mathbf{x}_1}
    \right)
    \mbox{, }
    \frac{\partial \mathbf{F}_{\mathbf{x}_0}}{\partial \mathbf{x}_3} = 
    - \left(
    \frac{\partial \mathbf{F}_{\mathbf{x}_1}}{\partial \mathbf{x}_3} +
    \frac{\partial \mathbf{F}_{\mathbf{x}_3}}{\partial \mathbf{x}_3}
    \right)\nonumber
\end{equation}

\begin{equation}
    \frac{\partial \mathbf{F}_{\mathbf{x}_0}}{\partial \mathbf{x}_0} = 
    - \left(
    \frac{\partial \mathbf{F}_{\mathbf{x}_1}}{\partial \mathbf{x}_0} +
    \frac{\partial \mathbf{F}_{\mathbf{x}_3}}{\partial \mathbf{x}_0}
    \right)\nonumber
\end{equation}

\subsection{Yarn-to-yarn collision}
\begin{equation}
     V_{0,1} = \frac{1}{2} k_c L \mbox{ReLU}(d - \Delta u)^2
\end{equation}
The yarn-to-yarn collision forces are:
\begin{equation}
    F_{u_0} = 
    - \frac{\partial V_{0,1}}{\partial u_0} = 
    k_c L (\Delta u - d)
\end{equation}

\begin{equation}
    F_{u_1} = 
    - \frac{\partial V_{0,1}}{\partial u_1} = 
    - k_c L (\Delta u - d)
\end{equation}
The derivatives of the forces with respect to the nodes' position in Eulerian coordiates:
\begin{equation}
    \frac{\partial F_{u_0}}{\partial u_0} = 
    \frac{\partial F_{u_1}}{\partial u_1} = -
    \frac{\partial F_{u_0}}{\partial u_1} = -
    \frac{\partial F_{u_1}}{\partial u_0} = 
    - k_c L 
\end{equation}

\subsection{Gravity}
 We define a gravitational energy which is computed segment-wise. To a warp segment $[\mathbf{q}_0,\mathbf{q}_1 ]$, it gravitational energy is defined as
\begin{equation}
    V_{0,1} = \rho \Delta u \mathbf{g}^{\top} \frac{\mathbf{x}_0 + \mathbf{x}_1}{2}
\end{equation}
where $\mathbf{g} \in \mathbb{R}_3$ is the gravity of earth which is approximately set to $(0, 0, 9.8)$. The gravity at the nodes are
\begin{equation}
    \mathbf{F}_{\mathbf{x}_0} = 
    - \frac{\partial V_{0,1}}{\partial \mathbf{x}_0} =
    - \frac{1}{2} \rho \mathbf{g} \Delta u
\end{equation}
\begin{equation}
    \mathbf{F}_{\mathbf{x}_1} = 
    - \frac{\partial V_{0,1}}{\partial \mathbf{x}_1} =
    - \frac{1}{2} \rho \mathbf{g} \Delta u
\end{equation}
\begin{equation}
    \mathbf{F}_{u_0} = 
    - \frac{\partial V_{0,1}}{\partial u_0} =
    \frac{1}{2} \rho \mathbf{g}^{\top} (\mathbf{x}_1 + \mathbf{x}_0)
\end{equation}
\begin{equation}
    \mathbf{F}_{u_1} = 
    - \frac{\partial V_{0,1}}{\partial u_1} =
    - \frac{1}{2} \rho \mathbf{g}^{\top} (\mathbf{x}_1 + \mathbf{x}_0)
\end{equation}
The derivative of the force with respect to the nodes' position are:
\begin{equation}
    \frac{\partial \mathbf{F}_{\mathbf{x}_0}}{\partial u_0} = 
    \frac{1}{2} \rho \mathbf{g}
    \quad
    \frac{\partial \mathbf{F}_{\mathbf{x}_0}}{\partial u_1} = 
    - \frac{1}{2} \rho \mathbf{g}
\end{equation}

\begin{equation}
    \frac{\partial \mathbf{F}_{\mathbf{x}_1}}{\partial u_0} = 
    \frac{1}{2} \rho \mathbf{g}
    \quad
    \frac{\partial \mathbf{F}_{\mathbf{x}_1}}{\partial u_1} = 
    - \frac{1}{2} \rho \mathbf{g}
\end{equation}

\begin{equation}
    \frac{\partial \mathbf{F}_{u_0}}{\partial \mathbf{x}_1} = 
    \frac{1}{2} \rho \mathbf{g}^{\top}
    \quad
    \frac{\partial \mathbf{F}_{u_0}}{\partial \mathbf{x}_0} = 
    \frac{1}{2} \rho \mathbf{g}^{\top}
\end{equation}

\begin{equation}
    \frac{\partial \mathbf{F}_{u_1}}{\partial \mathbf{x}_1} = 
    - \frac{1}{2} \rho \mathbf{g}^{\top}
    \quad
    \frac{\partial \mathbf{F}_{u_1}}{\partial \mathbf{x}_0} = 
    - \frac{1}{2} \rho \mathbf{g}^{\top}
\end{equation}

\subsection{Wind Force}
\begin{figure}
    \centering
    \includegraphics[width=0.4\textwidth]{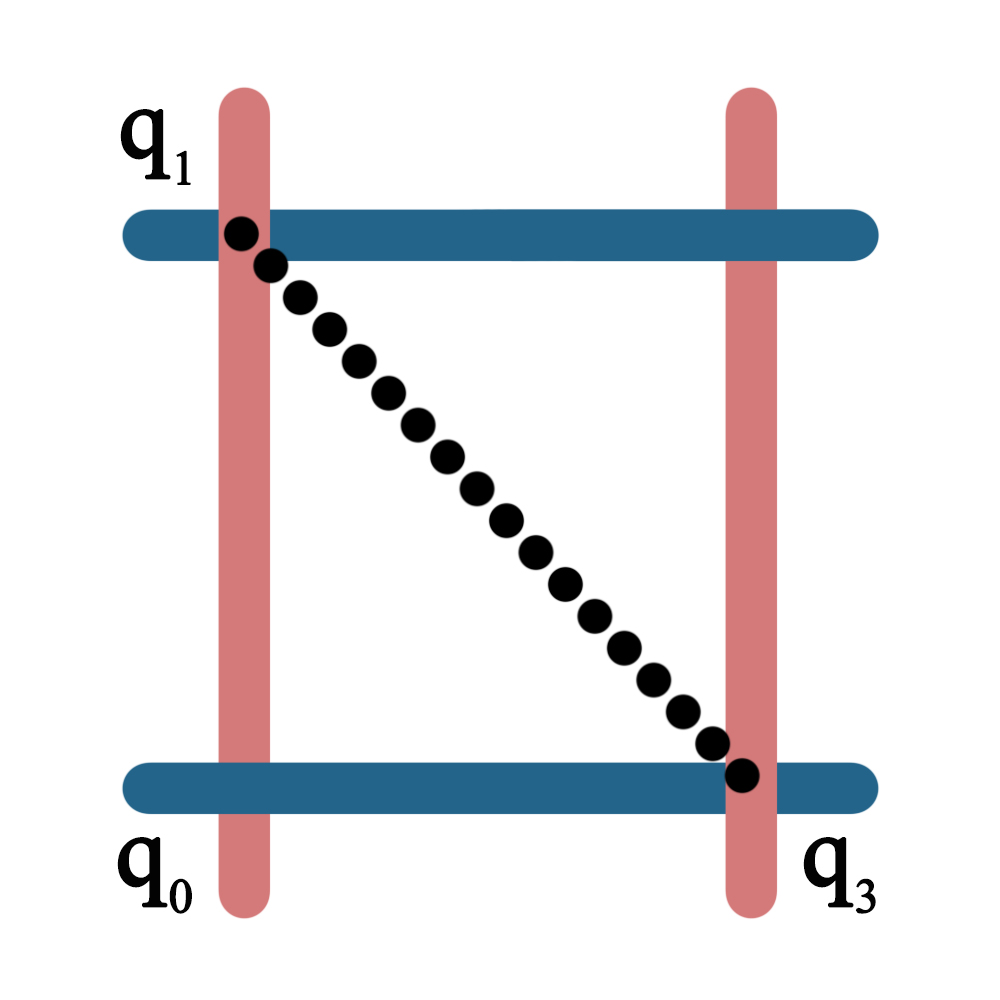}
    \caption{Treat a square hold in 4 segments as two triangles.}
    \label{fig:wind_triangle}
\end{figure}
To apply wind force to the surface of the cloth, we need to compute an area-based force. Every square composed of four segments can be split into two triangles when computing wind force( shown in \ref{fig:wind_triangle}). The wind force has three properties affecting its influence on the cloth: velocity $\mathbf{v}_w$, density $\rho_w$, and drag $d_w$. 
$\mathbf{v}_w = (0, 5, 0)$, density $\rho_w = 2$, and drag $d_w = 0.5$.
The wind force imposed on a triangle face $[\mathbf{q}_0, \mathbf{q}_1, \mathbf{q}_3]$ is:
\begin{equation}
    \mathbf{F}_w = \rho_w a |v_n| v_n \mathbf{n}_f + d_w \mathbf{v}_t
\end{equation}
where $a$ is face area, $\mathbf{n}_f$ is face normal, and 
\begin{equation}
    v_n = \mathbf{n}_f \left(\mathbf{v}_w - 
    \frac{\dot{\mathbf{x}}_0 + \dot{\mathbf{x}}_1 + \dot{\mathbf{x}}_3}{3} \right)
    \mbox{,} \nonumber
\end{equation}
\begin{equation}
    \mathbf{v}_t = \frac{\dot{\mathbf{x}}_0 + \dot{\mathbf{x}}_1 + \dot{\mathbf{x}}_3}{3}
    - v_n \mathbf{n}_f
    \mbox{.}\nonumber
\end{equation}
The forces on the nodes are 
\begin{equation}
    \mathbf{F}_{\mathbf{x}_0} = 
    \mathbf{F}_{\mathbf{x}_1} =
    \mathbf{F}_{\mathbf{x}_3} =
    \frac{1}{3} \mathbf{F}_w
    \mbox{.}
\end{equation}

\subsection{Collision Response}
We adopt a collision handling method originally designed for triangular meshes stored in bounding volume hierarchy \citep{tang_fast_2010} where continuous collision detection (CCD) can detect edge-edge and vertex-face collision. The detected vertices, edges, and faces are grouped into non-rigid impact zones \citep{harmon_robost_2008} for computing collision response. We treat collision response as a constrained optimization problem to prevent penetrations~\citep{liang_differentiable_2019}:
\begin{alignat*}{2}
& \underset{z}{\text{minimize}} &\quad&\frac{1}{2}(\mathbf{x}_{colli}-\mathbf{x})^{\top} \mathbf{W} (\mathbf{x}_{colli}-\mathbf{x})\\
& \mbox{subject to} &\quad&\mathbf{Gx}_{colli} + \mathbf{h} \leq \mathbf{0}
\end{alignat*}
where $\mathbf{W}$ is a weight matrix, $\mathbf{x}$ is the Lagrangian part of $\mathbf{q}$, $\mathbf{x}_{colli}$ is the updated $\mathbf{x}$ where no collision can be detected. $G$ and $h$ are constraint parameters. We assume neither self-collision nor cloth-object collision can generate considerable yarn-sliding motions, so we exclude the Eulerian terms.

\section{Derivatives of the simulator}
\label{sec:app_physicsSolve}
Now we have a fully differentiable cloth simulator. We then compute the loss $\mathcal{L}$ that indicates the difference between the predicted and ground truth cloth states. The loss gradients with respect to the parameters $\frac{\partial \mathcal{L}}{\partial w}$ can help learn the right physics parameters via back-propagation. For simplicity, we use $\mathbf{A\dot{q} = b}$ to represent Equation \ref{eqn:app_linear}. The differential of $\mathbf{A\dot{q} = b}$ is~\citep{magnus_matrix_2019}:
\begin{equation}
    \label{eqn:app_der_1}
     \mathbf{A} \mathsf{d} \dot{\mathbf{q}} = \mathsf{d} \mathbf{b} - \mathsf{d}\mathbf{A} \dot{\mathbf{q}}
\end{equation}
We can form the Jacobians of $\dot{\mathbf{q}}$ with respect to $\mathbf{A}$ or $\mathbf{b}$ with Equation \ref{eqn:app_der_1}. For example, to compute the $\frac{\partial \dot{\mathbf{q}}}{\partial \mathbf{A}}$, we need to set $\mathsf{d}\mathbf{A} = \mathbf{I}$ and $\mathsf{d}\mathbf{b}=\mathbf{0}$, then solve the equation and the result is $\frac{\partial \dot{\mathbf{q}}}{\partial \mathbf{A}}$. As pointed out by \citet{amos2017optnet}, it is unnecessary to explicitly compute these Jacobians in back-propagation. We want to compute the product of the vector passed from back-propagation, $\frac{\partial \mathcal{L}}{\partial \dot{\mathbf{q}}}$ and the Jacobians of $\dot{\mathbf{q}}$, i.e.$\frac{\partial \mathcal{L}}{\partial \dot{\mathbf{q}}} \frac{\partial \dot{\mathbf{q}}}{\partial \mathbf{A}}$ and $\frac{\partial \mathcal{L}}{\partial \dot{\mathbf{q}}} \frac{\partial \dot{\mathbf{q}}}{\partial \mathbf{b}}$. Assume $\mathbf{A} \in \mathbb{R}^{3\times3}$, $\dot{\mathbf{q}} \in \mathbb{R}^3$, and $\mathbf{b} \in \mathbb{R}^3$, then 
\begin{equation}
    \label{eqn:der_2}
    \frac{\partial \mathcal{L}}{\partial \mathbf{b}}
    = \frac{\partial \mathcal{L}}{\partial \dot{\mathbf{q}}} \frac{\partial \dot{\mathbf{q}}}{\partial \mathbf{b}}
    = \left(
    \begin{pmatrix}
        \frac{\partial \mathcal{L}}{\partial \dot{\mathbf{q}}_1} &
        \frac{\partial \mathcal{L}}{\partial \dot{\mathbf{q}}_2} &
        \frac{\partial \mathcal{L}}{\partial \dot{\mathbf{q}}_3}
    \end{pmatrix}
    \begin{pmatrix}
        \frac{\partial \dot{\mathbf{q}}_1}{\partial\mathbf{b}_1} & \frac{\partial \dot{\mathbf{q}}_1}{\partial \mathbf{b}_2} & \frac{\partial \dot{\mathbf{q}}_1}{\partial \mathbf{b}_3}\\
        \frac{\partial \dot{\mathbf{q}}_2}{\partial \mathbf{b}_1} & \frac{\partial \dot{\mathbf{q}}_2}{\partial \mathbf{b}_2} & \frac{\partial \dot{\mathbf{q}}_2}{\partial \mathbf{b}_3}\\
        \frac{\partial \dot{\mathbf{q}}_3}{\partial \mathbf{b}_1} & \frac{\partial \dot{\mathbf{q}}_3}{\partial \mathbf{b}_2} & \frac{\partial \dot{\mathbf{q}}_3}{\partial \mathbf{b}_3}
    \end{pmatrix}
    \right)^{\top} 
\end{equation}
As
\begin{equation}
    \label{eqn:der_3}
    \frac{\partial \dot{\mathbf{q}}_1}{\partial \mathbf{b}_1} 
    = \frac{ \partial
    \; (\mathbf{A}^{-1})_{1,1}\mathbf{b}_1
    + (\mathbf{A}^{-1})_{1,1}\mathbf{b}_2
    + (\mathbf{A}^{-1})_{1,1}\mathbf{b}_3}
    { \partial \mathbf{b}_{1}}
    = \mathbf{A}^{-1}_{1,1} \nonumber
\end{equation}
and similarly for $\frac{\partial \mathbf{x}_i}{\partial \mathbf{b}_j}$, Equation \ref{eqn:der_2} can be represented as: 
\begin{equation}
\label{eqn:der_4}
    \left(
    \begin{pmatrix}
        \frac{\partial \mathcal{L}}{\partial \dot{\mathbf{q}}_1} &
        \frac{\partial \mathcal{L}}{\partial \dot{\mathbf{q}}_2} &
        \frac{\partial \mathcal{L}}{\partial \dot{\mathbf{q}}_3}
    \end{pmatrix}
    \begin{pmatrix}
        (\mathbf{A}^{-1})_{1,1} & (\mathbf{A}^{-1})_{1,2} & (\mathbf{A}^{-1})_{1,3} \\
        (\mathbf{A}^{-1})_{2,1} & (\mathbf{A}^{-1})_{2,2} & (\mathbf{A}^{-1})_{2,3} \\
        (\mathbf{A}^{-1})_{3,1} & (\mathbf{A}^{-1})_{3,2} & (\mathbf{A}^{-1})_{3,3}
    \end{pmatrix}
    \right)^{\top}
    = (\mathbf{A}^{-1})^{\top} \frac{\partial \mathcal{L}}{\partial \dot{\mathbf{q}}}
\end{equation}
After computing $\frac{\partial \mathcal{L}}{\partial \mathbf{b}}$, we need to compute $\frac{\partial \mathcal{L}}{\partial \mathbf{A}}$. The $\mathsf{}\mathbf{b}$ in Equation \ref{eqn:app_der_1} can be set to 0 because it is irrelevant when computing $\frac{\partial \mathcal{L}}{\partial \mathbf{A}}$. Then we have
\begin{equation}
    \mathbf{A} \mathsf{d} \dot{\mathbf{q}} = - \mathsf{d}\mathbf{A} \dot{\mathbf{q}}
\end{equation}
The derivative of $\dot{\mathbf{q}}$ with respect to $\mathbf{A}_{i,j}$, the entry in the $i$th row and $j$th column of the matrix $\mathbf{A}$, is
\begin{equation}
    \frac{\partial \dot{\mathbf{q}}} {\partial \mathbf{A}_{i,j}} = \mathbf{A}^{-1}
    \begin{pmatrix}
        \mathbf{0} \\ -\dot{\mathbf{q}}_j \\ \mathbf{0}
    \end{pmatrix}
\end{equation}
According to chain rule, 
\begin{equation}
    \frac{\partial \mathcal{L}}{\partial \mathbf{A}_{i,j}} 
    = \frac{\partial \mathcal{L}}{\partial \dot{\mathbf{q}}} 
    \frac{\partial \dot{\mathbf{q}}}{\partial \mathbf{A}_{i,j}} 
    = \frac{\partial \mathcal{L}}{\partial \mathbf{b}}^{\top}\mathbf{A}\mathbf{A}^{-1}
    \begin{pmatrix}
        \mathbf{0} \\ -\dot{\mathbf{q}}_j \\ \mathbf{0}
    \end{pmatrix}
    = - \left(\frac{\partial \mathcal{L}}{\partial \mathbf{b}}\right)_i
    \dot{\mathbf{q}}_j
\end{equation}
The more general form is 
\begin{equation}
    \frac{\partial \mathcal{L}}{\partial \mathbf{A}} = -
    \frac{\partial \mathcal{L}}{\partial \mathbf{b}} \dot{\mathbf{q}}^{\top}
\end{equation}

\end{document}